\documentclass[12pt]{article}
\pdfoutput=1

\newcounter{dim-method}

\usepackage[margin=1in]{geometry} 

\usepackage{amsmath}
\usepackage{amssymb}

\usepackage{lscape}
\usepackage{times}
\usepackage{float}
\usepackage{placeins}
\usepackage{bm}
\usepackage{afterpage}
\usepackage{epsfig}
\usepackage{longtable}
\usepackage{setspace}
\usepackage{titlesec}
\usepackage{amsthm} 
\usepackage{mathtools}

\usepackage[utf8]{inputenc}
\usepackage{csquotes}
\usepackage{hyperref}

\usepackage[numbers]{natbib}
\usepackage{amsthm} 
\newtheoremstyle{myremark}
  {}
  {}
  {}
  {}
  {\bfseries}
  {.}
  { }
  {}
\theoremstyle{myremark}
\newtheorem{remark}{Remark}

\usepackage{graphicx, color}
\usepackage{blkarray}
\graphicspath{{figures/}}

\usepackage{algorithm, algpseudocode} 
\usepackage{mathrsfs} 

\usepackage{authblk}

\title{IAN: Iterated Adaptive Neighborhoods for manifold learning and dimensionality estimation}
\author[1]{Luciano Dyballa}
\author[1,2]{Steven W. Zucker}
\affil[1]{Department of Computer Science, Yale University, New Haven, CT}
\affil[2]{Department of Biomedical Engineering, Yale University, New Haven, CT}
\date{}                     
\begin{document}

\maketitle

	\begin{abstract}
Invoking the manifold assumption in machine learning requires knowledge of the manifold's geometry and dimension, and theory dictates how many samples are required. However, in most applications the data are limited, sampling may not be uniform, and the manifold's properties are unknown; this implies that neighborhoods must adapt to the local structure. We introduce an algorithm for inferring adaptive neighborhoods for data given by a similarity kernel. Starting with a locally-conservative neighborhood (Gabriel) graph, we sparsify it iteratively according to a weighted counterpart. In each step, a linear program yields minimal neighborhoods globally, and a volumetric statistic reveals neighbor outliers likely to violate manifold geometry. We apply our adaptive neighborhoods to non-linear dimensionality reduction, geodesic computation, and dimension estimation. A comparison against standard algorithms using, e.g., $k$-nearest neighbors, demonstrates the usefulness of our approach.
\end{abstract}
\bigskip
{\footnotesize Research supported by NIH Grant EY031059, NSF CRCNS Grant 1822598, and the Swartz Foundation. This project derived from problems in manifold inference involving neuroscience data. We thank G. Field and M. Stryker for motivating discussions.}



\section{Introduction}

A starting point for many algorithms in data science---from clustering to manifold inference---is knowing the neighbor relationships among data points. Clustering, for example, often begins with a ``$k$-nearest neighbor graph,'' while manifold inference involves a kernel, i.e., a measure of similarity between data points. In the first case, the neighborhoods are local and discrete; in the second, they are global and continuous, with concentration of influence controlled by the kernel bandwidth, or scale. Such neighbor relationships are fundamental to defining a topology. Moreover, dimensionality may be estimated based on the rate of change in the density of points within a ball, that is, within a neighborhood, with respect to its radius. It is helpful when the number of data points is large, a requirement that grows with dimensionality; asymptotic analysis is often favored by theoreticians.

In practice, we rarely have enough data points to satisfy asymptotic bounds. Nor are we given the precise number of neighbors, $k$, that each point should have. We often make the manifold assumption---that the data points are drawn randomly from a (or near a) manifold---but rarely try to assess the basic properties of the manifold assumed by theorists: its dimensionality, sampling density, curvature, medial axis, or reach (defined in the next section). All of these could influence $k$.

Instead, we rely on different visualization algorithms, such as Isomap, diffusion maps, t-SNE, and many others (references in the next section), to find a pleasing organization of the data. This is dangerous, of course, because these algorithms have free parameters. In particular, and central to this paper, most require specifying the number of neighbors, $k$ (or its equivalent): changing $k$ or other parameters changes the result. Unless one knows the answer, one is caught in a conundrum: imposing a prior belief amounts to ``fixing'' the solution (examples of changing $k$ are shown later in the paper).

\begin{figure}[!ht]
  \centering
  \includegraphics[width=1\textwidth]{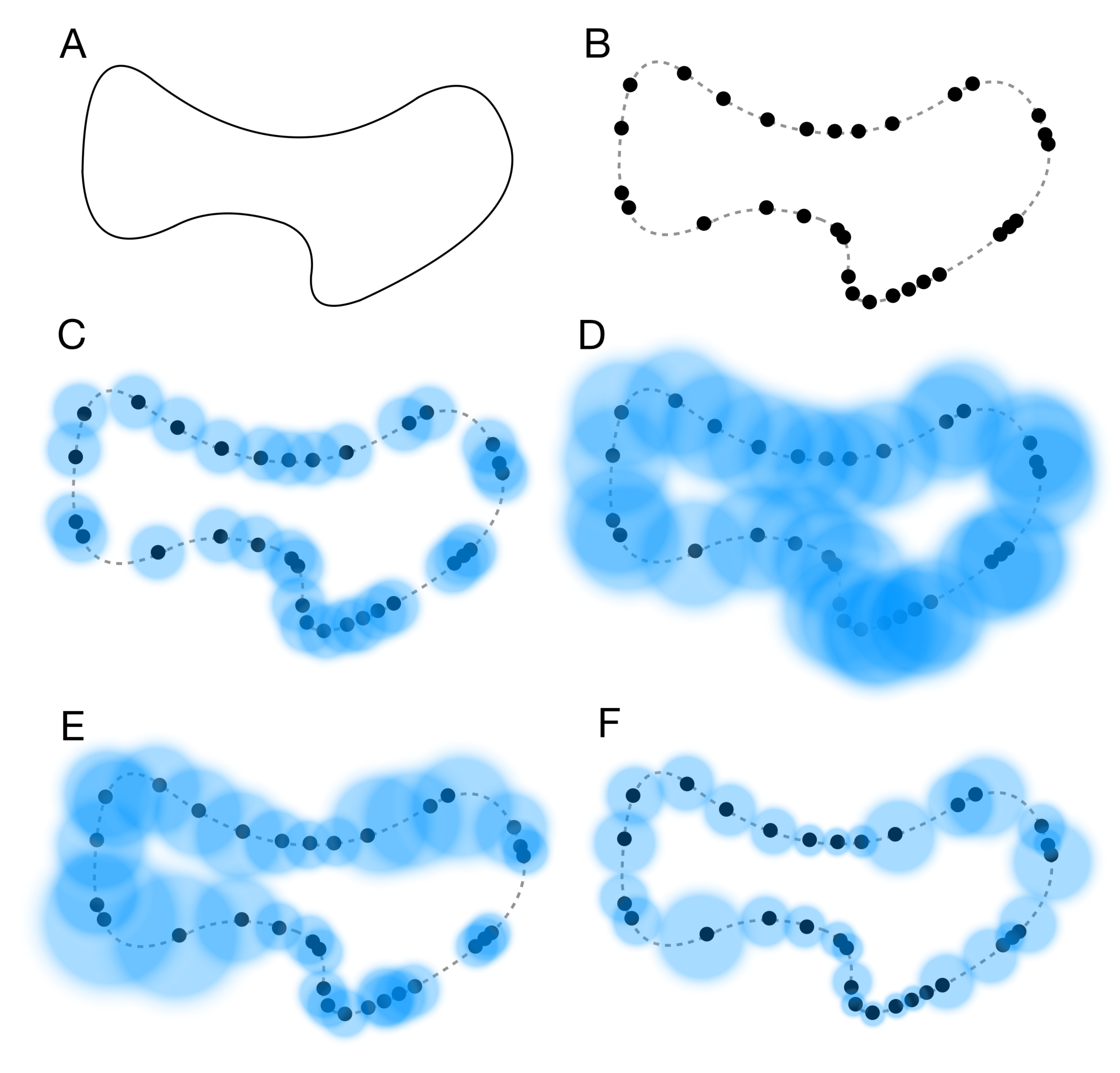}
  \caption{Inferring the geometry of manifolds requires neighborhoods around each given data point. Setting the correct scale for these neighborhoods, shown as balls, is fundamental. 
  (\textbf{A}) Example of a 1-dimensional manifold, $\mathcal{M}$.
  (\textbf{B}) Collection of points sampled from an unknown distribution over $\mathcal{M}$. Their pairwise distances are the only available data; properties of $\mathcal{M}$ are not given \textit{a priori}.
  (\textbf{C}) Using a global kernel scale: if it is too small, the manifold will appear disconnected, artificially producing clusters. Notice how some balls do not touch.
  (\textbf{D}) If it is too big, the manifold may collapse, giving rise to incorrect geometry/topology. Notice how the balls overlap (covering dimension).
  (\textbf{E}) The use of local scales based on a global number of nearest neighbors (in this example, $k$ = 2) is still susceptible to the problems above.
  (\textbf{F}) Our approach computes locally adaptive neighborhood sizes, resulting in scales that conform to the local geometry and sampling. 
  }
  \label{fig:manifold-intro}
\end{figure}

This gap between theory and practice shows up right from the start. If the manifold is not pure, i.e., if it consists of a union of manifolds of possibly different dimensionality, then there may be no global $k$ that suffices; furthermore, the manifold may have a boundary. Even if it is pure and without boundary, the temptation to choose $k$ large is common. But this can incorrectly fill in the open space around curved manifolds (``folding'', or ``short-circuiting''), linking distant points that should not be neighbors. On the other hand, choosing $k$ small can induce holes and break connectivity. Such phenomena are illustrated in Figure~\ref{fig:manifold-intro}. As we shall demonstrate, sampling issues and manifold geometry interact in causing these. Moreover, in real datasets the appropriate number of neighbors may differ from point to point. This final issue is a principal motivation for this paper.

We present an algorithm to estimate an effective neighborhood---the immediate neighbors, or scale of a similarity kernel---around each point. We seek to identify those nearest neighbors that are ``correct'' in the sense that they support dimensionality and volume estimates, and manifold inference in general, without covering holes or filling in concavities. It is inspired by the philosophical position that views discrete and continuous mathematics as ``two sides of the same,'' as argued by \citet{lovasz2010discrete}, and iterates between them.

Our algorithm builds from a conservative initial estimate of neighbors (based on a discrete construct, the Gabriel graph) toward a refined one, based on continuous estimates from a multiscale Gaussian kernel. The discrete and continuous volume estimates must be consistent, however, and this provides the glue for our iteration. Since not all of the initial putative neighbors may actually be closest neighbors, those neighbors that violate the volume relationship are pruned, and the process repeats until the two perspectives agree. Our algorithm, thus, can be considered an iterative graph sparsification.

Technically, it involves two different graphs: a discrete one, that links only putative nearest neighbors (pairs of points defining the diameter of an otherwise empty ball), and a weighted one, structured by a multiscale Gaussian kernel, whose individual scales must cover the neighborhood given in the discrete graph. Keeping the two graphs consistent is another way to think about our iteration. Each resulting graph can be applied to many different algorithms for data visualization, dimensionality reduction, and manifold inference.

\begin{figure}[!ht]
  \centering
  \includegraphics[width=1\textwidth]{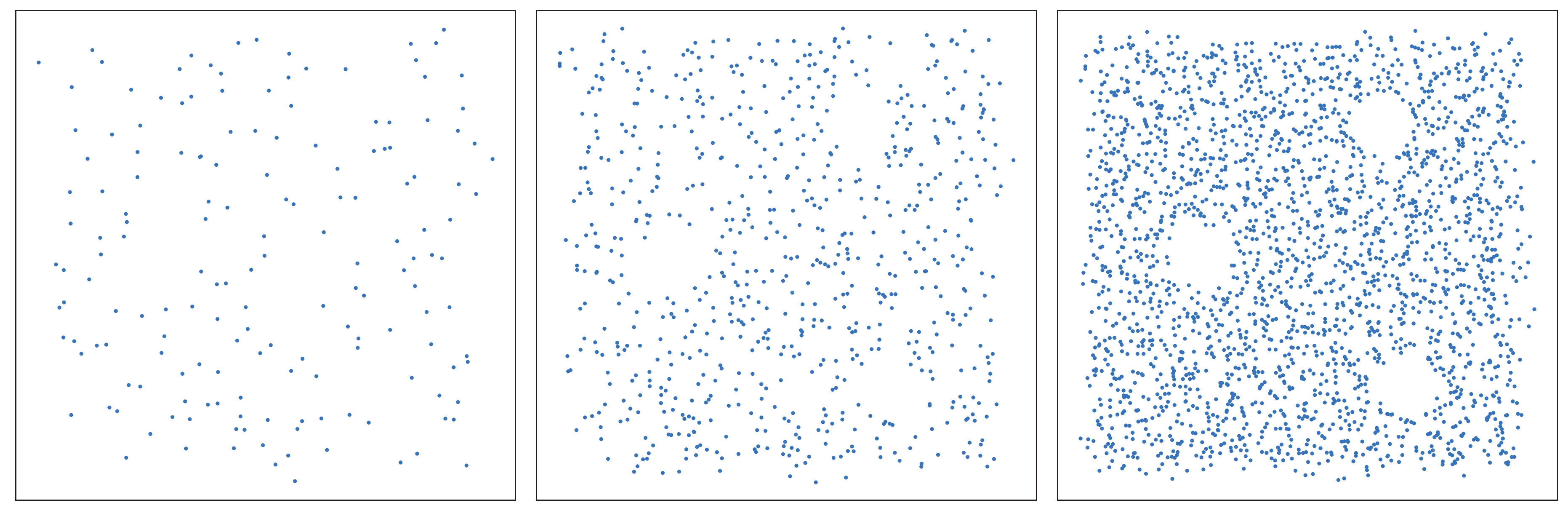}
  \caption{Sampling a Swiss cheese: the available data constrain manifold complexity. Viewed from left to right, as the number of
  sample points increases, the apparent manifold goes from a plane to a plane with holes. The central panel shows the sampling density for which actual holes in the manifold become roughly distinguishable from holes due to sampling. The results of our algorithm on these examples are shown in Figure~\ref{fig:swiss-cheese}. 
  }
  \label{fig:swiss-cheese-pts}
\end{figure}

Our approach to the problem is in the spirit of {\it exploratory data analysis}; it works with the available data. This provides another view regarding the interaction between sampling and geometry: one can only do as well as the available data allow (see Figure~\ref{fig:swiss-cheese-pts}). The situation is analogous to that in learning theory, where there is a trade-off between the accuracy of the learner and the coarseness of the hypothesis class over which she is learning \citep{alon1997scale}. Here, the space of manifolds over which inferences are made is dictated by the available samples. 

An overview of the paper is as follows. In the next section, we review the background in some detail, covering both the zoo of similarity kernels that exist plus several relevant notions, such as the reach of a manifold, that are well studied in the theory literature. The discussion is organized to emphasize the centrality of scale, or neighborhood, in all of the references. In section~\ref{sec:algorithm} we provide an overview of our algorithm. It includes a brief sketch of both graphs we work with, plus the connection back to manifolds. Pseudocode for the algorithm is given in Algorithm~\ref{algo1}, which also includes pointers to where each of its steps is developed.

We then expand on the algorithm. In section~\ref{section:connect-criterion} we study the Gabriel graph and putative neighbors. Two features are emphasized: scale-free neighborhoods and the relationship between node degree and local dimensionality. A structural criterion is revealed, showing how putative edges between neighbors fill ``volumes'' that block others from being neighbors. This graph serves as an initialization. It is then refined iteratively in several steps. First, continuous kernel scales are computed based on the discrete, putative neighbors. A linear program relaxation bridges local scales to a global cover, in which each node's weighted degree is comparable to the number of its neighbors. In other words, each neighborhood radius should not cover too many outside points. If it does, then it indicates that the neighborhood itself should be refined. That is, some putative scales are likely wrong, in the sense that their neighborhood contains an extreme outlier. This leads directly to a volumetric statistic (section~\ref{section:delta-stat}), and to a pruning technique for sparsifying edges from the discrete graph. The process iterates until there are no more outliers.

In section~\ref{section:applications}, we evaluate the results for estimating manifold low-dimensional embeddings, geodesics, and local intrinsic dimensionality. Comparisons against popular algorithms, such as UMAP and t-SNE, illustrate the power of the approach. In the end, we demonstrate that it is possible to infer data-driven local neighborhoods that remain consistent with geometric and topological properties of manifolds. Code for our algorithm is available at $\texttt{github.com/dyballa/IAN}$.


\section{Background}\label{section:background}

Manifold learning is a vast area of machine learning where high-dimensional data are analyzed based on the assumption that they were sampled from a low-dimensional manifold, $\mathcal{M}$ \citep{fefferman2016testing}, in which case geodesic distances over $\mathcal{M}$ provide a better description of the relationships between data points than Euclidean distances in ambient space \citep{belkin2004semi}. The manifold assumption finds applications in non-linear dimensionality reduction \citep{vdmaaten-review}, de-noising \citep{hein2006denoising}, interpolation \citep{bregler1994interp}, dimensionality estimation \citep{camastra2016intrinsic}, computational geometry \citep{crane-geoheat}, and more.

Since $\mathcal{M}$ locally resembles Euclidean space, it is standard to define a similarity kernel to define (possibly weighted) neighborhoods around each point in terms of other points. This naturally leads to a graph having data points as nodes and similarity values as edge weights. Then, by computing the graph Laplacian, one can apply a variety of methods from spectral graph theory \citep[see, e.g.,][]{spielman2012spectral}. Formal analysis involves the limit as the number of data points grows large; the practical success of such methods depends on how well graph neighborhoods capture the topology and geometry of $\mathcal{M}$. 

We here review the many approaches to specifying a similarity kernel or a local neighborhood. Let $\mathcal{M}$ be a $d$-dimensional manifold in ambient space $\mathbb{R}^n$. When only pairwise distances are known, an intuitive approach is to define the neighbors of a point, $\bm{x}_i$, as those within a certain distance threshold, or, equivalently, inside an $n$-dimensional ball around $\bm{x}_i$. A kernel function assumes the role of this ball, by assigning values to neighboring points as a function (discrete or continuous) of how close they are to $\bm{x}_i$. The question becomes: what kernel size should be used for each point? 

\subsection{Similarity kernels}\label{section:kernels}

Consider a set of points $\mathcal{X} \in \mathbb{R}^n$. Typically, a symmetric, positive semi-definite similarity kernel \citep{scholkopf2002learning} is chosen to determine weighted connections between data points based on the ambient Euclidean distances between them. For each pair of data points ${\bm{x}_i},{\bm{x}_j} \in \mathbb{R}^n$, it returns a number between 0 and 1 which determines how close, or strongly connected, they are. This effectively defines a neighborhood around each point.

\subsubsection{Discrete kernels}

Possibly the simplest choice for a kernel is the $\varepsilon$-neighborhood \citep[e.g.,][]{eigenmaps}:
\begin{equation}\label{eq:eps-nbrhood}
K_{ij}(\varepsilon)= 
\begin{dcases}
    1,& \text{if } \|\bm{x}_i-\bm{x}_j\| < \varepsilon\\
    0,              & \text{otherwise},
\end{dcases}
\end{equation}
where $\|\cdot\|$ is typically the Euclidean norm in $\mathbb{R}^n$. This results in discrete-like neighborhoods whose sizes may be quite sensitive to the choice of $\varepsilon$, so implicit is the assumption that sampling is approximately uniform.

Instead of defining a neighborhood radius, a more common approach is to specify the number of neighboring points, $k$. Letting $\mathcal{N}_k(\bm{x}_i)$ be the set containing the $k$ points closest to $\bm{x}_i$ in $\mathbb{R}^n$ (not including $\bm{x}_i$\footnote{Throughout, when referring to a point's set of $k$-nearest neighbors, we shall not include the point itself (unless otherwise stated), and further assume that no two points are identical.}), a $k$-nearest neighbors kernel can be defined as:
\begin{equation}\label{eq:knn-graph}
K_{ij}(k)= 
\begin{dcases}
    1,& \text{if } \bm{x}_j \in \mathcal{N}_k(\bm{x}_i)\\
    0,              & \text{otherwise},
\end{dcases}
\end{equation}
which is commonly symmetrized by making $K_{ij}(k) = 1$ if $\bm{x}_j \in \mathcal{N}_k(\bm{x}_i) \vee \bm{x}_i \in \mathcal{N}_k(\bm{x}_j)$.

\subsubsection{Continuous kernels -- global scale}

In order to have the kernel values decrease with increasing distance between data points, a Gaussian kernel is commonly used:
\begin{equation}\label{eq:gauss-kernel}
    K_{ij}(\sigma) = \mathrm{exp}\left({\frac{-{\left \| \bm{x}_i - \bm{x}_j \right \|^2}}{\sigma^2}}\right).
\end{equation}
This gives a continuous similarity scale from 1 (when ${\bm{x}_i}$ and ${\bm{x}_j}$ are identical) down to some predetermined cutoff below which the kernel is considered to be zero (meaning no connection in the data graph). Such a threshold is typically chosen to be a very small value, often at the limit of numerical precision, and is often required to ensure compactness of the kernel.

One would like the parameter $\sigma$ to be just large enough to be able to capture local manifold patches. There are several heuristics for finding such a scale: the median of all pairwise distances in $\mathcal{X}$ (or another percentile), the mean (or median) of the distances to each point's $k^{\mathrm{th}}$ nearest neighbor \citep{lafonthesis}, or a scalar multiple of the maximal distance from a point to its nearest neighbor in the data \citep{audiovisual}. Also common is to choose a scale so that each data point is sufficiently connected to at least one other point \citep{datafusion}.

A different approach is based on inspection of the curve given by the sum of pairwise kernel values. When the double-sum $\sum_{i,j} K_{ij}(\sigma)$ is plotted against $\sigma$ using a log-log scale, the slope
\begin{equation}\label{eq:slope-singer}
\frac{\mathrm{d}\log \sum_{i,j} K_{ij}(\sigma)}{\mathrm{d}\log \sigma}
\end{equation}
is proportional to the intrinsic dimensionality of the data \citep{laplaciantomography}. A global scale is then chosen from within a linear region of such curve.
In \cite{theiss}, a similar procedure is proposed that considers, instead, the curve given by the weighted average of the degrees $Z_i(\sigma) = \sum_j K_{ij}$ of each data point $\bm{x}_i$, after taking the logarithm:
\begin{equation}\label{eq:meannbrsumeps}
\left \langle \log Z_i(\sigma) \right \rangle =  \frac{\sum_i\log Z_i(\sigma)\cdot(1/Z_i(\sigma))} {\sum_i(1/Z_i(\sigma))},
\end{equation}
The use of the inverse of each point's degree as weights is intended to compensate for density heterogeneities.
The choice of $\sigma$ is then made precise by choosing the argmax of the slope of $\left \langle \log Z_i(\sigma) \right \rangle$ plotted against $\log \sigma$, which in many cases should occur near the center of the linear region of equation~\ref{eq:slope-singer}. 
One complication occurring in both approaches, however, is that more than one linear section (and, equivalently, more than one local maximum of the slope) may exist, requiring that additional criteria be defined to make the choice of $\sigma$ truly automated. 

\subsubsection{Continuous kernels -- multiscale}

A more localized strategy is to use a \emph{multiscale kernel}, where each point has an individual scale, or bandwidth. Instead of a single, global scale, there are now $N$ parameters. The advantage is that, if the scale selection is adequate, the kernel may capture the characteristics of more complex datasets and manifolds that have non-uniform sampling and geometry.

In the self-tuning method \citep{zelnikmanor}, local scales are used in a Gaussian kernel by replacing the global scale $\sigma$, from equation~\ref{eq:gauss-kernel}, by $\sqrt{\sigma_i\sigma_j}$, where $\sigma_i$ and $\sigma_j$ are the scales assigned to $\bm{x}_i$ and $\bm{x}_j$, respectively. This results in the symmetric kernel:
\begin{equation}\label{eq:multiscale}
    K_{ij}(\sigma_i,\sigma_j) = \mathrm{exp}\left(\frac{-{\left \| \bm{x}_i - \bm{x}_j \right \|^2}}{\sigma_i\sigma_j}\right).
\end{equation}
Each $\sigma_i$ is set as the distance to the $k^{\mathrm{th}}$ nearest neighbor of $\bm{x}_i$; authors recommend $k$ = 7 \citep{zelnikmanor,mishnemultiscale}.

In \cite{BGH1} and \cite{BGH2}, a variable bandwidth kernel is proposed that combines the use of local bandwidths with a global scale parameter, $\epsilon$. The kernel then takes the form:
 \begin{equation}\label{eq:bgh}
    K_{\varepsilon}\left (\bm{x}_i,\bm{x}_j\right ) = \mathrm{exp}\left(\frac{-{\left \| \bm{x}_i - \bm{x}_j \right \|^2}}{4\epsilon(q_{\epsilon}(\bm{x}_i)q_{\epsilon}(\bm{x}_j))^\beta}\right),
\end{equation}
where $q_{\epsilon}$ is a local density function and $\beta$ an additional (non-positive) parameter. An initial estimate for the local bandwidth around each point $\bm{x}_i$ is set as the square-root of the mean squared distance to the $k$-nearest neighbors of $\bm{x}_i$, with $k$ = 8. Finally, $\epsilon$ is automatically tuned as the argmax of equation~\ref{eq:slope-singer} above; however, the authors do not consider cases in which more than one local maximum may exist.

Other methods also adopt individual bandwidth parameters, but use asymmetric kernels that are symmetrized \textit{a posteriori}. In the t-SNE algorithm \citep{tsne}, the single-scale Gaussian kernel 
\begin{equation}\label{eq:tsne-kernel}
    K_{ij}(\sigma_i) = \mathrm{exp}\left(\frac{-{\left \| \bm{x}_i - \bm{x}_j \right \|^2}}{2\sigma_i^2}\right)
\end{equation}
gives a measure of affinity, or similarity, between pairs of points. It is then normalized as 
\begin{equation}
    p_{j|i}(\sigma_i) = \frac{K_{ij}(\sigma_i)}
    {\sum_{k\neq i} K_{ik}(\sigma_i)}
\end{equation}
to yield transition probabilities, and finally symmetrized as
\begin{equation}\label{eq:tsne-sym}
    p_{ij}(\sigma_i,\sigma_j) = \frac{1}{2N}\left(p_{j|i}(\sigma_i) + p_{i|j}(\sigma_j) \right).
\end{equation}
Each $\sigma_i$ is fit to $\bm{x}_i$ so that the distribution of $p_{j|i}, \forall j$ attains entropy $H_i$ such that its perplexity, $2^{H_i}$ (a real-valued number representing the ``effective number of neighbors''), approximates some prespecified value, $k$. The authors recommend a value for $k$ between 5 and 50.

In the UMAP algorithm \citep{umap}, an exponential kernel is used instead of the typical Gaussian. Using a prespecified neighborhood size, $k$, let $\mathcal{N}_k(i)$ be the set of $k$-nearest neighbors of $\bm{x}_i$. With $\rho_i$ as the distance to the nearest neighbor of $\bm{x}_i$, the kernel has the form
\begin{equation}\label{eq:umap-kernel}
    K_{ij}(\sigma_i) = \mathrm{exp}\left(\frac{-{\max\{0,\left \| \bm{x}_i - \bm{x}_j \right \| - \rho_i\}}}{\sigma_i}\right), j \in \mathcal{N}_k(i),
\end{equation}
and is symmetrized as
\begin{equation}\label{eq:umap-sym}
    U_{ij}(\sigma_i,\sigma_j) = K_{ij}(\sigma_i) + K_{ji}(\sigma_j) - K_{ij}(\sigma_i)K_{ji}(\sigma_j).
\end{equation}
It can be seen as a hybrid between continuous and discrete, since $U_{il}$ is set to zero for any point $\bm{x}_l$ not in $\mathcal{N}_k(i)$. Each $\sigma_i$ is fit to $\bm{x}_i$ so that $\sum_j K_{ij}(\sigma_i)$ approximates $\log_2 k$ (loosely analogous to the perplexity approach from t-SNE).

\subsubsection{Adaptive neighborhood size methods}

Other methods attempt to automatically determine optimal
neighborhoods. Most of these are based on determining an optimal $k$ for a $k$-nearest neighbors ($k$-NN) graph; this can be done either globally or by selecting a local neighborhood size $k_i$ around each point $\bm{x}_i$, known as adaptive neighborhood selection \citep{vdmaaten-review}.

Some approaches optimize a global $k$ based on its performance in a specific embedding algorithm. For instance, the method from \cite{samko2006selection} is tailored to Isomap \citep{isomap}, while others \citep{kouropteva2002selection,alvarez2011global} apply to LLE \citep{LLE}. In \cite{alvarez2011global}, a local method is additionally proposed that produces a nearest-neighbor graph with variable $k_i$, under the assumption that the manifold is connected.

Others are based on first estimating the local tangent space around each point, then setting $k_i$ to include as neighbors those points that are close to it. Such methods \citep[e.g.,][]{wang2004adaptive,mekuz2006parameterless} typically work with positional information for the tangent space computation (usually via SVD).

Also available are methods that are not based on the nearest-neighbors concept. In computational geometry, the idea of refining an initial estimate of connectivity from a simplicial mesh has been used before, usually specific to the case when $d$ = 2 and $n$ = 3, i.e., surfaces in 3-D space \citep{amenta1998new,amenta1999surface, ballpivoting,belkin2008discrete}. Other approaches extend this idea to arbitrary dimension  \citep{belkin2009constructing,boissonnat2009witness}, but still require knowledge of $d$. Most of the algorithms in this class use point clouds as input, so they can exploit positional information to decide on the appropriate neighborhood/connectivity.

Among the myriad ways of estimating neighborhoods, there is little agreement on which is most successful; see \cite{averbuch} for a review. Before proceeding to our algorithm, then, it is helpful to first understand what makes this such a hard problem. How can it fail, and what requirements must it fulfill in order to properly capture the topology and geometry of $\mathcal{M}$? This brings us to the geometry of manifolds.

\subsection{Reach and the geometry of manifolds}\label{section:reach}

The neighborhoods implied by a kernel should agree with $\mathcal{M}$, or at least approximate a tubular neighborhood of it. As exemplified in Figure~\ref{fig:manifold-intro}, if neighborhoods are too small, the implied manifold may become disconnected, i.e., falsely divided into disjoint sub-manifolds or clusters \citep{samko2006selection}; if too large, they may cause $\mathcal{M}$ to self-intersect, collapsing bottlenecks or curved regions, or cause ``smoothing,'' or ``folding.'' Such shortcomings are well-known in the manifold inference literature---while the former case typically occurs due to non-uniform sampling, the latter is mainly caused by an incompatibility between the sampling rate and the \emph{reach} of $\mathcal{M}$ \citep{federer1959curvature, thale200850}. We now expand on these points.

\begin{figure}[!ht]
  \centering
  \includegraphics[width=0.85\textwidth]{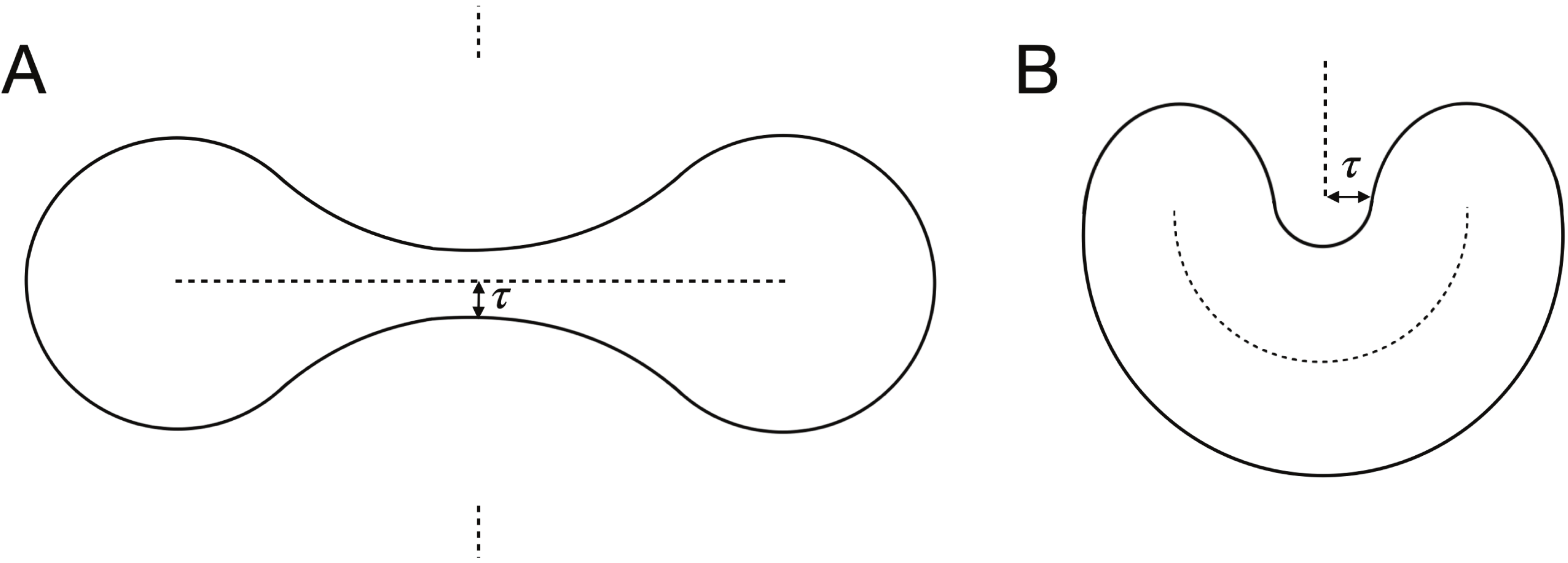}
  \caption{The reach, $\tau$, is a measure of the shape of a manifold. (\textbf{A}) A 1-dimensional manifold $\mathcal{M}$ with a bottleneck; the reach (double arrow) is the smallest distance between $\mathcal{M}$ and its medial axis (dashed curves). (\textbf{B}) A highly curved manifold; now the reach indicates the high curvature region.
  }
  \label{fig:manifold-reach}
\end{figure}

Letting the \emph{medial axis} of $\mathcal{M}$ be the set of points in $\mathbb{R}^n$ with at least two closest points in $\mathcal{M}$, the reach, $\tau$, can be defined as the minimum distance from $\mathcal{M}$ to its medial axis. Locally, it is constrained by the minimal radius of curvature (i.e., maximal curvature of a geodesic through $\mathcal{M}$); globally, it is constrained by the presence of bottlenecks (Figure~\ref{fig:manifold-reach}). The reach encodes essential geometric properties of $\mathcal{M}$, and has been widely used in the manifold learning community \citep{amenta1999surface,belkin2008discrete,niyogi2008finding,boissonnat2009witness,niyogi2011topological,genovese2012minimax,little2017multiscale,fefferman18fitting, aamari2019estimating,boissonnat2019reach}. It approximates the size of the largest ball in ambient $\mathbb{R}^n$ such that points in $\mathcal{M}$ can be seen as lying in Euclidean space $\mathbb{R}^d$ \citep{block2021intrinsic}.
A related concept, the \emph{local feature size} of a point $\bm{x}_i \in \mathcal{M}$, is the smallest distance between $\bm{x}_i$ and the medial axis of $\mathcal{M}$, so $\tau$ can be seen as the infimum of the local feature size anywhere on $\mathcal{M}$ \citep{belkin2009constructing}.

When $\tau$ is positive, it provides a measure of the ``local distortion'' \citep{block2021intrinsic}; the larger it is, the easier inference becomes. Some authors \citep[e.g.,][]{narayanan2010sample, fefferman2016testing} assume large reach in order to test the manifold hypothesis and to find bounds on the required sample size. In \cite{block2021intrinsic}, the reach is used when establishing bounds on the quality of an intrinsic dimensionality estimation based on $k$-nearest neighbors.

Obtaining a good representation of $\mathcal{M}$, therefore, requires consideration of its reach. In terms of our problem of finding an appropriate kernel, this effectively means that no neighborhood radius should cross the medial axis of $\mathcal{M}$.

Sampling is a further complication, and essentially what makes this a hard problem: when it is nonuniform and sparse (common in real-life datasets), it is not always clear whether the space between points constitutes an undersampled piece of $\mathcal{M}$, a hole, or a gap between disjoint submanifolds (cf. Figure~\ref{fig:swiss-cheese-pts}). The latter two conditions, of course, relate to reach. \citet{narayanan2010sample} prove that the number of required samples depends polynomially on curvature, exponentially on intrinsic dimension, and linearly on intrinsic volume. Aspects of our algorithm address each of these during the iteration process.

In all such cases, choosing a globally-fixed radius is likely to be problematic. While defining neighborhood size based on a fixed number $k$ of neighbors can be helpful to deal with nonuniform density (since the neighborhood radius adapts to the local pairwise distances), it is bound to violate the reach if $k$ is too large. It will also be a problem when the intrinsic dimensionality is not constant throughout $\mathcal{M}$, as higher dimensions require exponentially more neighbors.

\citet{mekuz2006parameterless} point out the lack of a principled way for setting this parameter, which in practice is often tuned empirically based on prior knowledge of the desired output. As put by \citet{wang2004adaptive}, the effectiveness of manifold learning algorithms depends on how nearby neighborhoods overlap and on the interplay between the curvature of the manifold and sampling density.

In terms relevant to this paper, the neighborhood radius should be smaller than the local feature size, but large enough to account for sampling variability and local dimensionality. We propose an iterative approach to developing a kernel, so that it can adapt appropriately to the neighborhood characteristics around each point.


\section{The algorithm}\label{sec:algorithm}

We here overview our algorithm for finding the neighborhood scale around each point in a manner that makes it globally consistent as a covering of the data points. As is common in manifold learning, we start with a pairwise distance matrix, not the points themselves. The first step is to build a graph in which each datum is connected to an appropriate neighborhood containing other data points. This data graph defines a topology; we refer to it as the {\it neighborhood graph}.

As we reviewed above, in the discrete case one might choose $k$-nearest neighbors, while in the continuous kernel case there is a bandwidth parameter that effectively defines a ``ball of influence'' around each point. \emph{Scale} is the radius of such a ball; a level set of the kernel function that essentially contains those neighbors whose weights are non-trivial. Our goal, then, is to find those scales---or neighborhoods---that support non-linear dimensionality reduction, geodesic estimation and, in general, manifold inference from the given pairwise distances. We do not have sampling guarantees, so will develop a statistic to check whether reach and curvature constraints might be violated. 

\subsection{Subtleties of scale}\label{section:subtleties}

Since scale may not be constant across the data set, we argue that it should be the first property to be inferred from the data. We start by imposing the manifold assumption, but from an empirical perspective. Unlike most theoretical studies, we do not assume the manifold is pure, i.e., that it has constant dimension. In a simple case, the data may be drawn from a union of different manifolds whose dimensions are not known \textit{a priori}---such datasets have been considered infrequently, although exceptions exist \citep[e.g.,][]{haro2008translated,little2017multiscale}.

Second, we do not know the sampling rate, or density. Rather, we build it up, conservatively, with putative nearest neighbors to each data point, by imposing a necessary (but not sufficient) condition. These putative neighbors will be refined, as the algorithm iterates, to achieve sufficiency. While the manifold assumption does imply the existence of local neighborhoods, their size may vary over the dataset; we require that the sampling be nearly constant over each of them. In effect, the density of points must be determined locally while respecting the global manifold geometry. 

\begin{figure}[h!]
  \centering
  \includegraphics[width=1\textwidth]{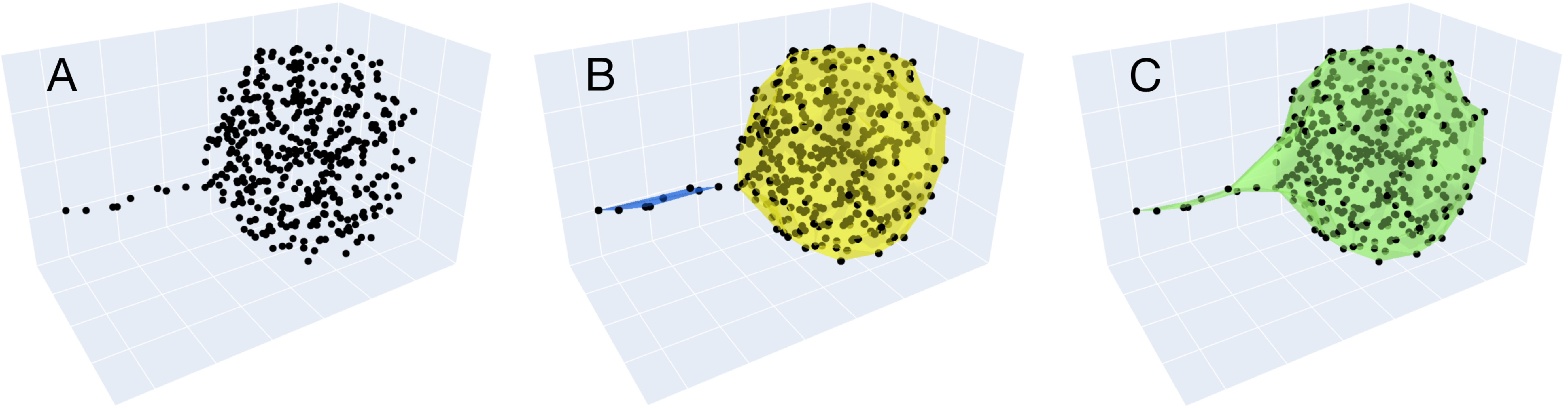}
  \caption{The ``manifold'' subtleties of complex datasets. (\textbf{A}) Sampled data from a non-linear system that includes two regimes. (\textbf{B}) It may be the case that the data in each regime define separate manifolds, shown by color. After sampling their union, however, the evidence for the separation is absent. (\textbf{C}) Or the data may be drawn from a single, connected manifold whose geometric properties change rapidly. In both cases, the intrinsic dimensionality appears different in the spike \textit{vs.} the ball. Colored meshes indicate underlying manifolds.
  }
  \label{fig:manifold-probs}
\end{figure}

We illustrate the complexity of this situation in Figure~\ref{fig:manifold-probs}. Shown is a data sphere with an apparent spike emerging from it. On one hand, such complex datasets could derive from two unrelated systems, which only appear to connect through their embeddings. On the other hand, the data could derive from a non-linear system that includes two regimes, one responsible for the spherical data and the other for the spike.  To handle the first situation, we must allow datasets to consist of unions of manifolds. This suggests the interpretation in Figure~\ref{fig:manifold-probs}-B, where the separation is obscured by sampling. Since manifolds with boundary and high curvature are also possible, the situation in Figure~\ref{fig:manifold-probs}-C arises. There is an apparent change in {\em intrinsic dimension} due to the small reach in the spike and the large boundary curvature. Because the (3-D) spike is so narrow, sampling suggests it is 1-dimensional, while the bulk of the points derive from a 3-D manifold.

We submit that such situations occur in real datasets and, since the data are fixed, we cannot appeal to knowing the sampling density or the manifold dimensions and reach. Instead, we address the interplay between manifold reach and sampling density pragmatically. Along the spike, the data appear to be 1-D; in the ball, 3-D. We seek a neighborhood graph that supports these inferences, so ``most'' points enjoy a neighborhood that agrees with their apparent dimension. At the join (or high-curvature neck), it is unclear. Moving from the spike to the ball suggests that dimension should be increasing; from the ball to the spike, it should be decreasing. For the neighborhood graph, most points along the spike should see $\sim$2 neighbors, and most points in the ball should see $\sim$2$^3$ neighbors; the problematic points should see something intermediate. Such results will be shown to follow from our algorithm.

We claim that either of the alternatives is worse; one should not impose an apparent dimensionality (or connectivity in the neighborhood graph) globally. To wit, if small numbers of neighbors (appropriate for the spike) are enforced on the ball, then holes are likely to be introduced. Or, if too many neighbors are enforced on the spike, it will collapse on itself. Both change the topology drastically (these situations are illustrated later, in Figures~\ref{fig:graphs-stingray}--\ref{fig:scales-stingray}).

\subsection{Overview of the algorithm}

Let the dataset, $\mathcal{X}$, be a sampling of a (possibly non-pure) manifold $\mathcal{M} = \cup_\alpha \mathcal{M}_\alpha$, with the dimension of each component $\mathcal{M}_\alpha$ denoted by $d_\alpha$. It consists of $N$ points in ambient space $\mathbb{R}^n$, where $n \geq d_\alpha, \forall \alpha$. The manifold may have a boundary, and the number of components is not known \textit{a priori}.

We work with two graphs: the first unweighted, and the second with edge weights given by a kernel. Our strategy is to begin with a conservative estimate of the unweighted graph, and extend it to a global weighted graph that suggests an estimated manifold covering. The validity of this extension is evaluated by a measure of volume in both graphs; an iterative algorithm is used to infer individual local scales for each point $\bm{x}_i$. Before presenting the algorithm, we introduce the two graphs.

Let the unweighted graph be $G=(V,E)$, with $|V|=N$ and adjacency matrix $A$ with entries $a_{ij}$, where to each point $\bm{x}_i \in \mathcal{X}$ is associated a node $i \in V$. We denote its initial estimate by $G^{(0)}$; successive refinements are indicated as $G^{(t)}$ until convergence ($G^{\star}$).

Since we seek a scale for each data point, we work with a multiscale Gaussian similarity kernel, defined as in section~\ref{section:kernels}:
\begin{equation}\label{eq:ms-kernel}
K_{ij} = \exp\left({\frac{-\|\bm{x}_i-\bm{x}_j\|^2}{\sigma_i\sigma_j}}\right).
\end{equation}
The kernel value $K_{ij}$ is therefore symmetric and equivalent to that of a traditional Gaussian kernel (equation~\ref{eq:gauss-kernel}), except using the geometric mean of $\sigma_i$ and $\sigma_j$ as its scale. Notice, in particular, how the scales and the kernel value are coupled: setting the scale incorrectly could make distant points $\bm{x}_i$ and $\bm{x}_j$ appear close in similarity. 

Given a set of individual point scales $\sigma_i$  (sometimes collected into the vector $\bm{\sigma} \in \mathbb{R}^N$), we define a second, weighted graph $\mathcal{G}=(V,\mathcal{E},W)$ as the complete graph on all pairs of data points in $\mathcal{X}$. Its weighted adjacency matrix, $W$, has entries $w_{ij} = K_{ij}$.

While the unweighted graph will be related to nearest neighbors and computational geometry, the weighted graph will be related to spectral methods on manifold inference. In particular, we expect the Laplacian of $\mathcal{G}$ to approximate the Laplace-Beltrami operator on $\mathcal{M}$, subject to the number of data points and their sampling.

The algorithm is initialized by computing a coarse estimate of $G$. As described later in section~\ref{section:connect-criterion}, this is achieved by exploiting the geometry of medial balls between pairs of points to produce a \emph{Gabriel graph} \citep{GG0,GGproperties}. A Gabriel graph is that in which there is an edge between two points $\bm{x}_i$ and $\bm{x}_j$ if and only if they are the only two closest points to the midpoint of the line segment joining them. The main advantages of using a Gabriel graph as a starting point are: (\textit{i}) it is scale invariant, so a prespecified $\varepsilon$-neighborhood (equation~\ref{eq:eps-nbrhood}) is not required; (\textit{ii}) there is no global constant $k$ (it can vary); and (\textit{iii}) neighbors are not limited to the closest neighbors in ambient space. Thus, it allows for connections to ``jump across'' sampling gaps while keeping the data graph sparse.

However, as described in section~\ref{section:reach}, obtaining a good inference of $\mathcal{M}$ amounts to finding reasonable estimates of its reach and local feature size. For that to occur, no edge segment $\ell_{ij}$ between two points $\bm{x}_i$ and $\bm{x}_j$ should cross a medial axis of $\mathcal{M}$. As the examples that follow will show, there are several cases in which the Gabriel graph will violate this. Therefore, additional steps are necessary to refine it. The Gabriel graph provides a necessary condition (all the correct connections are present, but possibly others as well); our refinement moves toward sufficiency. 

In order to estimate $\mathcal{G}$---the weighted counterpart of $G$---we will use the weights that are obtained by applying a continuous kernel (equation~\ref{eq:ms-kernel}) over the points in $\mathcal{X}$. Such a kernel requires scales, or bandwidths, $\bm{\sigma}$ that must be estimated from $G$. 
These will be obtained from an optimization procedure that finds the smallest such scales ensuring that all discrete edges have a minimum kernel value as weight. At this point, a weighted graph $\mathcal{G}$ can be obtained from $\bm{\sigma}$.

\begin{figure}[ht!]
  \centering
  \includegraphics[width=0.75\textwidth]{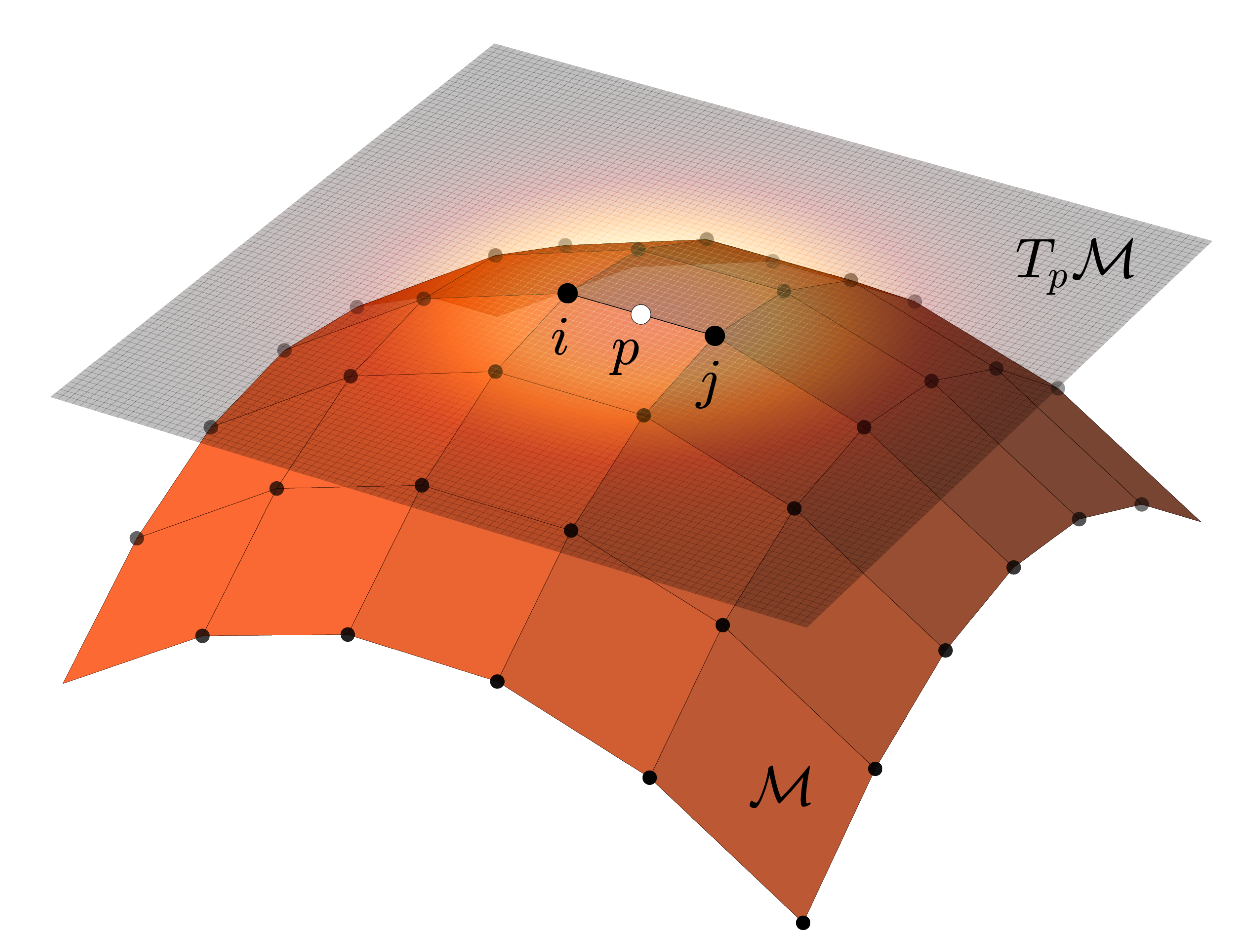}
  \caption{Relating the discrete neighborhood graph to manifold geometry. Nearby sampled points ($i$ and $j$) on a patch of manifold $\mathcal{M}$ lie in (or near) the tangent plane $T_p\mathcal{M}$ to the midpoint ($p$). Line segments (edges) between neighboring points lift, via the exponential map, to geodesics in $\mathcal{M}$. The continuous kernel extends this discrete relationship to the full tangent plane. The values of the kernel centered at $p$ are shown as shading, extending in every direction in $T_p\mathcal{M}$. Our algorithm shall enforce this relationship, i.e., the consistency between discrete edges and large kernel values.
  }
  \label{fig:tangent-plane}
\end{figure}

It is now helpful to articulate the geometry more carefully; Figure~\ref{fig:tangent-plane} depicts how the discrete connectivity relates to the manifold geometry. In particular,
for a real dataset, the few closest points surrounding $\bm{x}_i$ are the best candidates for ``nearest'' neighbors---this is all that can be asserted locally. Let $\bm{p}_i$ and $\bm{p}_j$ be the projections of two neighbors $\bm{x}_i$ and $\bm{x}_j$ onto $\mathcal{M}$, respectively. Then, any point along the geodesic between $\bm{p}_i$ and $\bm{p}_j$ should be closer to no sampled point other than $\bm{x}_i$ or $\bm{x}_j$. By further assuming $\bm{x}_i \in \mathcal{M}, \forall i$ or at least that $\|\bm{x}_i - \mathcal{M}\|_{\mathbb{R}^n} < \varepsilon, \forall i$ and small $\varepsilon$, then $\|\bm{x}_i - \bm{x}_j\|_{\mathbb{R}^n}$ approximates the geodesic when the curvature between $\bm{p}_i$ and $\bm{p}_j$ is small. Equivalently, the line segment $\ell_{ij}$ between $\bm{x}_i$ and $\bm{x}_j$ lies on the tangent space $T_p\mathcal{M}$, where $p$ is the midpoint between $\bm{p}_i$ and $\bm{p}_j$; see Figure~\ref{fig:tangent-plane}. The existence of a geodesic follows from identifying the tangent plane that includes the points with the exponential map of the manifold around them. 

Such an ``edge-centric'' approach connects differential geometry to the underlying graph. This is illustrated in Figure~\ref{fig:tangent-plane}, where the kernel values are shown as shading in the tangent plane. Notice how $\bm{x}_i$ and its neighbor $\bm{x}_j$ both fall under the bright kernel values; i.e., they are very similar (in this measure) to each other. Stated in geometric terms, we assume that the neighbors lie within the {\em injectivity radius} around $p$. In fact, we will show (Figure~\ref{fig:ms-covering}) that the value of a multiscale kernel between two data points is equivalent to that of a rescaled, single-scale kernel centered at the midpoint between those two points. 

Now, the optimized scales can be used to evaluate the current approximation and identify the edges in $G$ that are ``too expensive,'' i.e., are likely to violate the local feature size. We proceed by computing successive refinements of both $G$ and $\bm{\sigma}$, in an iterative manner, until no further change is observed. We then return the final version of the discrete and weighted graphs (denoted by $G^{\star}$ and $\mathcal{G}^{\star}$, respectively).

One can view the computation of $\mathcal{G}$ as a relaxation of the discrete connectivity in $G$. In fact, as we shall see in section~\ref{section:stat-pruning}, a relaxation statistic, $\delta'_i$, will be used to prune discrete edges that produce a poor approximation. More specifically, when a node $i$ with degree, $\mathrm{deg}(i)$, in $G$ has $\delta'_i$ close to 1, it means $i$ has retained approximately the same degree in $\mathcal{G}$, only continuously spread as a Gaussian around it. 

\begin{figure}[ht!]
  \centering
  \includegraphics[height=0.9\textheight]{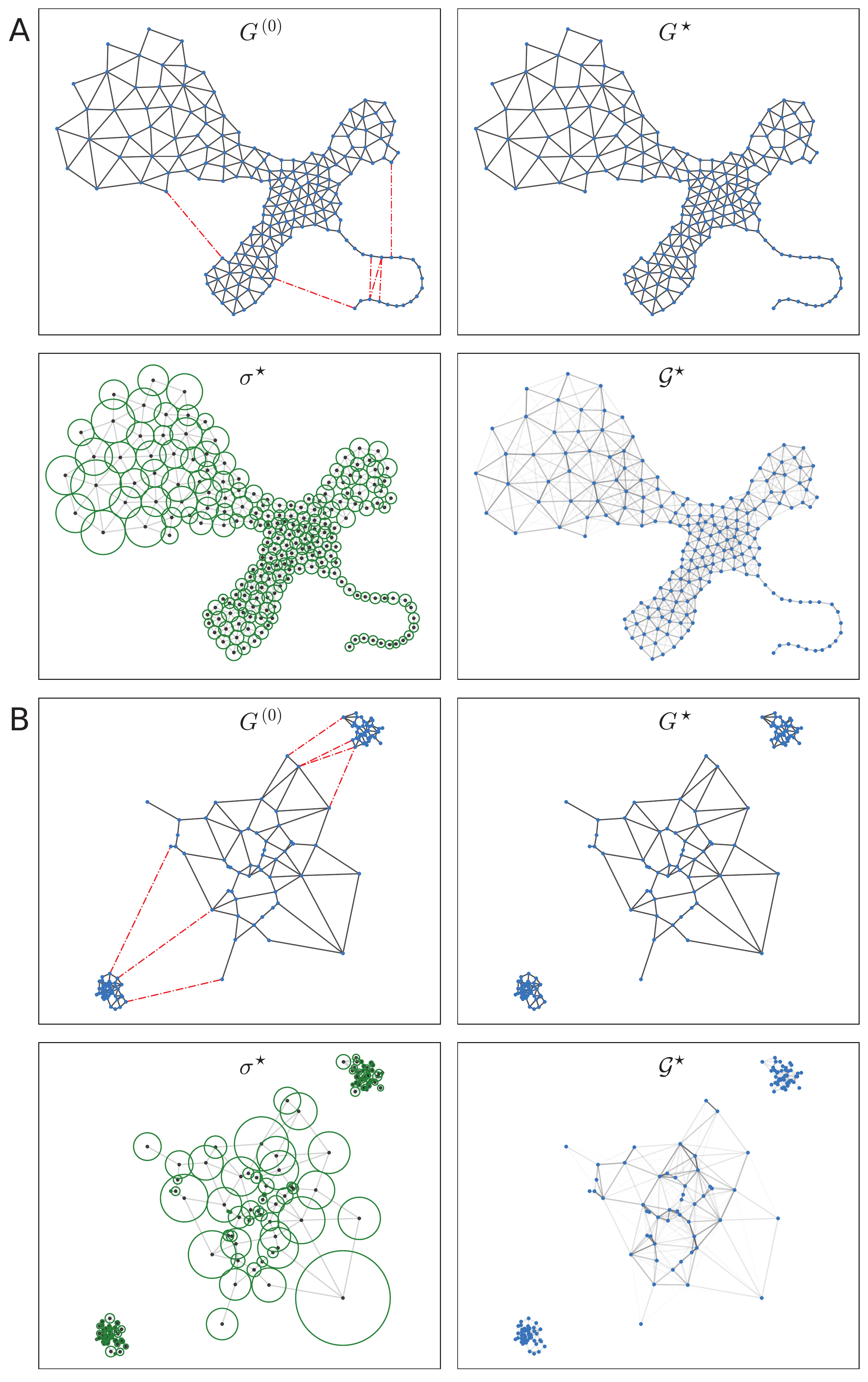}
  \caption{\emph{See next page.}
  }
  \label{fig:algo-examples}
\end{figure}

Each of the steps above are listed in Algorithm~\ref{algo1} and will be described in detail. We begin with the discrete connectivity rule (Gabriel graph); then the scale optimization is developed, followed by the edge-pruning step. Figure~\ref{fig:algo-examples} illustrates the results of our algorithm on datasets for which the Gabriel graph alone cannot infer a good approximation of the manifold connectivity.

\addtocounter{figure}{-1}

\begin{figure}[t!]
  \caption{
Steps of Algorithm~\ref{algo1} on toy datasets.
  (\textbf{A}) Dataset with several challenges: non-uniform density, non-uniform dimension, and high curvature. After pruning 6 edges (dashed red lines) from the original Gabriel graph, $G^{(0)}$, the algorithm converges, inferring reasonable discrete neighborhoods ($G^{\star}$); the optimal scales $\bm{\sigma}^{\star}$ produce a weighted graph $\mathcal{G}^{\star}$ whose connectivity closely approximates that of $G^{\star}$.
  (\textbf{B}) Dataset with three Gaussian clusters of non-uniform density. The Gabriel graph approximation, $G^{(0)}$, naively connects all clusters using multiple edges. After convergence, the clusters become disconnected in $G^{\star}$, and its weighted version follows this by assigning negligible weights (due to $\bm{\sigma}^{\star}$) between points in different clusters.
  }
\end{figure}

\begin{algorithm}
\caption{Iterated Adaptive Neighborhoods kernel}\label{algo1} \begin{algorithmic}[1] \Procedure{IANkernel}{$D$}\Comment{Input: distance matrix, $D$}
\State $G^{(0)}\gets$ \Call{GabrielGraph}{$D$}\Comment{Compute initial $G$ (sec.~\ref{section:connect-criterion})}
\Repeat \ Iteration
   \State $\bm{\sigma}^{(t)},\gets$ \Call{OptimizeScales}{$G^{(t)},D$}  \Comment{Update scales $\bm{\sigma}$ (sec.~\ref{section:optimization})}
   \State $\mathcal{G}^{(t)}\gets$ \Call{MultiScaleKernel}{$D,\bm{\sigma}^{(t)}$} \Comment{Weighted graph (eq.~\ref{eq:ms-kernel})}
   \State $\delta', C\gets$ \Call{ComputeVolumeRatios}{$G^{(t)},\bm{\sigma}^{(t)}$} \Comment{Statistic $\delta'$ (sec.~\ref{section:delta-stat})}
   \State $G^{(t+1)}\gets$ \Call{Sparsify}{$G^{(t)},\delta'$} \Comment{Update $G$ (sec.~\ref{section:Ctuning})}
\Until{no further change in $G$}
   \State \textbf{return} $G^{\star}, \mathcal{G}^{\star}, \bm{\sigma}^{\star}$\Comment{Output: final graphs and optimal scales}
\EndProcedure 
\end{algorithmic}
\end{algorithm}


\subsection{Neighbors in a Gabriel graph}\label{section:connect-criterion}

\begin{figure}[ht!]
  \centering
  \includegraphics[width=1\textwidth]{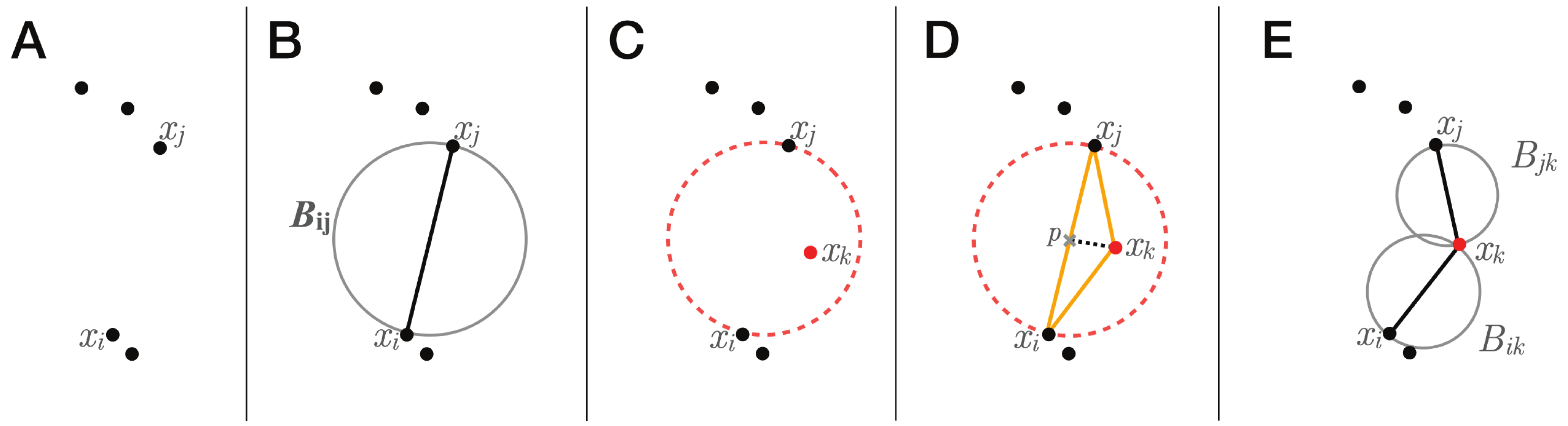}
  \caption{Connecting ``nearest neighbors.'' (\textbf{A}) A set of data points in space. (\textbf{B}) An edge can be formed between $\bm{x}_i$ and $\bm{x}_j$ because there is no other point in the interior of the ball $\mathcal{B}_{ij}$ centered halfway between $\bm{x}_i$ and $\bm{x}_j$.
(\textbf{C}) Here, because of the presence of a third point $\bm{x}_j$ inside $\mathcal{B}_{ij}$, $\bm{x}_i$ and $\bm{x}_j$ cannot be neighbors.
(\textbf{D}) Even in the absence of the original data point coordinates, i.e., given only the distances between all pairs of points, Apollonius's formula can be used to determine the length of the segment $p$--$\bm{x}_k$, where $p$ is the center of $\mathcal{B}_{ij}$. Namely, $p$--$\bm{x}_k$ is a median of the depicted triangle. Here, because the length of the median is less than the radius of $\mathcal{B}_{ij}$, $\bm{x}_i$ and $\bm{x}_j$ cannot be neighbors.
(\textbf{E}) Edges are drawn connecting points $\bm{x}_i$ to $\bm{x}_k$ and $\bm{x}_k$ to $\bm{x}_j$ because both $\mathcal{B}_{ik}$ and $\mathcal{B}_{jk}$ are empty except for those pairs of points, respectively.
  }
  \label{fig:pruning}
\end{figure}

We begin by defining a set of putative neighboring points of $\bm{x}_i$ (denoted by $\mathcal{N}(i)$), which uses the connectivity rule found in a \emph{Gabriel graph} \citep{GG0,GGproperties}. It directly incorporates the observation that closest neighbors should have no points ``between'' them.
\begin{remark}
Two points, $\bm{x}_i$ and $\bm{x}_j$, are \emph{Gabriel-nearest neighbors} to each other if and only if they both touch the same closed ball, $\mathcal{B}_{ij}$, that is empty except for $\bm{x}_i$ and $\bm{x}_j$.
\end{remark} 
\noindent Note that $\mathcal{B}_{ij}$ is therefore a \emph{medial ball}, i.e., a ball whose center point is a medial axis (with respect to the set of sampled points). Thus, this connectivity criterion can be restated as creating an edge for all those medial balls, and only those, touching exclusively two points (to be clear, if a third point touches $\mathcal{B}_{ij}$ no edge shall be formed between $\bm{x}_i$ and $\bm{x}_j$). Hence, to each edge $e_{ij}$ is associated a medial ball $\mathcal{B}_{ij}$ centered and the midpoint between $\bm{x}_i$ and $\bm{x}_j$ with radius $\|\bm{x}_i-\bm{x}_j\|/2$ (see Figure~\ref{fig:pruning}). This is furthermore equivalent to the following alternative definitions:
\begin{remark}
Points $\bm{x}_i$ and $\bm{x}_j$ are Gabriel-nearest neighbors if and only if any point along the line segment $\ell_{ij}=\overline{\bm{x}_i\bm{x}_j}$ in $\mathbb{R}^n$ has either $\bm{x}_i$ or $\bm{x}_j$ (or both) as its only closest point(s).
\end{remark}
\begin{remark}
In terms of the Voronoi diagram \citep{fortune1995voronoi} of $\mathcal{X}$ (with the cell around $\bm{x}_i$ denoted by $V_i$), $\bm{x}_i$ and $\bm{x}_j$ are neighbors when $\ell_{ij}$ crosses a single Voronoi hyperplane $H_{ij}$ (namely that between the cells $V_i$ and $V_j$) and the midpoint between $\bm{x}_i$ and $\bm{x}_j$ is in $H_{ij}$. 
\end{remark}

As a concrete example (refer to Figure~\ref{fig:pruning}), consider two points $\bm{x}_i$ and $\bm{x}_j$ at a distance $r_{ij}$ from each other, with midpoint $p$. Assume the region in the manifold between them is uniformly sampled. Now consider the ball centered at $p$ with radius $r_{ij}/2$, therefore touching $\bm{x}_i$ and $\bm{x}_j$. If there are no points in its interior, we say $\bm{x}_i$ and $\bm{x}_j$ are nearest neighbors. Conversely, if it contains other points in its interior, under our assumption of uniform density this means that there is at least one other point $\bm{x}_k$ ``between'' $\bm{x}_i$ and $\bm{x}_j$. So we say that $\bm{x}_i$ and $\bm{x}_j$ are \emph{not} nearest neighbors, in the sense that connecting $\bm{x}_i$ and $\bm{x}_j$ directly would be ``crossing over'' $\bm{x}_k$; this implies that an edge $e_{ij}$ in the resulting graph would be a poor approximation to a geodesic in $\mathcal{M}$ (i.e., if $\mathcal{M}$ is ``locally uniformly sampled,'' the segment $\ell_{ij}$ would be passing outside of $\mathcal{M}$). Note that, even when the input to the algorithm is solely a distance matrix (i.e., with no position information), this connectivity criterion may still be evaluated by considering the triangle $\bm{x}_i$--$\bm{x}_j$--$\bm{x}_k$ and using Apollonius's theorem to compute the length of the median from $\bm{x}_k$ to $p$ (Figure~\ref{fig:pruning}-D).

The Gabriel graph is a subgraph of the Delaunay graph \citep{gg-delaunay}, and enjoys a number of key properties \citep{GGproperties}. We emphasize: (\textit{i}) they are scale invariant, i.e., there is no pre-specified threshold on the diameter of medial balls that can form connections; (\textit{ii}) the guarantee that Gabriel graphs connect points to their true nearest neighbors when $\mathcal{M}$ is uniformly sampled as a grid (shown in Figure~\ref{fig:dim-degs}); and  (\textit{iii}), Gabriel graphs provide a locally-adapted neighborhood size $k_i$, for each point $\bm{x}_i$, based on the local geometry. Crucially, they do not require an initial guess of the number of neighbors, of the intrinsic dimensionality, or of a maximum neighborhood radius. 

Nevertheless, the neighborhoods given by the Gabriel graph are not sufficent. We now expand on a few of their properties---these will be useful in motivating the rest of the algorithm.

\subsubsection{Closing triangles}\label{section:triangles}

\begin{figure}[ht!]
  \centering
  \includegraphics[width=1\textwidth]{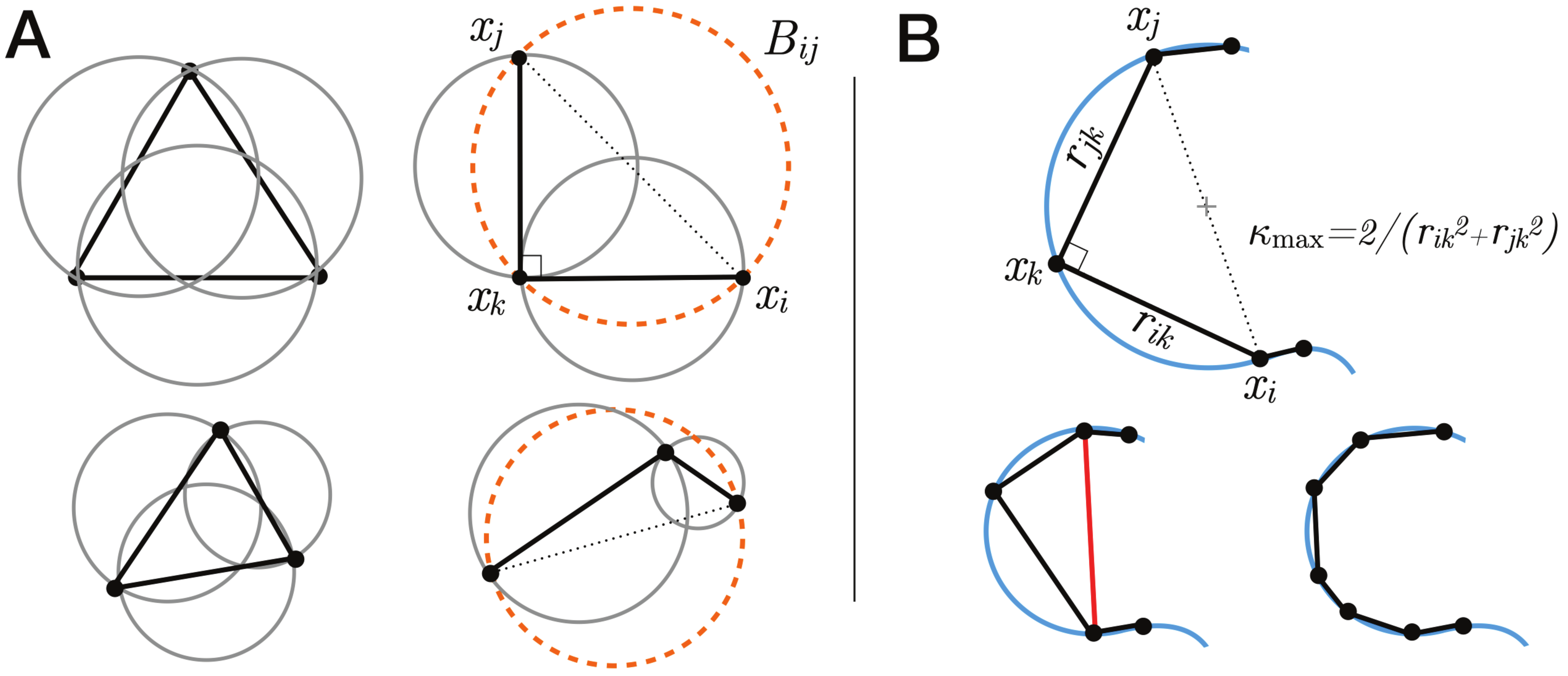}
  \caption{Implications of the connectivity rule in a Gabriel graph.
  (\textbf{A}) Closing triangles from edges: three points will be mutual neighbors if and only if they form an acute triangle (left). If the angle between $\bm{x}_i$ and $\bm{x}_j$ at $\bm{x}_k$ is at least $\pi/2$, all three points will lie in $\mathcal{B}_{ij}$, so no edge is created (right).
  (\textbf{B}) The maximum principal curvature in $\mathcal{M}$ (shown in blue) that can be reasonably approximated by the resulting graph geodesic (path) is constrained by the sampling interval. The limiting case occurs when three points form a right triangle (top), cf. equation~\ref{eq:curv1}). When sampling is too sparse (bottom left), a triangle may be formed, in this case preventing the graph from adequately capturing the manifold's geometry. As sampling frequency increases (bottom right), higher curvatures can be better approximated.
  }
  \label{fig:triangles-curvature}
\end{figure}

Here we show that the edges created using the above connectivity rule can only form acute triangles in $\mathbb{R}^n$. Let three points $\bm{x}_i$, $\bm{x}_j$, $\bm{x}_k$ be such that $\bm{x}_i$ and $\bm{x}_k$ are connected, as well as $\bm{x}_j$ and $\bm{x}_k$. The rule says, $\bm{x}_i$ and $\bm{x}_j$ shall be connected only if $\bm{x}_k$ is outside the closed ball $\mathcal{B}_{ij}$ of radius $R = r_{ij}/2$ centered half-way between $\bm{x}_i$ and $\bm{x}_j$ (where $r_{ij}$ stands for the Euclidean distance between $\bm{x}_i$ and $\bm{x}_j$). Using Apollonius's formula for the squared distance $m^2$ between $\bm{x}_k$ and the midpoint between $\bm{x}_i$ and $\bm{x}_j$, we obtain
\begin{equation}
    m^2 = \frac{1}{4}(2r_{ik}^2 + 2r_{jk}^2 - r_{ij}^2).
\end{equation}
Then, $\bm{x}_k$ is in $\mathcal{B}_{ij}$ if and only if $m^2 \leq R^2$, so
\begin{equation}\label{eq:pythagoras}
\begin{aligned}
    \frac{1}{4}(2r_{ik}^2 + 2r_{jk}^2 - r_{ij}^2) &\leq R^2 = (\frac{r_{ij}}{2})^2 \\
    r_{ik}^2 + r_{jk}^2 &\leq r_{ij}^2.
\end{aligned}
\end{equation}
Notice that equality will hold when $\bm{x}_i$--$\bm{x}_j$--$\bm{x}_k$ is a right triangle. Therefore:
\begin{remark}
A triangle will be formed by edges in a Gabriel graph only when it is acute (see Figure~\ref{fig:triangles-curvature}-A).
\end{remark}

\subsubsection{Maximum curvature}\label{section:max-curvature}

The above result leads to a bound on the maximum principal curvature that is allowed locally on $\mathcal{M}$ such that the Gabriel graph correctly approximates it (i.e., without closing a triangle).
Assume $\bm{x}_i$, $\bm{x}_j$, and $\bm{x}_k$ are points in a smooth manifold $\mathcal{M}$ as in Figure~\ref{fig:triangles-curvature}-B, up to the level that the sampling defines. If we assume that the curvature, $\kappa$, is locally constant, then the geodesic from $\bm{x}_i$ to $\bm{x}_j$ passing through $\bm{x}_k$ is an arc of a circle $\mathcal{C}$. Therefore, the segments $\ell_{ik}$ and $\ell_{kj}$ approximate geodesics on $\mathcal{M}$, but not $\ell_{ij}$ (which would cause ``folding''). Hence, values of curvature that can be correctly inferred are those that do not create an edge between $\bm{x}_i$ and $\bm{x}_j$ (i.e., those for which the ball $\mathcal{B}_{ij}$ is non-empty). In this case, from equation~\ref{eq:pythagoras}, the maximum such curvature, $\kappa^{\mathrm{max}}$, occurs when $\bm{x}_i$, $\bm{x}_k$, and $\bm{x}_j$ form a right triangle in space (as any larger value would cause this triangle to be acute, connecting $\bm{x}_i$ to $\bm{x}_j$). Then, from Thales's theorem, the diameter $D$ of $\mathcal{C}$ would equal that of the hypotenuse $\ell_{ij}$, so
\begin{equation}\label{eq:curv1}
    \kappa^{\mathrm{max}} = \frac{1}{D/2} = \frac{2}{D} = \frac{2}{\sqrt{r_{ik}^2 + r_{jk}^2}}.
\end{equation}

A special case to consider is when $\mathcal{M}$ is uniformly sampled with constant interval $T$ over arc length.  Then, the arc length $s$ between $i$ and $j$ is $2T$; but, since $r_{ij} = D$, $s$ covers half the circle and we have $2T = \pi D/2$. Equation~\ref{eq:curv1} then becomes
\begin{equation}\label{eq:curv2}
    \kappa^{\mathrm{max}}(T) = \frac{\pi}{2T}.
\end{equation}
\begin{remark}
Equations~\ref{eq:curv1} and \ref{eq:curv2} define the maximum geodesic curvature in $\mathcal{M}$ that can be adequately inferred from a Gabriel graph. As a consequence, the reach is lower-bounded by $1/\kappa^{\mathrm{max}}$.
\end{remark}

\subsubsection{Degree distribution in Gabriel graphs}\label{section:GG-degrees}

We now study the above connectivity rule starting with flat, uniformly sampled manifolds (i.e., ``regular grids'') to illustrate how Gabriel graphs naturally adapt to both their geometry and dimensionality. As shown in Figure~\ref{fig:dim-degs}-A, in such ideal cases the degree of an interior node in the Gabriel graph agrees with the true number of (literal) nearest neighbors, i.e.: 2 for collinear points, 4 for a square grid, and 6 for a triangular grid.

\begin{figure}[th!]
  \centering
  \includegraphics[width=.95\textwidth]{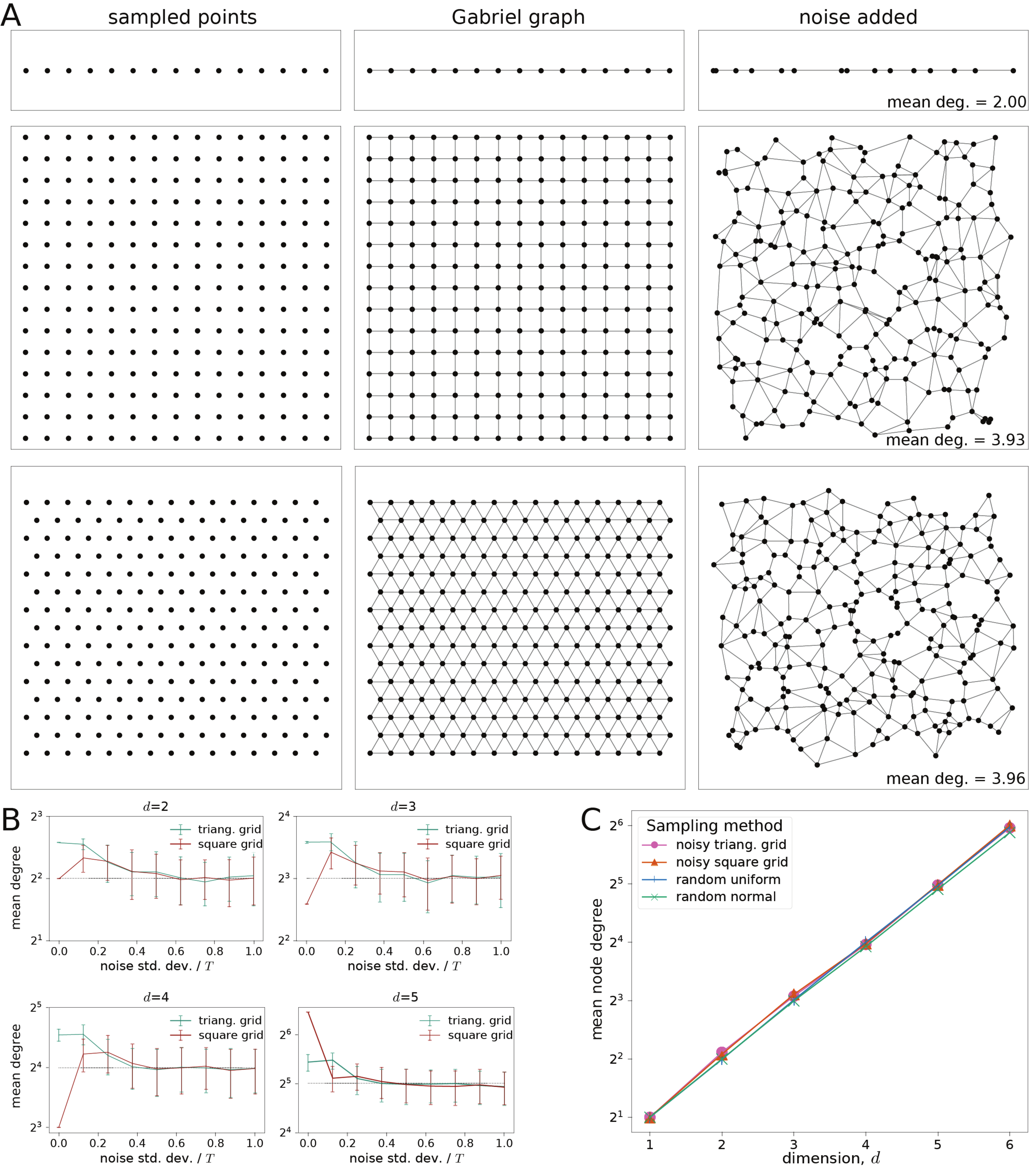}
  \caption{Regularity of node degree distribution in Gabriel graphs with random sampling.
  (\textbf{A}) Node degree in graphs computed from regular grids (constant sampling interval, $T$) and their jittered versions (Gaussian noise with std. dev. $0.5T$). \textit{Top:} A sequence of collinear points (left) produces a one-dimensional grid (center). Addition of noise (right) does not change the mean degree (constant 2 for interior points). 
  \textit{Middle:} A square grid (left) results in a quadrilateral mesh with constant degree 4 in its interior. Although addition of noise considerably scrambles the points, the mean degree is roughly unchanged.
  \textit{Bottom:} Points arranged as a triangular grid (left) result in a triangular mesh where every interior node has degree 6. Its noisy version looks similar to a noisy square grid, with mean degree also approximating 4. \emph{(Cont. next page.)}}
  \label{fig:dim-degs}
\end{figure}
\addtocounter{figure}{-1}

\begin{figure}[t!]
  \caption{
  \emph{(Cont. from previous page.)}
  (\textbf{B}) Degree distribution for interior points of $d$-dimensional triangular and square grids after addition of Gaussian noise. Moderate amounts of noise are sufficient to make the mean degree become approximately $2^d$. Error bars indicate standard deviation; dotted lines show constant $2^n$ values for reference.
  (\textbf{C}) Mean degree of $d$-dimensional manifolds sampled using different strategies: uniformly at random, normally at random, and as jittered versions of regular triangular and square grids (as in (\textbf{A}), added Gaussian noise with std. dev. $0.5T$). Remarkably, mean degree grows approximately as $2^n$ regardless of the sampling strategy. 
  }
\end{figure}

\begin{figure}[!ht]
  \centering
  \includegraphics[width=1\textwidth]{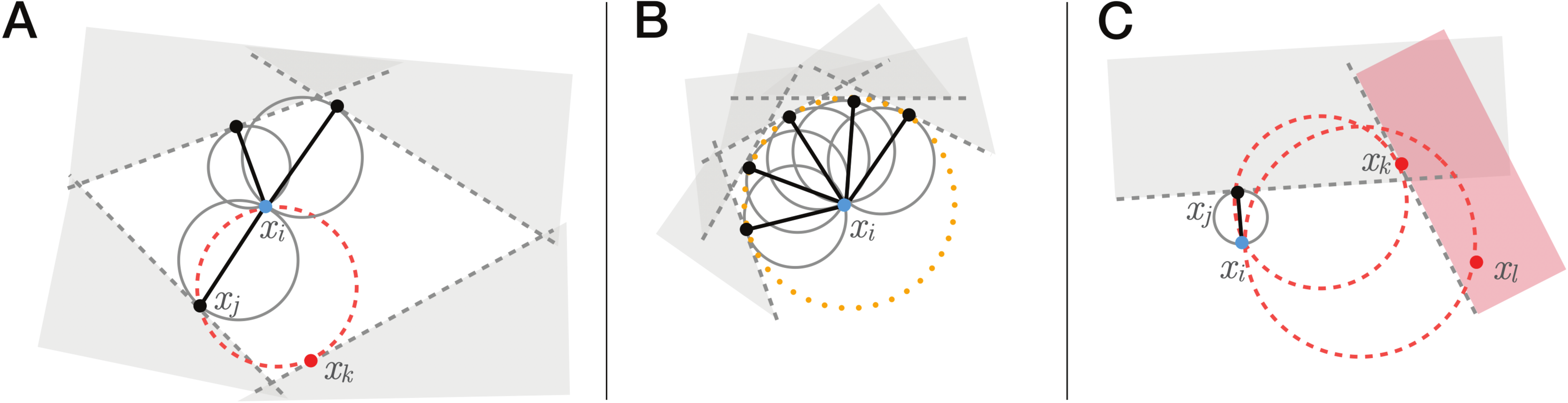}
  \caption{
  (\textbf{A}) A central point $\bm{x}_i$ (in blue) and its neighbors (in black). Every neighbor $\bm{x}_j$ of $\bm{x}_i$ will ``occlude'' 
the entire area behind a hyperplane tangent to $\mathcal{B}_{ij}$ at $\bm{x}_j$ (dashed lines). That is, no point inside the occluded areas (shaded region) can form a connection with $\bm{x}_i$. Here, the dashed ball does not form a connection  between $\bm{x}_i$ and $\bm{x}_k$ because $\bm{x}_j$ lies exactly on its boundary; despite this, $\bm{x}_k$ still contributes with an occluding hyperplane, preventing farther points from connecting to $\bm{x}_i$.
(\textbf{B}) In principle, there is no limit to the number of neighbors a point in ambient space $\mathbb{R}^n$ may have (when $n \geq 2$); e.g., any number of points lying exactly on a hypersphere around $\bm{x}_i$ (dotted curve, in orange) will not occlude one another. Sets of nodes with connectivity such as this are termed ``wheels'' in graph theory, and the more points they contain, the less likely they are to occur in real datasets. In this example, any appreciable variability in the distance from $\bm{x}_i$ to its neighbors would cause one (or several) of them to become occluded. 
(\textbf{C}) Points inside occluded areas can also contribute with additional occluding hyperplanes. Here, although $\bm{x}_k$ lies inside the region occluded by $\bm{x}_j$ (and therefore cannot form a connection with $\bm{x}_i$), it produces further occlusion behind a hyperplane of its own (region shaded in red). So $\bm{x}_l$ cannot connect to $\bm{x}_i$, either, due to the presence of $\bm{x}_k$ (even though it is not occluded by $\bm{x}_j$).
  }
  \label{fig:occlusion}
\end{figure}

\begin{figure}[h!]
  \centering
  \includegraphics[width=1\textwidth]{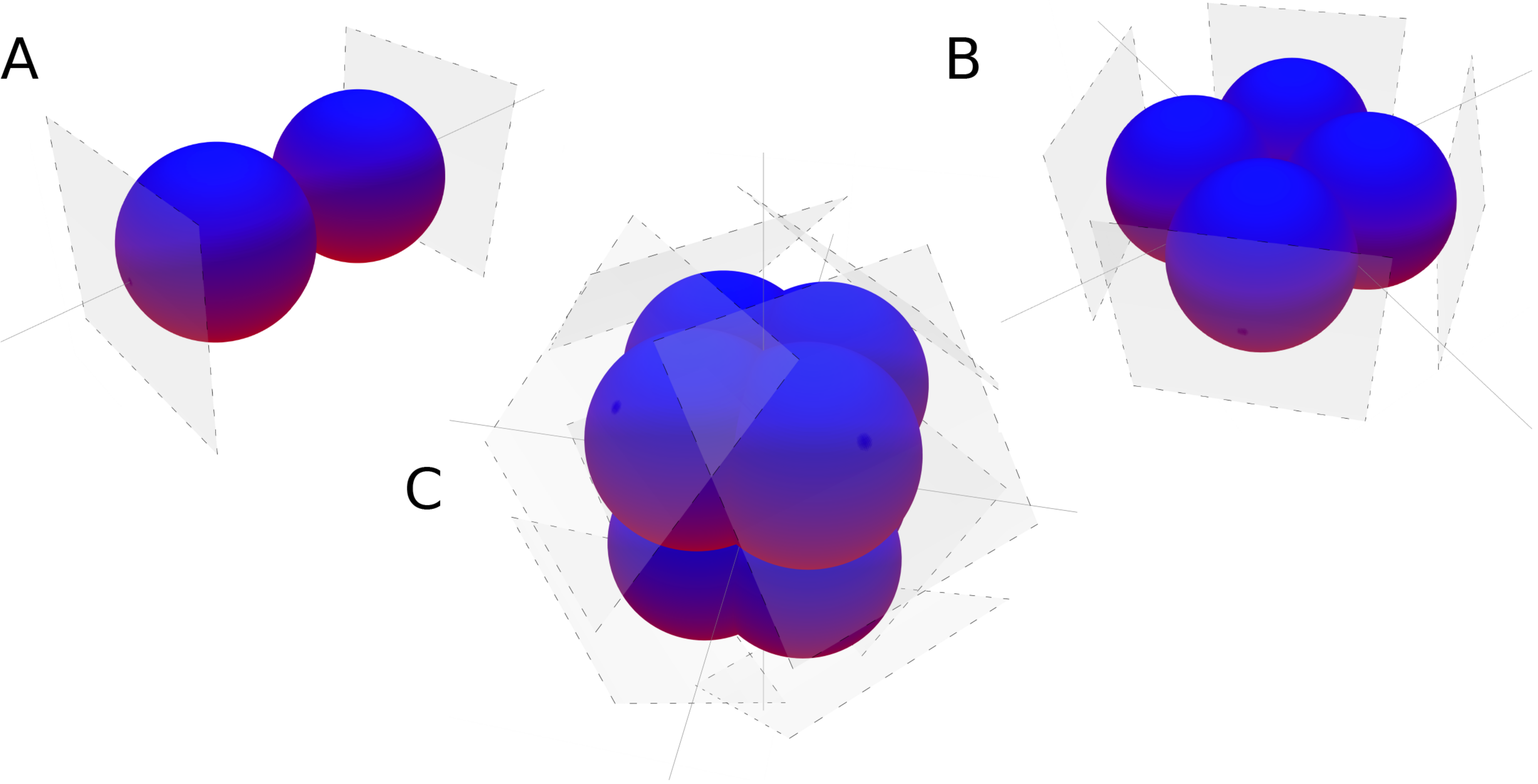}
  \caption{Occlusion hyperplanes (shown in gray) due to neighbors in dimensions 1, 2, and 3 (\textbf{A}--\textbf{C}, respectively); compare with Figure~\ref{fig:occlusion}. Every additional dimension adds a new coordinate axis along which the previous constraints are duplicated, roughly doubling the average amount of directions available from which neighbors can connect.  Once $2^d$ Gabriel balls are ``attached'' to $\bm{x}_i$, the remaining space is greatly reduced, and so is the probability of drawing a sample point from inside the region enclosed by the hyperplanes.}
  \label{fig:orthoplexes}
\end{figure}

\begin{figure}[!ht]
  \centering
  \includegraphics[width=1\textwidth]{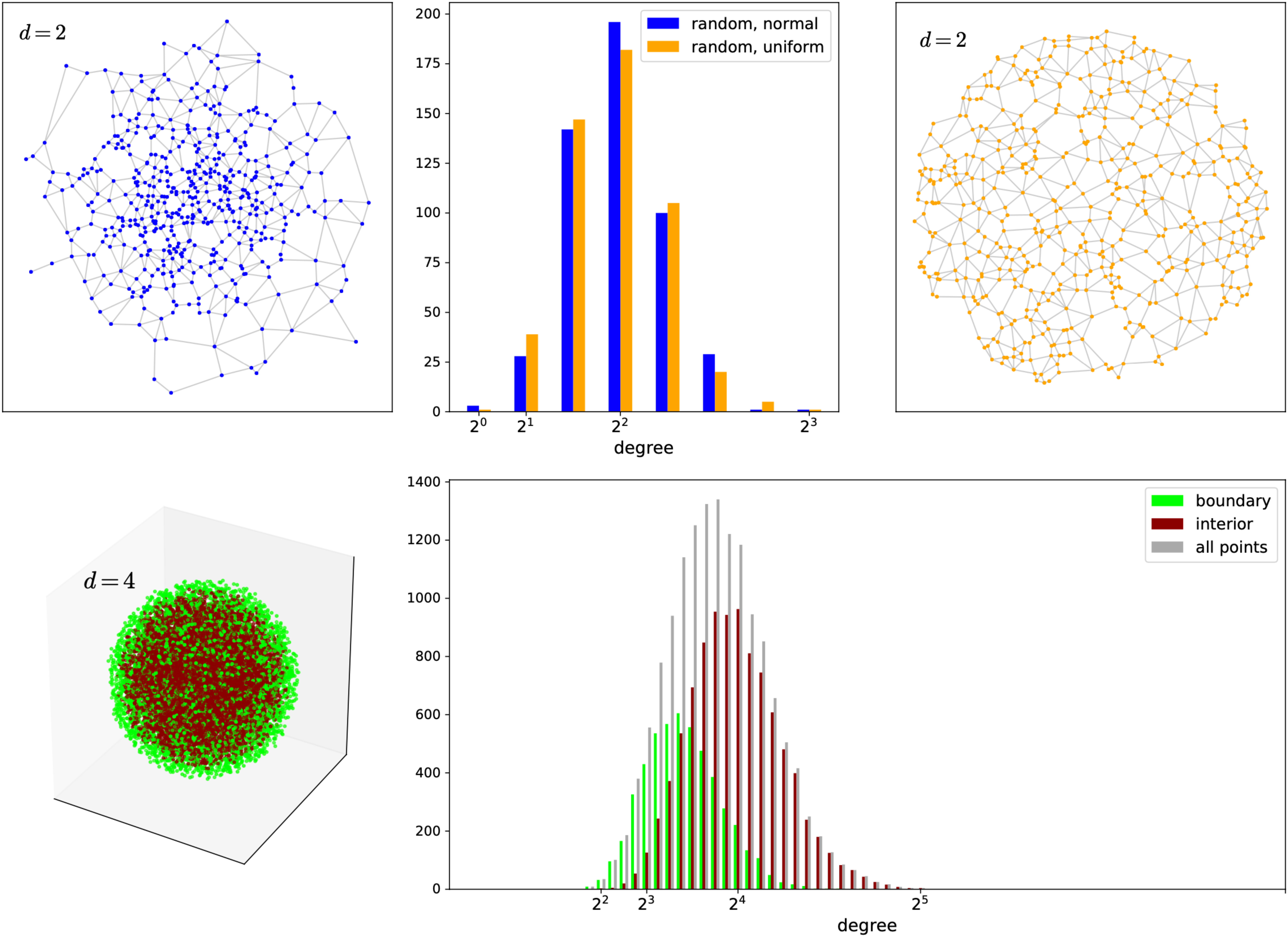}
  \caption{
  Distribution of node degree in the Gabriel graph of datasets with different sampling strategies and dimensionalities.
  \emph{Top:} Points sampled normally (blue) or uniformly (orange) at random from a two-dimensional ball result in similar degree distributions centered at $2^2$.
  \emph{Bottom:} In higher dimensions, interior points continue to follow this pattern. On the left, a 4-dimensional unit ball sampled uniformly at random is shown projected onto $\mathbb{R}^3$, with boundary points labeled as those with vector norm $> 0.9$ (edges omitted for clarity). It produces a Gabriel graph where interior points have degree distribution centered at $\sim$2$^4$, and the mean degree of boundary points is close to $2^3$.
  }
  \label{fig:deg-histograms}
\end{figure}

Node degree appears to grow with dimension as $2^d$, except for the triangular grid (which, in some sense, looks too ``non-generic''). Adding noise (Gaussian, with standard deviation equal to half the spacing between neighboring points) supports this conjecture, as the degree then approaches $2^d$ regardless of the original grid structure. This holds in higher dimensions as well, for both normal and uniform sampling at random (Figures~\ref{fig:dim-degs}-B,C and \ref{fig:deg-histograms}).
\begin{remark}
The expected number of neighbors in a Gabriel graph approximately follows a distribution centered at $2^d$ (where $d$ is the intrinsic dimension of the data) for a variety of sampling strategies (Figure~\ref{fig:dim-degs}-C).
\end{remark}
\noindent Importantly, because Gabriel graphs are inherently scale invariant, this degree distribution is largely independent of sampling density.

How to explain such remarkable regularity despite the randomness of sampling? A complementary geometric view of the Gabriel graph connectivity rule is illuminating: each edge between data points implies an ``occluding hyperplane'' that blocks other points from becoming neighbors (see Figure~\ref{fig:occlusion}). For example, when $d$ = 1, two points necessarily occlude any additional connections, and every non-boundary point must have 2 neighbors. Now, using the diagrams in Figure~\ref{fig:orthoplexes} as reference, we find that, when $d$ = 2, on average $\sim$4 points are sufficient to occlude a point $\bm{x}_i$ from all sides. For $d$ = 3 this number is doubled again, and the expected number of neighbors becomes $\sim$8, revealing the trend.  Every additional dimension adds a new coordinate axis along which the previous constraints are duplicated, roughly doubling the average number of directions available from which neighbors can connect. Once $2^d$ balls are ``attached'' to $\bm{x}_i$, the remaining space is greatly reduced, and so is the probability of drawing a sample point from inside the region $\mathcal{H}$ enclosed by the hyperplanes.

When the neighbors are regularly spread around $\bm{x}_i$, by construction this region $\mathcal{H}$ is equivalent to a $d$-dimensional orthoplex\footnote{An orthoplex is a line segment in 1-D, a square in 2-D, a regular octahedron in 3-D, a 16-cell in 4-D, etc.} (or cross-polytope). A $d$-orthoplex has $2^d$ facets (or $(d$-$1)$-faces), and is one of the three finite, regular, convex polytopes that exist in dimension higher than 4 (the other two being hypercubes and simplices). Naturally, when sampling is not uniform, we should find irregular orthoplexes instead.

While this geometric construction supports our empirical results, and implies they should hold in higher dimensions, it also suggests the following:
\begin{remark}
Our experiments on the growth in dimension of randomly sampled points agree with a model in which Gabriel neighbors lie approximately in the facets of an orthoplex. 
\end{remark}
\noindent We shall later use the additional observation that the dual polytope (a $d$-hypercube) of an orthoplex is obtained by placing a vertex (i.e., a neighbor) in each of its $2^d$ facets.

The Gabriel graph enjoys many attractive properties, and provides the starting point for our algorithm.  The above arguments show how the space is largely filled by ``Gabriel balls'' within the manifold, but such balls may also fill space across holes and bottlenecks; curvature must be dealt with. Examples were given in Figure~\ref{fig:algo-examples}, where we showed that Gabriel connections can arise incorrectly and must be removed. To do so, one must ``look'' in every direction (of the tangent plane), and  past immediate neighbors. For this, we now develop the weighted graph counterpart to the Gabriel graph, exploiting the kernel to extend local information globally. This begins to connect the graph construction more directly to manifold properties.

\FloatBarrier


\subsection{Multiscale optimization}\label{section:optimization}

We now begin to develop the iteration in Algorithm 1, given the initial Gabriel neighborhood graph, $G^{(0)}$.
Assuming (temporarily) that this gives correct local neighborhoods, what should the corresponding scales be for a Gaussian kernel? In effect this is an extension of $G$ into a weighted counterpart, $\mathcal{G}$. From Figure~\ref{fig:tangent-plane}, this weighted graph is also a type of approximation of (aspects of) the continuous manifold. Because density is not necessarily uniform, different points might have different neighborhood radii, so a multiscale Gaussian similarity kernel (equation~\ref{eq:ms-kernel}) is used. Each point $\bm{x}_i$ has its own associated scale, $\sigma_i$. To develop the computation of such scales, we now move into the continuous domain and exploit the geometric notion of a cover.

\subsubsection{Covering criterion}

A criterion for separability between two Gaussians has been developed in the mixture-of-Gaussians literature \citep{dasgupta-mog,vempala-mixture,arora-nonspherical}: two spherical Gaussians, $i$ and $j$, can be distinguished (in the sense of solving a classification problem) with reasonable probability when they have a separation of at least
\begin{equation}\label{eq:C-separation}
\|\bm{\mu}_i - \bm{\mu}_j\| > C\max\{\sigma_i,\sigma_j\},
\end{equation}
at which the overlap in their probability mass is a constant fraction \citep{vempala-mixture}.

\begin{figure}[!ht]
\centering
\includegraphics[width=1\textwidth]{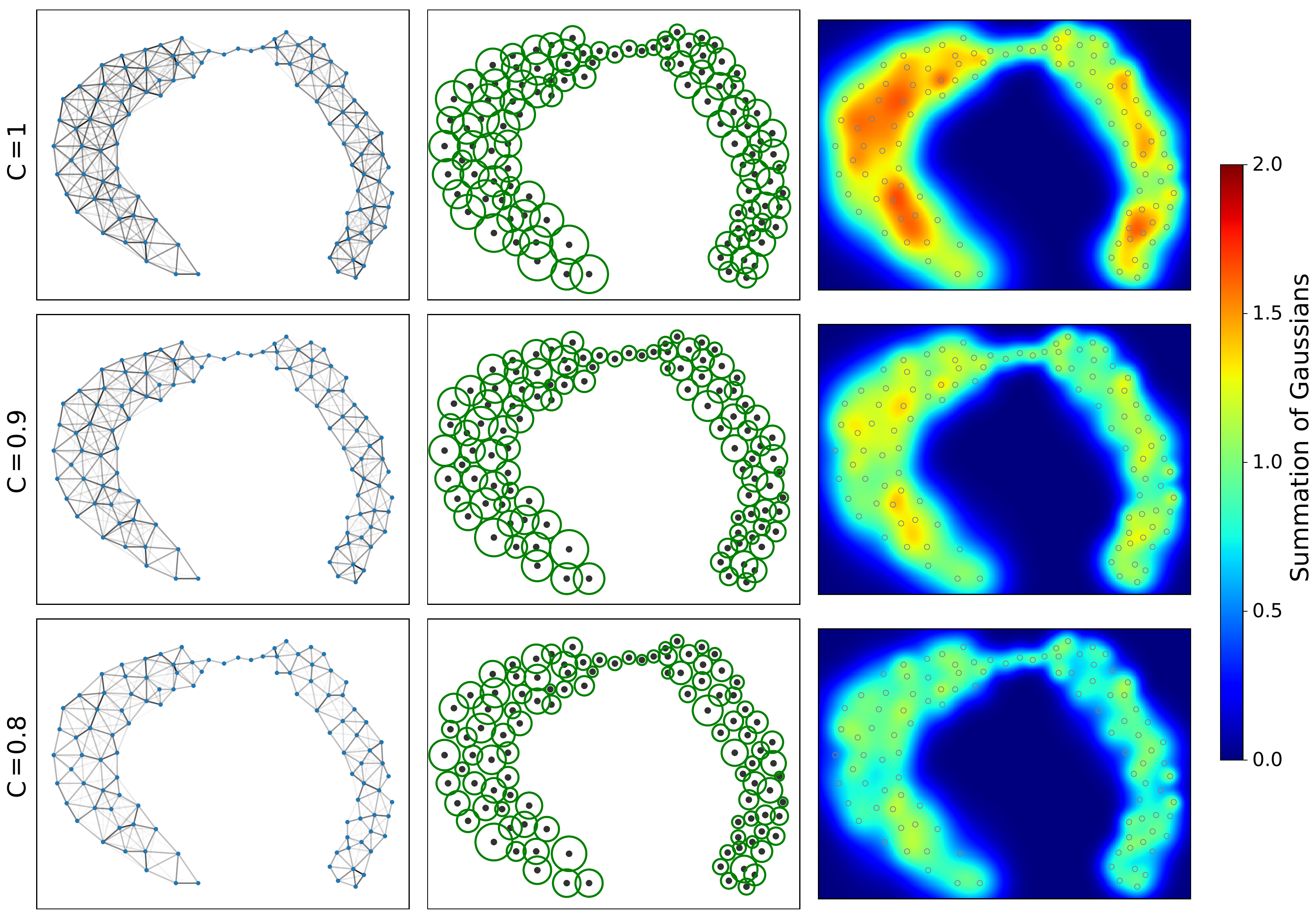}
\caption{
Effect of hyperparameter $C$ from equation~\ref{eq:C-connectivity} on the resulting weighted graph (left), optimal scales (middle), and manifold approximation (right, shown as the resulting summation over the Gaussian kernels around each point using their individual scales). For $C$ = 1 (top), the scales overlap too much and, as a result, the Gaussian summation (right) is highly non-uniform. For $C$ = 0.8 (bottom), the scales are not sufficiently large to properly cover the underlying manifold, resulting in holes (right). When $C$ = 0.9, there is a good compromise between covering and keeping a uniform density, so the Gaussian summation approximates a partition of unity (summing to $\sim$1 everywhere) when the scales correctly conform to the local sampling characteristics. Our approach will allow us to tune $C$ based on a relaxation statistic, $\delta_i'$.
}
\label{fig:partition-unity}
\end{figure}

We flip this around by using a different, but related, construction: consider
Gaussians now centered at the midpoints (i.e., not on data points) to indicate whether nearby points should be connected, not separated (Figure~\ref{fig:tangent-plane} illustrates this construction directly). Furthermore, because we use a multiscale kernel (equation~\ref{eq:ms-kernel}), the (non-normalized) Gaussian density becomes a function of $\sqrt{\sigma_i\sigma_j}$. Hence, we obtain a criterion for what we term  $C$\emph{-connectivity}:
\begin{displayquote}
\emph{\textbf{Definition:}}  Two neighbors $i$ and $j$ in the discrete graph $G=(E,V)$ are $C$-connected by the multiscale kernel when the geometric mean of their individual scales is at least the distance between $\bm{x}_i$ and $\bm{x}_j$ scaled by a positive constant, $C$:
\begin{equation}\label{eq:C-connectivity}
C\|\bm{x}_i - \bm{x}_j\| \leq \sqrt{\sigma_i\sigma_j}.
\end{equation}
\end{displayquote}
\noindent
The constant $C$ plays a role in normalizing for unknown density; it will be developed in section~\ref{section:Ctuning}. For now, we illustrate its role in the connection from graphs to manifolds. Figure~\ref{fig:partition-unity} shows the graph over a set of data points, and the local scales obtained (by the algorithm below) for different values of $C$. Choosing $C$ too large yields scales (and therefore Gaussians) that are too large, that is, their overlap has peaks. Choosing it too small yields scales that introduce holes. Choosing it correctly, the Gaussians form a covering of the manifold that approximates a partition of unity. Such partitions of unity are used in differential geometry to extend local information (in our case, the scales) to global information (a covering of the manifold).

\begin{figure}[ht!]
\centering
\includegraphics[width=1\textwidth]{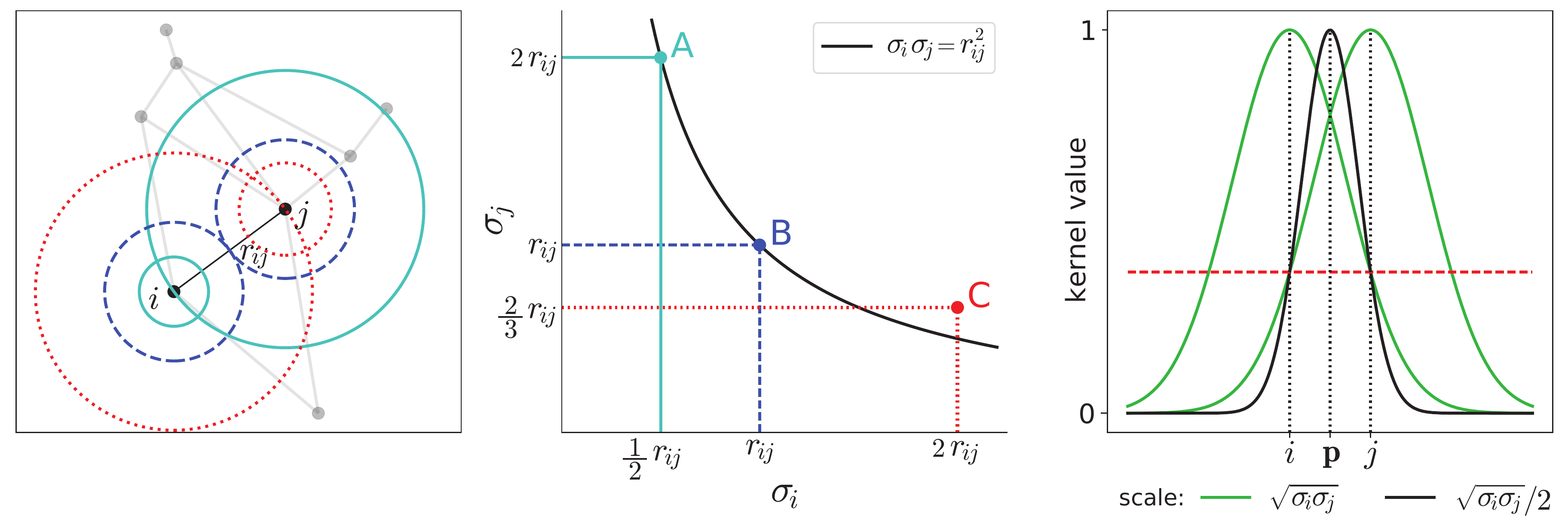}
\caption{
Covering constraint for the multiscale kernel of equation~\ref{eq:ms-kernel}.
\textit{Left:} A graph $G$ with two nodes $i$ and $j$ at a distance $r_{ij}$ from each other in $\mathbb{R}^n$.
Since they are connected, their assigned individual scales $\sigma_i$ and $\sigma_j$ 
must satisfy $\sigma_i\sigma_j \geq r_{ij}^2$, i.e., the covering constraint (here we assume $C$ = 1).
\textit{Center:} All feasible pairs ($\sigma_i,\sigma_j$) lie inside the region above a positive hyperbola, three of which are indicated as colored points; pairs A and B satisfy exactly, while C satisfies in excess.
Each one is also depicted as a pair of circles on the left plot using the same color code, each one centered at its corresponding node (radii are set to half the scale, for clarity).
Although pairs A and B differ in their ratio $\sigma_i/\sigma_j$, both result in the same multiscale kernel value for the edge ($i,j$), since the product $\sigma_i\sigma_j$ is the same; pair C yields a slightly higher value.
This illustrates the freedom that might exist in choosing an optimal combination of scales for all nodes (i.e., a covering).
\textit{Right:} multiscale kernel values, $K_{ij}$, centered at either $i$ or $j$, shown in green, are symmetric (with scale $\sqrt{\sigma_i\sigma_j}$). Horizontal axis represents position over the line in $\mathbb{R}^n$ passing through $i$ and $j$. A kernel centered at the midpoint $\bm{p}$ between $i$ and $j$ using half the scale (black curve) attains the same value as $K_{ij}$ at $i$ and $j$. Dashed red line indicates the common value between the three kernels.}
  \label{fig:ms-covering}
\end{figure}

By choosing appropriate scales, i.e., scales that meet our criterion for all edges in $E$, we also ensure a covering of the edges, in the following sense: the value of the multiscale kernel $K_{ij}$ between $\bm{x}_i$ and $\bm{x}_j$ is identical to that of a kernel re-centered at the midpoint $\bm{p} \equiv (\bm{x}_i + \bm{x}_j)/2$ and re-scaled using half the geometric mean of $\sigma_i$ and $\sigma_j$ as its scale, $\sigma_{\bm{p}}$:
\begin{equation}
K_{ij} = \exp{\frac{-\|\bm{x}_i-\bm{x}_j\|^2}{\sigma_i\sigma_j}} = \exp{\frac{-\|(\bm{x}_i-\bm{x}_j)/2\|^2}{\sigma_i\sigma_j/2^2}} = 
\exp{\frac{-\|(\bm{p}-\bm{x}_i)\|^2}{\sigma_{\bm{p}}^2}},
\end{equation}
with $\sigma_{\bm{p}} \equiv \sqrt{\sigma_i\sigma_j}/2$ (Figure~\ref{fig:ms-covering}).
\begin{remark}
We say a $C$\emph{-covering} is attained when every pair $(i,j) \in E$ is $C$\emph{-connected} (equation~\ref{eq:C-connectivity}). Additionally, when the spacing between neighboring points is approximately uniform locally, the pointwise summation over all Gaussian kernel bumps given by the individual scales provides an (un-normalized) partition of unity of $\mathcal{M}$.
\end{remark}

We now use the covering constraints to solve for the set of scales, $\bm{\sigma}$. It is desirable that the scales be small (respecting the reach), while at the same time maintaining the connectivity in $\mathcal{G}$ close to that of $G$. Thus, one idea is to find scales such that the sum of edge weights in $\mathcal{G}$ incident to a node $i$ from its neighbors in $G$ approximate the degree of $i$ in $G$, for all $i$, while at the same time ensuring a $C$-covering. This, however, amounts to a non-convex problem in which the cost function involves a summation of multiscale kernel values. We are unable to solve this efficiently. Instead, we find the smallest individual scales such that our covering criterion is satisfied for all edges (a ``minimal covering''), and later address the quality of the relaxation by using a statistical pruning (edge sparsification). This can be transformed into a convex, linear program with linear constraints by which all scales can be solved for simultaneously, as we show next. (We also present, in Appendix~\ref{section:greedy}, a greedy approach to this optimization that may be convenient when dealing with very large datasets.)

\subsubsection{Linear program relaxation}\label{section:constraints}

To achieve a minimal covering, one might minimize $\sum_i \sigma_i$ (or, equivalently, the 1-norm of the vector $\bm{\sigma}$, since scales are positive) subject to the covering constraint\footnote{Another possibility is to use a weighted sum $\sum_i \nu_i \sigma_i$ while keeping the same constraints, thus still guaranteeing a covering. The weights $\nu_i$ add a bias to how the length of an edge is split between its two incident nodes (by balancing their individual scales). One interesting option is to set $\nu_i=r_i^{\mathrm{non}}/r_i^{\mathrm{FN}}$, i.e., the ratio between the distance to the nearest non-neighboring point, $r_i^{\mathrm{non}}$, and the farthest neighbor, $r_i^{\mathrm{FN}}$.}.
This suggests the following:

\begin{displayquote}
\emph{\textbf{Optimization Problem:}}
\begin{equation}\label{eq:obj-ftn0}
\begin{aligned}
\min_{\bm{\sigma}} \quad & \bm{1}^\intercal\bm{\sigma} \\
\textrm{s.t.} \quad & (i,j)\ \textrm{is}\ C\textrm{-connected},\ \forall\ (i,j) \in E\\
  &\sigma_i\ \textrm{is bounded,}\ \forall\ i \in V,   \\
\end{aligned}
\end{equation}
\end{displayquote}

\noindent
where $\bm{\sigma}$ is the vector of individual scales, $\sigma_i$, and $\bm{1}$ is the all-ones vector.  Now it remains to represent the $C$-covering requirement by a set of constraints.

Looking in detail at $C$-connectedness (equation~\ref{eq:C-connectivity}) as a function of $\sigma_i$ and $\sigma_j$, observe that it represents a region delimited by a single-branched hyperbola (since the distance and scales are positive):
\begin{equation}\label{eq:hyperbola}
\sigma_i\sigma_j \geq (Cr_{ij})^2, \quad \sigma_i > 0, \sigma_j > 0,
\end{equation}
where $r_{ij} \equiv \|\bm{x}_i - \bm{x}_j\|_{\mathbb{R}^n}$). Each $\sigma_i$ is naturally bounded above by the distance to $i$'s farthest neighbor, $r_i^{\textrm{FN}}$:
\begin{equation}\label{eq:upper-bound}
\sigma_i \leq r_i^{\textrm{FN}},
\end{equation}
beyond which all neighbors are satisfied\footnote{That is assuming $C\leq1$ (a natural choice). If for some reason one needs to allow $C>1$, then the upper bounds must be scaled by $C$ in order to ensure feasibility.}, so further increasing either scale would make the weights to non-neighbors larger than strictly necessary (thereby hurting the kernel graph relaxation).
These bounds, combined, specify a bounding box for each edge that must necessarily be crossed (or at least touched) by the hyperbola, since $r_{ij} > 0$.

\begin{figure}[h!]
\centering
\includegraphics[width=1\textwidth]{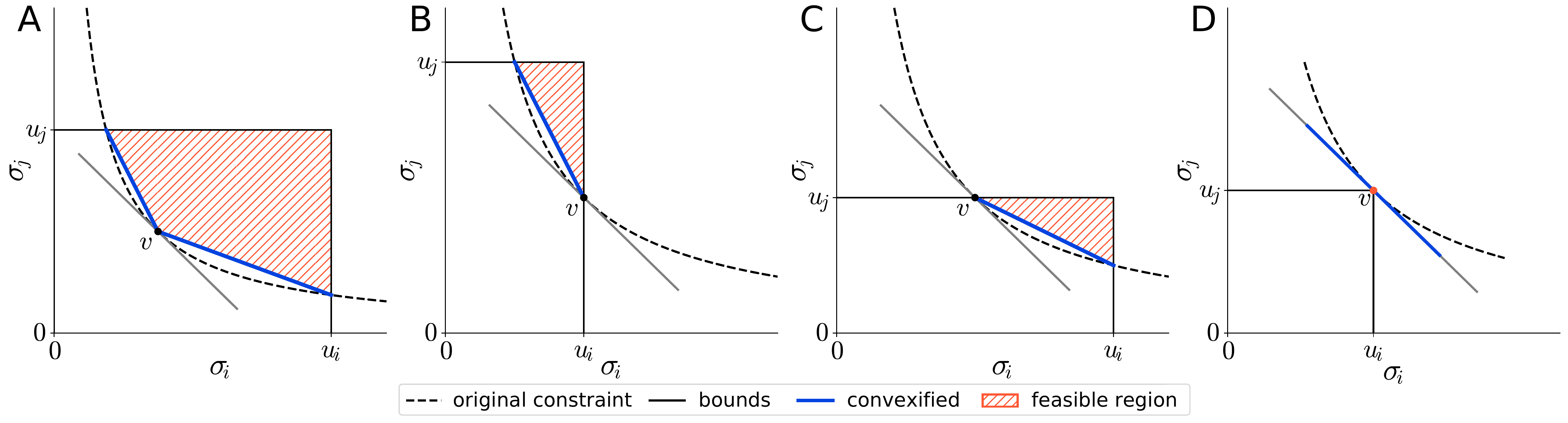}
\caption{Examples of constraints introduced by an edge, $e_{ij}$, in $G$. The $C$-connectivity rule, i.e., the hyperbola given by $\sigma_i \sigma_j = (C\|\bm{x}_i-\bm{x}_j\|)^2$ (dashed curve), when convexified, may give rise to one or two linear constraints, depending on whether the hyperbola's vertex $v$ (point where $\sigma_i = \sigma_j$) intersects the bounding box given by the lines $\sigma_i = 0$, $\sigma_j = 0$, $\sigma_i = u_i$, and $\sigma_j = u_j$, where $u_i$ and $u_j$ denote upper bounds. Hatched area (in orange) shows feasible region using convexified constraints; tangent line at $v$ is shown in gray.
When $v$ is interior to the bounding box (\textbf{A}), two secants (in blue) define the feasible region (namely, the lines passing through $v$ and the points where the hyperbola intersects the lines $\sigma_i = u_i$ and $\sigma_j = u_j$);
when either $v = u_i$ (\textbf{B}) or $v = u_j$ (\textbf{C}), only one secant is necessary;
when $v$ coincides with both $u_i$ and $u_j$ (\textbf{D}) (which may occur if $C$ is set to 1), again only one inequality is necessary, namely the tangent line at $v$.
  }
  \label{fig:constraints}
\end{figure}

Due to the hyperbolae, this amounts to a non-linear, non-convex set of constraints. However, we can convexify the feasible set by considering, for each edge $(i,j)$, the line(s) passing through the hyperbola's vertex (the point at which $\sigma_i$ = $\sigma_j$ = $Cr_{ij}$) and the points where the hyperbola intersects the bounding box. The four possibilities are shown in Figure~\ref{fig:constraints}. The feasible region for each edge, therefore, is bounded by a convex envelope given by such line(s) and those defined by the upper bounds to $\sigma_i$ and to $\sigma_j$.
Such envelopes for all edges, combined, define the boundaries of a convex polytope. Note that this convexification is conservative in the sense that only the objective is relaxed---the feasible scales are always at least as large as required by the original non-convex problem, therefore our covering requirement is not relaxed. (Because of the presence of a later pruning stage in the algorithm, it is better to over-connect here than to inadvertently disconnect nodes that should otherwise be connected.) 

Letting $m \leq 2|E|$ be the total number of linear constraints obtained as above, and $N$ the number of nodes in $G$, we define the $m\times N$ matrix $\bm{\Lambda}$ and the $m\times 1$ vector $\bm{b}$.
Now, for each edge, $e_{ij}$, let its two possible constraints be expressed as
\begin{align}
\sigma_j &\geq \alpha_{ij}^{(1)}\sigma_i + \beta_{ij}^{(1)} \\
\sigma_j &\geq \alpha_{ij}^{(2)}\sigma_i + \beta_{ij}^{(2)}
\end{align}
with $\alpha_{ij}$ and $\beta_{ij}$ denoting, respectively, the slope and intercept of the corresponding line(s) forming its convex envelope. Rearranging, we obtain $\alpha_{ij}\sigma_i - \sigma_j \leq - \beta_{ij}$ for each line, which is encoded as a row in $\bm{\Lambda}$ with values $\alpha_{ij}$ and $-1$ at columns $i$ and $j$, respectively (with zeros everywhere else), and an entry in $\bm{b}$ with value $-\beta_{ij}$:
                
\[
\renewcommand\arraystretch{1.3}
\begin{blockarray}{cccccccccc}
&  & & & & \bm{\Lambda} & & & &  \\
&\dots & &i& & \dots & &j& &  \dots\\
\begin{block}{@{}c[ccccccccc]}
  \vdots & & \vdots & & \vdots &  & \vdots & & \vdots &  \\
  e_{ij}^{(1)} & 0  & \dots &  \alpha_{ij}^{(1)} & \cdots & 0 & \dots & -1 & \dots & 0 \\
  e_{ij}^{(2)} & 0 & \dots &  \alpha_{ij}^{(2)} & \cdots & 0 & \dots &  -1 & \dots& 0 \\
  \vdots & & \vdots & & \vdots &  & \vdots & & \vdots & \\
\end{block}
& & & & & m\times N & & &  &  \\
\end{blockarray}
\begin{blockarray}{c}
  \\ \\ \\
  \bm{\sigma} \\
  \\ \\
  N \times 1 \\
\end{blockarray}
\leq
\begin{blockarray}{c}
\bm{b} \\ \\
\begin{block}{[c]}
  \vdots \\
  -\beta_{ij}^{(1)} \\
  -\beta_{ij}^{(2)} \\
  \vdots \\
\end{block}
m \times 1 \\
\end{blockarray}.
 \]
 
\begin{remark}
The convex envelope defining the constraints can be expressed by the linear inqualities:
\begin{equation}
\begin{aligned}
\bm{\Lambda} &\bm{\sigma} \leq \bm{b} \\
\bm{0} <\ &\bm{\sigma} \leq \bm{r}^{\text{FN}}, \\
\end{aligned}
\end{equation}
where $\bm{r}^{\text{FN}}$ is the vector of distances to each node's farthest neighbor.
\end{remark}
\noindent
Hence the problem now amounts to a convex, linear program (LP) with linear constraints:

\begin{displayquote}
\emph{\textbf{Optimization Problem:}}
\emph{\textbf{LP Relaxation:}}
\begin{equation}\label{eq:obj-ftn}
\begin{aligned}
\min_{\bm{\sigma}} \quad & \bm{1}^\intercal\bm{\sigma} \\
  \textrm{s.t.}\quad \bm{\Lambda} &\bm{\sigma} \leq \bm{b} \\
  \bm{0} <\ &\bm{\sigma} \leq \bm{r}^{\text{FN}}, \\
\end{aligned}
\end{equation}
\end{displayquote}
\noindent
which can be readily solved by a variety of methods \citep[see, e.g.,][]{boyd}. Figures~\ref{fig:horseshoe-iters}, \ref{fig:gauss3-iters}, and \ref{fig:limbs-iters} show the results of running this optimization on different examples.


\subsection{Sparsification}\label{section:stat-pruning}

\begin{figure}[!ht]
\centering
\includegraphics[width=1\textwidth]{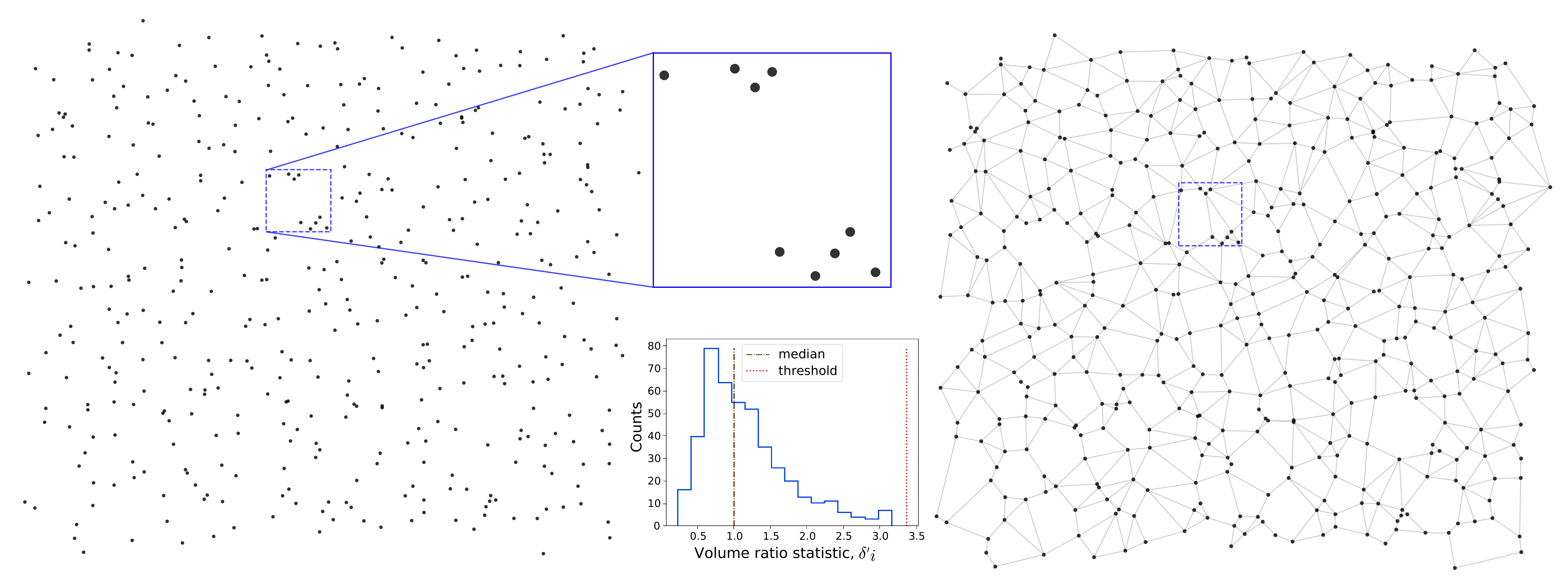}
\caption{
Local \textit{vs.} global assessment of neighborhoods. \emph{Left:} The points inside the cropped window appear to form two well-defined clusters when looked at up close (local estimation). However, when considered in the context of the full dataset (global estimation), the apparent gap between the top and bottom groups ``disappears,'' i.e., it is well within the range of gaps observed throughout the data. More precisely, it does not significantly deviate from the average sampling interval.
\emph{Right:} The converged graph $G$ indeed connects the two groups by edges, and the distribution of volume ratios, $\delta_i'$ (lower inset), confirms that all edges are reasonable.
}
  \label{fig:zoomed-pts}
\end{figure}

Summarizing what we have seen so far, the Gabriel graph provides an initial estimate of connectivity, while the LP optimization provides minimal scales for a continuous kernel to cover those connections. However, since the initial estimate of the discrete graph might contain incorrect connections, its resulting optimal scales might also be inadequate. An example of this can be seen in Figure~\ref{fig:horseshoe-iters}: initially, two pairs of nodes are connected across the central gap since a Gabriel ball exists between them. This will require very large scales to ``cover'' these edges. Furthermore, the Gabriel graph is based on a local connectivity rule; however, as illustrated in Figure~\ref{fig:zoomed-pts}, decisions about connecting nodes across a gap should not be local. We here address both of these issues, by introducing a global statistic based on how frequently such a gap occurs in the data. In terms of Algorithm~\ref{algo1}, we are now at steps 6 and 7.

\subsubsection{Volume ratio}\label{section:delta-stat}

Because incorrect connections can be given by Gabriel balls lying in the free space between parts of a manifold, i.e., across the medial axis, it is tempting to simply prune the longest connections. Note, however, that the size of a scale by itself is not necessarily important: in both examples shown in Figure~\ref{fig:algo-examples}, the non-uniform density causes scale sizes to vary considerably, and even the largest ones are appropriate, that is, are still consistent with the distances to neighboring points.

Conversely (and importantly), a scale that is excessively large will likely cover ``too many'' points. That is, it will cover neighbors in excess of the number of discrete neighbors of its corresponding node in $G$.  We quantify this notion by observing that an individual scale, $\sigma_i$, should produce kernel values whose sum is comparable to the discrete degree, $\mathrm{deg}(i)$, of node $i$ in $G$. As will be shown, after proper normalization this also means $\sigma_i$ shall relate to a local volume element around $\bm{x}_i$, or the inverse of the local density. Since each connection in $G$ can be seen as having unit weight, a Gaussian kernel around $\bm{x}_i$ with scale $\sigma_i$ should distribute that same amount, $\mathrm{deg}(i)$, only continuously over ambient space. 

We start our derivation with a definition:
\begin{displayquote}
\emph{\textbf{Definition:}}
let $w_{ij}^{(\sigma_i)}$ be the Gaussian kernel value between $\bm{x}_i$ and $\bm{x}_j$ using scale $\sigma^{(t)}_i$ at iteration $t$. A (non-isolated) node's \emph{volume ratio} at iteration $t$, denoted by $\delta_i^{(t)}$, is defined as
\begin{equation}\label{eq:naive-deg-ratio}
    \delta_i^{(t)} \equiv \frac{\sum_{j \in V} w_{ij}^{(\sigma_i^{(t)})}}
    {\sum_{j \in V} a_{ij}},
\end{equation}
\end{displayquote}
\noindent
i.e., the ratio between node $i$'s weighted degree due to $\sigma_i^{(t)}$ and its discrete degree in $G^{(t)}$ (henceforth we suppress the iteration dependency ($t$) to simplify notation).
An individual-scale Gaussian kernel is needed to correctly assess the impact of $\sigma_i$ on the relaxation from the perspective of $i$ alone---the multiscale kernel here might artificially increase the weighted degree of $i$ when other nodes (even non-neighbors of $i$!) have incorrect scales. (Nevertheless, as discussed below, a corresponding ratio using the actual weights in $\mathcal{G}$ may eventually be used for convergence purposes.)

\begin{figure}[h!]
  \centering
\includegraphics[width=1\textwidth]{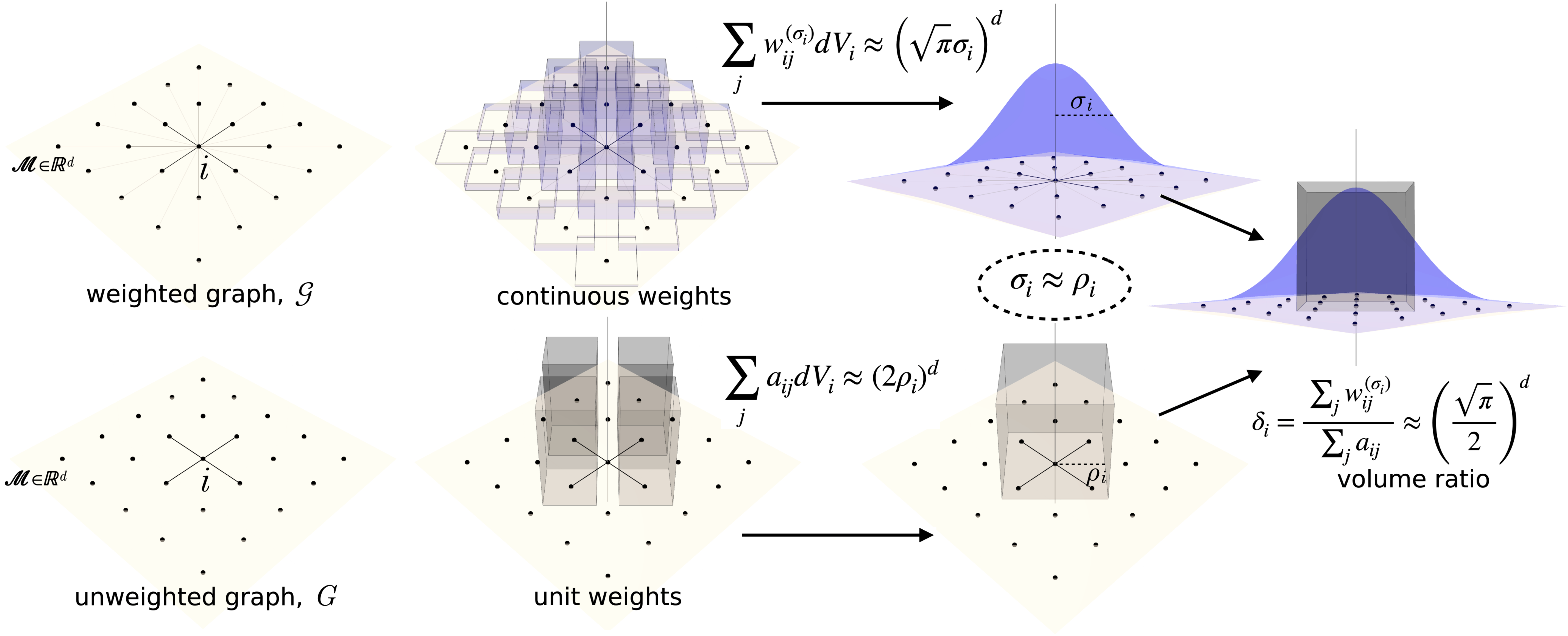}
  \caption{Computing the volume ratio between continuous and discrete degrees of a node $i$ with neighboring points sampled uniformly over a $d$-dimensional manifold $\mathcal{M}$. \emph{Top row:} Using a Gaussian kernel, the weighted degree of $i$ (sum of kernel values $\sum_j w_{ij}^{(\sigma_i)}$) in $\mathcal{G}$  approximates the volume of a Gaussian with scale $\sigma_i$ (equation~\ref{eq:sumw-gaussvol}). \emph{Bottom row:} The number of edges adjacent to $i$ in $G$ (sum of unit weights) approximates the volume of a box with unit height and a hypercube of side $2\rho_i$ as its base, where $\rho_i$ is the radius of a local volume element of $\mathcal{M}$ around $\bm{x}_i$ (equation~\ref{eq:meanvaluebox}). \emph{Right:} When the scale $\sigma_i$ is compatible with $\rho_i$, the volume ratio, $\delta_i$, is expected to be approximately $(\sqrt{\pi}/2)^d$, and therefore is a scale-invariant quantity.
  }
  \label{fig:volumes}
\end{figure}

Now, using a mean-value integral \citep[as in][]{laplaciantomography}, the numerator approximates the volume under the continuous Gaussian kernel over $\mathcal{M}$, and can be further approximated by
\begin{equation}\label{eq:meanvaluegauss1}
    \sum_j w_{ij}^{(\sigma_i)} \approx \frac{N}{\mathrm{vol}(\mathcal{M})} \int_{\mathcal{M}} \mathrm{exp}\left( \frac{-\|\bm{x}_i - \bm{x}_j\|^2}{\sigma_i^2}\right) d\bm{x}_j
\end{equation}
when $\mathcal{M}$ has uniform density and low curvature. In practice, the kernel will have compact support due to numerical precision (i.e., its values become effectively zero for sufficiently large distances), so by defining the volume element $dV_i \equiv \mathrm{vol}(\mathcal{N}(\bm{x}_i))/|\mathcal{N}(\bm{x}_i)|$ of a neighborhood $\mathcal{N}(\bm{x}_i) \in \mathcal{M}$ around $\bm{x}_i$, we may rewrite equation~\ref{eq:meanvaluegauss1} as
\begin{equation}
    \sum_j w_{ij}^{(\sigma_i)} dV_i \approx \int_{\mathcal{M}} \mathrm{exp}\left( \frac{-\|\bm{x}_i - \bm{x}_j\|^2}{\sigma_i^2}\right) d\bm{x}_j
\end{equation}
when the sampling is approximately uniform around $\bm{x}_i$. By further assuming that $\sigma_i$ is small, and that $\mathcal{M}$ can be well-approximated locally by its tangent space $\mathbb{R}^d$, then 
\begin{equation}\label{eq:meanvaluegauss}
    \int_{\mathcal{M}} \mathrm{exp}\left( \frac{-\|\bm{x}_i - \bm{x}_j\|^2}{\sigma_i^2}\right) d\bm{x}_j \approx \int_{\mathbb{R}^d} \mathrm{exp}\left( \frac{-\|\bm{x}_i - \bm{x}_j\|^2}{\sigma_i^2}\right) d\bm{x}_j = (\sqrt{\pi}\sigma_i)^d,
\end{equation}
so
\begin{equation}\label{eq:sumw-gaussvol}
    \sum_j w_{ij}^{(\sigma_i)} dV_i \approx (\sqrt{\pi}\sigma_i)^d,
\end{equation} 
as shown in Figure~\ref{fig:volumes}.

An analogous derivation for the discrete degree summation is as follows. First, note that the edge weight in this case is a constant (unity); it remains to determine its support over $\mathcal{M}$. From section~\ref{section:GG-degrees}, we know that, for simple manifolds with random sampling, the node degree $\mathrm{deg}(i)$ in a Gabriel graph is approximately $2^{d_i}$ within a region of constant intrinsic dimensionality, where $d_i$ denotes the local intrinsic dimension around $\bm{x}_i$ (possibly different around other points in $\mathcal{X}$)\footnote{We abuse notation, therefore, when we say ``$d$-dimensional manifold'', or ``$\mathcal{M} \in \mathbb{R}^d$''.}.
In more general manifolds, we expect the converged graph $G^{\star}$ instead to approach such a property. This means $\sum_j a_{ij} \approx 2^{d_i}$ will approximate the volume of a hyperrectangle (or box) of unit height and having a ${d_i}$-dimensional hypercube of side 2 as its base\footnote{This agrees with our observation (section~\ref{section:GG-degrees}) that the unoccluded region around $\bm{x}_i$ is similar to a $d_i$-orthoplex: by placing a vertex (i.e., a neighbor) in each of its $2^{d_i}$ facets, we obtain a $d_i$-hypercube, which is the dual polytope of an orthoplex.}. So, by defining $\rho_i$ as the radius of the local volume element $dV_i$ (such that $\rho_i = \sqrt[d_i]{dV_i}$), we may write:
\begin{equation}\label{eq:meanvaluebox}
    \sum_j a_{ij} dV_i \approx \int_{-\rho_i}^{\rho_i} \cdots \int_{-\rho_i}^{\rho_i} 1 dx_{j_1}\ldots dx_{j_d} = (2\rho_i)^{d_i},
\end{equation}
as illustrated in Figure~\ref{fig:volumes}. Hence, $\rho_i$ is a kind of ``neighborhood radius'' of $\bm{x}_i$.

From equations~\ref{eq:sumw-gaussvol}--\ref{eq:meanvaluebox}, equation~\ref{eq:naive-deg-ratio} becomes
\begin{equation}\label{eq:empir-vol}
\frac{\sum_j w_{ij}^{(\sigma_i)}}{\sum_j a_{ij}} =  \frac{\sum_j w_{ij}^{(\sigma_i)}\ dV_i }{\sum_j a_{ij} \ \ dV_i} \approx \left ( \frac{\sqrt{\pi}\sigma_i}{2 \rho_i} \right )^{d_i},
\end{equation}
representing the ratio between the volume of a Gaussian with scale $\sigma_i$ and that of a box of side $2\rho_i$ and height 1 (cf. Figure~\ref{fig:volumes}). As the algorithm approaches convergence, we expect $\sigma_i \approx \rho_i$ (scales are compatible with neighborhood radius) and $\mathrm{deg}(i)$ should approach, on average, the empirically-observed value of $2^{d_i}$ (meaning that the number of neighbors in $G$ is compatible with dimensionality of $\mathcal{M}$). This results in
\begin{equation}\label{eq:expected-ratio}
\frac{\sum_j w_{ij}^{(\sigma_i)}}{\sum_j a_{ij}} \approx \left ( \frac{\sqrt{\pi}}{2}\right )^{d_i}.
\end{equation}
Finally, we can estimate $d_i$ as
\begin{equation}\label{eq:estimate-dim}
    \tilde{d}_i \equiv \log_2 \sum_j a_{ij},
\end{equation}
based on the empirical degree distribution of $G^{(t)}$. From this, we can compute a \emph{normalized} volume ratio, $\delta_i'^{(t)}$, dividing $\delta_i^{(t)}$ by the value from equation~\ref{eq:expected-ratio}:
\begin{displayquote}
\emph{\textbf{Definition:}} A node's \emph{normalized volume ratio} is computed as
\begin{equation}\label{eq:corr-vol-ratio}
    \delta_i'^{(t)} \equiv 
    \frac{\sum_j w_{ij}^{(\sigma_i)}}{\sum_j a_{ij}}
    \left(\frac{2}{\sqrt{\pi}}\right)^{\tilde{d}_i}.
\end{equation}
\end{displayquote}

Nodes whose degree deviate from exactly $2^{d_i}$ will, likewise, under- or overestimate the local dimension, so reasonable volume estimates are still obtained regardless. However, in order to avoid dimension less than 1 for connected nodes, in practice when  $\mathrm{deg}(i)$ = 1 we replace $\sum_j a_{ij}$ with $\max\{2,\sum_j a_{ij}\}$.

Thus, we expect $\delta_i' \approx 1$ for points obeying $\sigma_i \approx \rho_i$ and $\tilde{d}_i \approx {d_i}$. Crucially, points for which these conditions are not met (those having ``wrong'' neighbors in the original Gabriel graph, $G^{(0)}$) will depart from this by having $\delta_i' \gg 1$. In the next section, we shall use this fact to guide a sparsification of edges in $G^{(0)}$ based on $\delta_i'$.

Interestingly, $\delta_i'$ can also be interpreted as measuring how well the scale $\sigma_i$ fits the local volume element $dV_i$ (or, equivalently, how it counteracts the local sampling density, $1/dV_i$). Since $dV_i = \rho_i^{d_i}$ (from the definition of $\rho_i$), we may rewrite equation~\ref{eq:empir-vol} as:
\begin{equation}
\frac{\sum_j w_{ij}^{(\sigma_i)}}{\sum_j a_{ij}} \approx \frac{(\sqrt{\pi}\sigma_i)^{d_i}}{2^{d_i} dV_i}.
\end{equation}
Summarizing the above, when $\tilde{d}_i \approx d_i$ and $\sigma_i \approx \rho_i$ we have:

\begin{remark}
A node's normalized volume ratio may alternatively be expressed as
\begin{equation}
    \delta_i'^{(t)} \equiv 
    \frac{\sum_j w_{ij}^{(\sigma_i)}}{\sum_j a_{ij}}
    \left(\frac{2}{\sqrt{\pi}}\right)^{\tilde{d}_i} \approx
    \frac{(\sqrt{\pi}\sigma_i)^{d_i}}{2^{d_i} dV_i}
    \left(\frac{2}{\sqrt{\pi}}\right)^{\tilde{d}_i}
    \approx \frac{\sigma_i^{d_i}}{dV_i}.
\end{equation}
\end{remark}

\noindent
Therefore, $\delta_i'$ can be thought of as the product between kernel scale and local density. When $\sigma_i$ is optimal, it should be approximately equal to the inverse of the local density, so $\delta_i' \approx 1$. 

\subsubsection{Uniformity of sampling and edge pruning} \label{section:Ctuning}

Since $\delta_i'^{(t)}$ is evaluated for every node $\bm{x}_i$, we can collect it across nodes and view it as a statistic. This has two consequences: (\textit{i}) it can be used to enforce consistency in sampling, and (\textit{ii}) outliers in this statistic are likely candidates for edge pruning. We address consistency of sampling first.

We have several times stated that sampling is required to be locally uniform, although its rate may change over the manifold. Examples of this were shown in, e.g., Figure~\ref{fig:deg-histograms}, where the sampling was denser in the center of the Gaussian distribution than in the periphery. This example differs from the regular grids, in which all nearest neighbors had exactly the same distance. Putting this together, we have:
\begin{remark}
{\it Locally Uniform Sampling:} Let node $i$ have $k_i$ neighbors in $G^{(t)}$. Among these, let $r_i^{\mathrm{FN}}$ denote the distance from $\bm{x}_i$ to its farthest neighbor, and $r_i^{\mathrm{NN}}$ that to its nearest neighbor. When $r_i^{\mathrm{FN}} \approx r_i^{\mathrm{NN}}$ for all $i$, we say the sampling is locally uniform. 
\end{remark}

This is useful because a departure from the assumption that sampling is locally uniform will cause $\delta_i'$ to be on average greater than 1 throughout the dataset. To see this, when sampling is not uniform, we have $r_i^{\mathrm{FN}} > r_i^{\mathrm{NN}}$. Now, since $\sigma_i$ is optimized to cover all of $i$'s neighbors, it will have in most cases the same order of magnitude as $r_i^{\mathrm{FN}}$ (minus some possible slack due to the multiscale interaction). Therefore, the higher the variability in the neighbors' distances, the larger the difference between $r_i^{\mathrm{FN}}$ and $r_i^{\mathrm{NN}}$ will be, making $\sigma_i$, in turn, be larger than the distance to most neighbors of $i$. Ultimately, this will increase $\sum_j w_{ij}^{(\sigma_i)}$ beyond what we would have in a uniform-sampling scenario (in which $r_i^{\mathrm{FN}} \approx r_i^{\mathrm{NN}}$).

When data are acquired using a global sampling strategy, this variability in the neighbors' distances should be roughly constant throughout the dataset (rather than the distances). So we use the scalar parameter, $C$, from equation~\ref{eq:C-connectivity} to correct for this ``bias'' and bring the median of the distribution of $\delta_i'^{(t)}$ (denoted as $\langle \delta_i'^{(t)} \rangle$) close to 1\footnote{Although the mean typically gives smoother tuning curves, the median is more robust. This matters, because of the possible outlying $\delta_i'$ values.}. 
\begin{remark}
Let the tuned $C^{\star (t)}$ be that which causes $\langle \delta_i'^{(t)} \rangle$ to be closest to 1.
\end{remark}

\begin{figure}[!ht]
  \centering
  \includegraphics[width=1\textwidth]{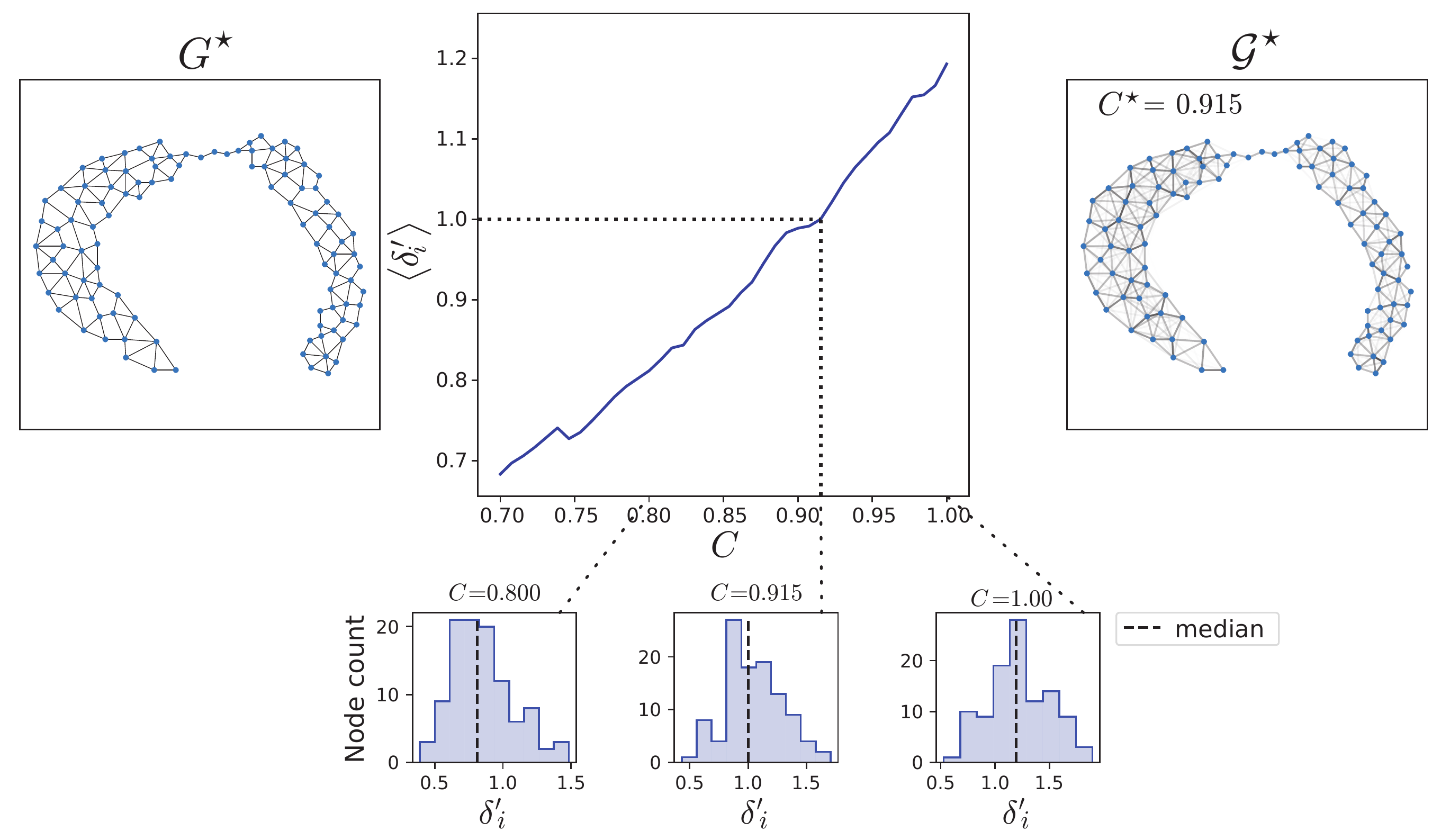}
    \caption{Tuning the hyperparameter $C$ based on the median of the distribution of normalized volume ratios, $\langle \delta_i' \rangle$.
    \emph{Left:} Converged unweighted graph $G^{\star}$ obtained for the dataset from Figure~\ref{fig:horseshoe-iters}.
    \emph{Center:} After computing $\langle \delta_i' \rangle$ for a range of values of $C \leq 1$, the optimal $C^{\star}$ is that resulting in a $\langle \delta_i' \rangle$ closest to 1. Histograms below show the distribution of $\delta_i'$ for different values of $C$, including $C^{\star}$ = 0.915.
    \emph{Right:} The resulting weighted graph $\mathcal{G}^{\star}$ after $C$-tuning typically exhibits a more uniform connectivity throughout (see Figure~\ref{fig:partition-unity}).
    }
  \label{fig:tuningC}
\end{figure}

Typically, $C^{\star (t)} < 1$, which, in the scale optimization procedure, means that the covering constraints (equation~\ref{eq:hyperbola}) are being relaxed using the distribution of $\delta_i'$ as a guide (Figure~\ref{fig:tuningC}).
Note that, although the tuning of $C$ is not necessary for finding candidates for sparsification, it attributes a quantitative meaning to the value of $\delta_i'$, so any $\delta_i' \gg 1$ is guaranteed to indicate the need for edge pruning. Such tuning should be performed at $t$ = 0, and repeated as needed over the iterations whenever $\langle \delta_i'^{(t)} \rangle$ deviates too much from unity (which may happen after several edges have been pruned). Most commonly, we find $0.5 < C^{\star(t)} < 1$.

Thus, we have a data-driven way of finding an appropriate value for $C$. Because it is a global constant applied to all connection constraints, it shifts the distribution of $\delta_i'$ to have median around 1 without changing its general shape.

This leads us to the second use of our statistic: any node whose normalized volume ratio is much greater than the median of the population should be identified as an outlier. Such nodes will have a neighbor considerably farther than its other neighbors (relative to the median variability of such neighboring distances throughout the data), and are candidates for the sparsification step.
\begin{remark}
Nodes that are robust outliers according to the $\delta_i'$ statistic have an overly distant neighbor (relative to the other neighbors for that node) and hence are likely to be in violation of reach or other geometric constraints. These {\em relatively distant neighbors} are candidates for having an edge pruned.
\end{remark}

\begin{figure}[!ht]
  \centering
  \includegraphics[width=1\textwidth]{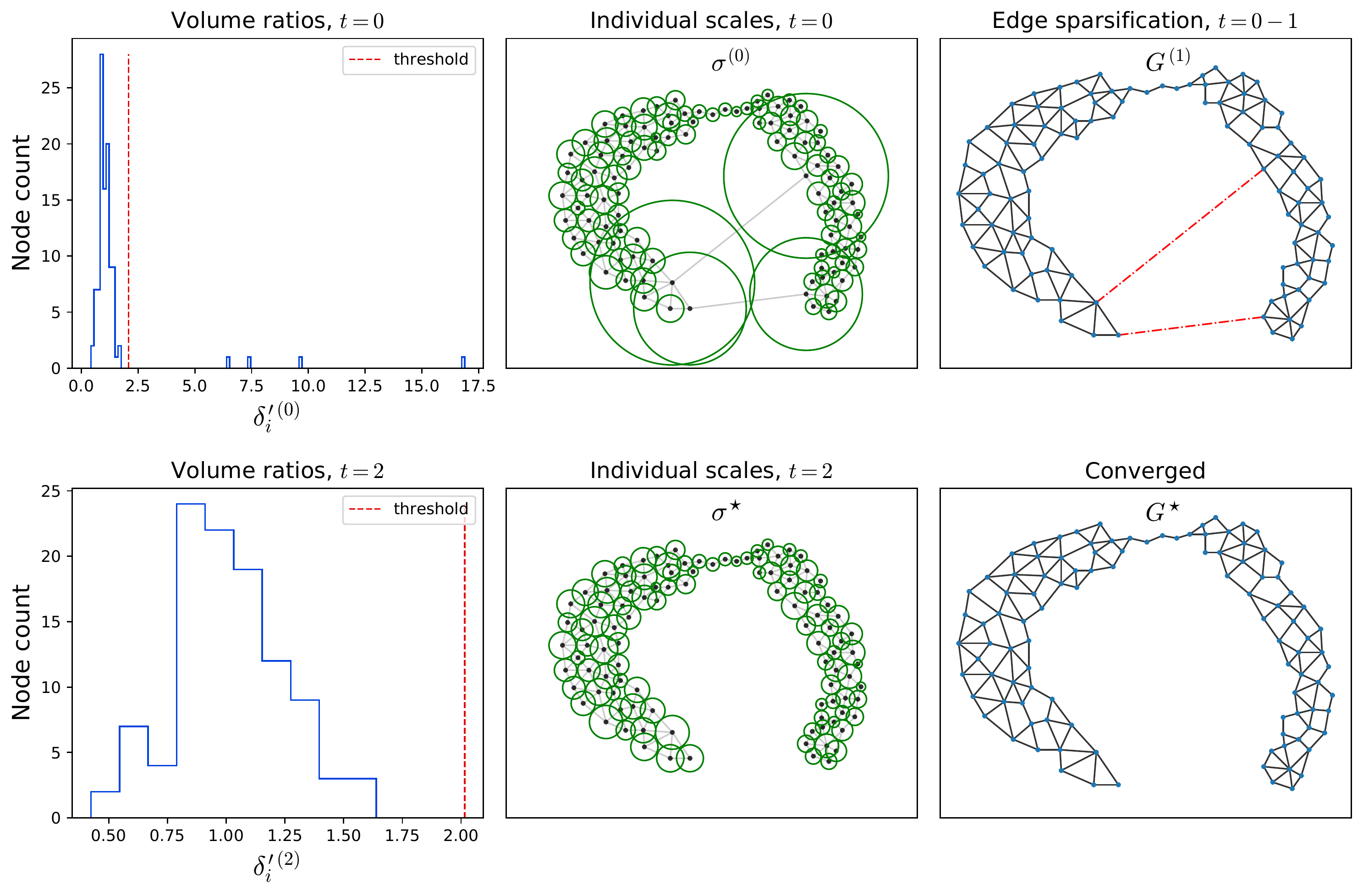}
  \caption{Optimal scales and associated normalized volume ratios at iterations 0 and 2 of Algorithm~\ref{algo1} on the horseshoe dataset (see Figure~\ref{fig:partition-unity}). \emph{Top row:} the $\delta_i'$ statistic has median around 1 and several outliers. These are caused by the long edges and huge scales (middle). \emph{Right column:} $G^{(t)}$ after iteration 1, with edges deleted shown in red (top), and after iteration 2 (bottom). 
  }
  \label{fig:horseshoe-iters}
\end{figure}

\begin{figure}[!ht]
  \centering
  \includegraphics[width=1\textwidth]{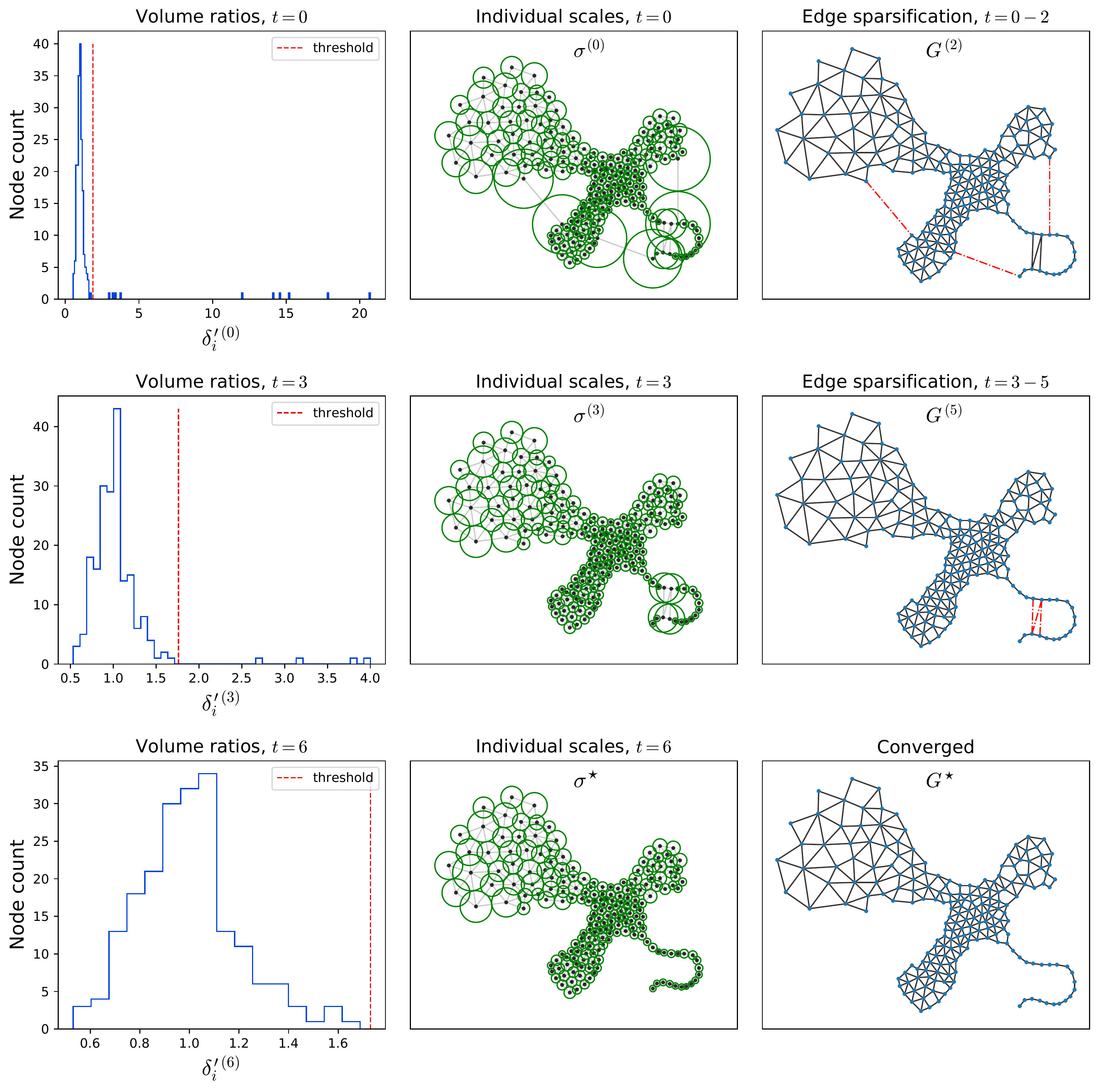}
  \caption{Optimal scales and associated normalized volume ratios at different iterations of Algorithm~\ref{algo1} on the dataset from Figure~\ref{fig:algo-examples}. The distribution of $\delta_i'$ (left) indicates those connections that are least likely to represent reasonable geodesics over the underlying manifold. Right column shows $G^{(t)}$ after iterations 2, 5 and 6 (deleted edges in red).
  }
  \label{fig:limbs-iters}
\end{figure}

\begin{figure}[!ht]
  \centering
  \includegraphics[width=1\textwidth]{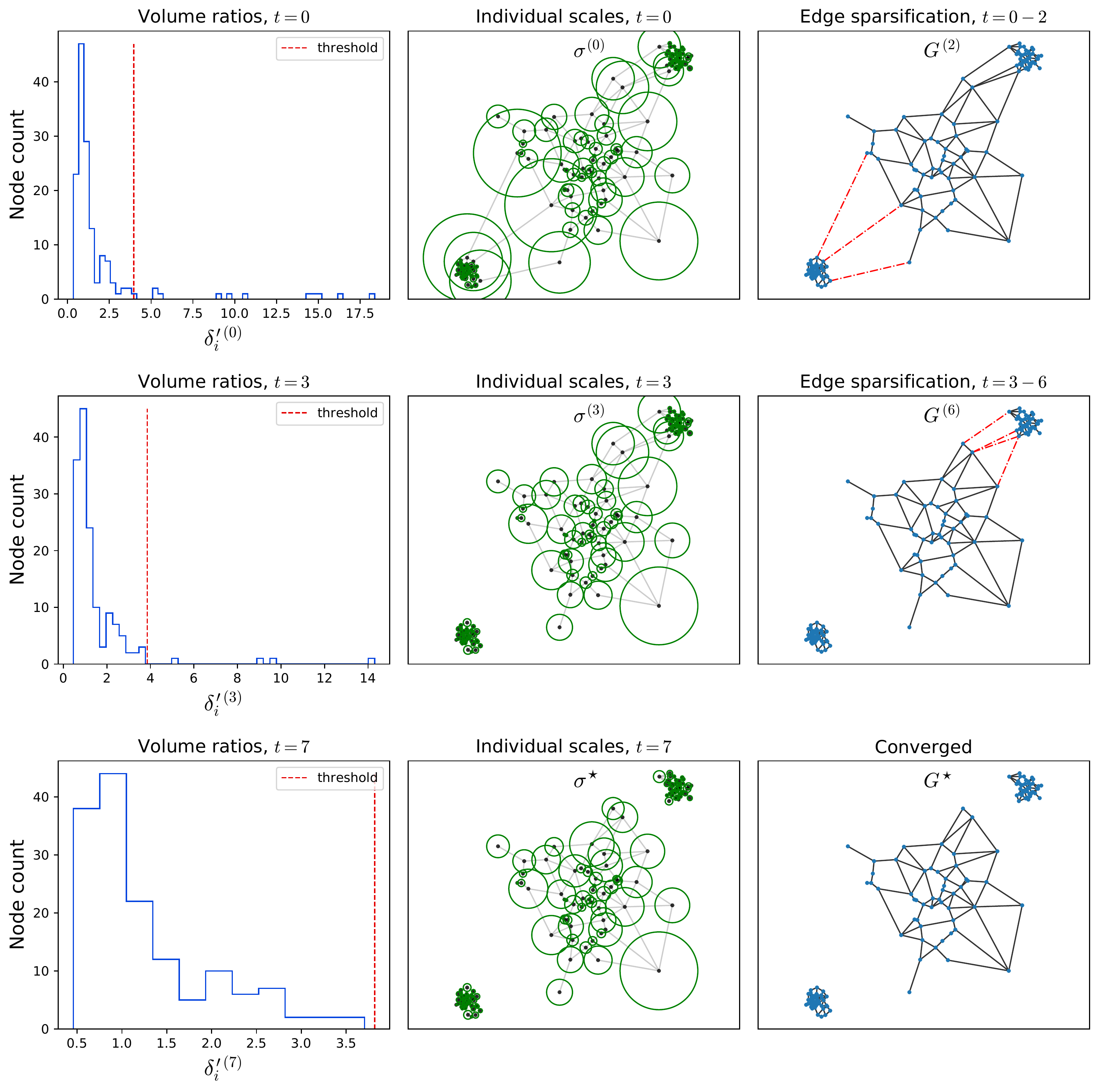}
  \caption{Optimal scales and associated volume ratio statistics at different iterations of Algorithm~\ref{algo1} on the clustered dataset from Figure~\ref{fig:algo-examples}. Pruned edges (in red) are precisely those connecting the three clusters together.
  }
  \label{fig:gauss3-iters}
\end{figure}

\begin{figure}[!ht]
  \centering
  \includegraphics[width=1\textwidth]{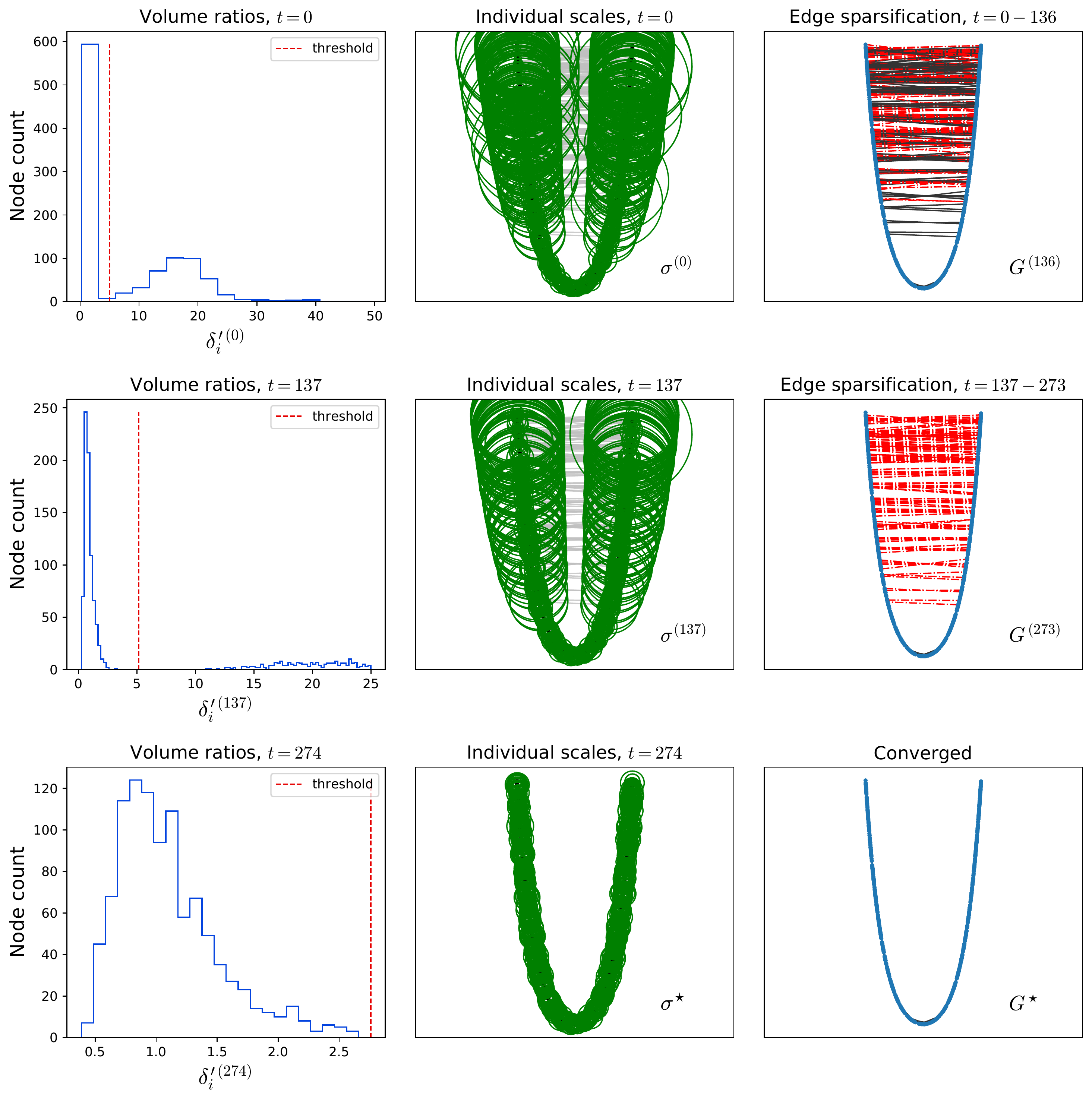}
  \caption{Optimal scales and associated normalized volume ratios $\delta_i'$ after each iteration of the algorithm on the dataset from Figure~\ref{fig:embeds-bentplane} (here, seen from a lateral view). The number of initial connections in $G^{(0)}$ (Gabriel graph) is very large, so the initial distribution of $\delta_i'$ shows two modes. However, ratios in right-side peak are very high, and are therefore easily identified as outliers. The algorithm converges soon after all edges crossing the gap are eliminated.
  }
  \label{fig:bentplane-iters}
\end{figure}

Given the distribution of normalized volume ratios, statistical models can be used to define a threshold for identifying outliers (see Figures~\ref{fig:horseshoe-iters}--\ref{fig:bentplane-iters}). It is likely that datasets with a large number of problematic connections will exhibit a distribution with a heavy tail, or that looks like a mixture of two distributions (cf. example in Figure~\ref{fig:bentplane-iters}), so using the distribution's quartiles may give a more robust result. One option that seems to work particularly well is to use estimates of the sample mean and standard deviation from the quartiles, as in \cite{mean-std-estim} (throughout, we make use of the C3 method derived therein, setting the $\delta_i'$ threshold to 4.5 standard deviations above the mean thus estimated). Still, we found that results are typically quite invariant to this particular choice, especially in real-life datasets. Finally, we note that our algorithm can be run interactively, so the user can analyze the histogram of the distribution after each iteration to judge whether the choice of threshold is reasonable and thus be confident in the results.

Nodes with $\delta_i'$ above the threshold should have their connection to their farthest neighbor deleted. Ideally, only one such connection is pruned after each iteration; however, should that become impractical with large datasets, a compromise is to limit the pruning, at each iteration, to a single edge from each node that is above the threshold (giving the chance for its $\delta_i'$ value to be updated before the next pruning). 

\subsubsection{Convergence}\label{section:convergence}

The algorithm converges at iteration $t$ when no point $i$ has an outlying $\delta_i'^{(t)}$ (i.e., greater than a statistical threshold). This implies that no edges will be pruned, so $G^{(t+1)} = G^{(t)}$ and therefore no further changes can occur to either $\bm{\sigma}^{(t)}$ or $\mathcal{G}^{(t)}$. Note that convergence is guaranteed: since at every iteration $t$ an edge must be removed, the algorithm necessarily reaches a certain $t$ at which all outliers (if there were any to begin with) have been pruned.

If one is solely interested in obtaining $\mathcal{G}^{\star}$ (i.e., not interested in $G^{\star}$), an alternative convergence condition may be adopted that looks at the distribution of the (normalized) \emph{multiscale volume ratio}, $\delta'^{(t)}_{i_{\mathrm{MS}}}$:
\begin{equation}\label{eq:ms-vol-ratio}
    \delta'^{(t)}_{i_{\mathrm{MS}}} \equiv 
    \frac{\sum_j w_{ij}}{\sum_j a_{ij}}
    \left(\frac{2}{\sqrt{\pi}}\right)^{\tilde{d}_i},
\end{equation}
analogous to equation~\ref{eq:corr-vol-ratio} but using the weights from $\mathcal{G}^{(t)}$ directly.
Since the multiscale kernel takes into account the interaction between individual scales, the distribution of $\delta'_{i_{\mathrm{MS}}}$ will be typically tighter than that of $\delta'_i$ (i.e., some of the excessively large scales might be compensated by small neighboring scales). Therefore, one may wish to allow for an earlier convergence when there are no remaining outliers in the distribution of $\delta'_{i_{\mathrm{MS}}}$.

Finally, in applications where it is required that $G^{\star}$ be connected, pruning can simply be stopped before disconnection. Naturally, $\mathcal{G}^{\star}$ is always connected up to machine precision or some numerical tolerance.

\begin{figure}[h!]
  \centering
  \includegraphics[width=1\textwidth]{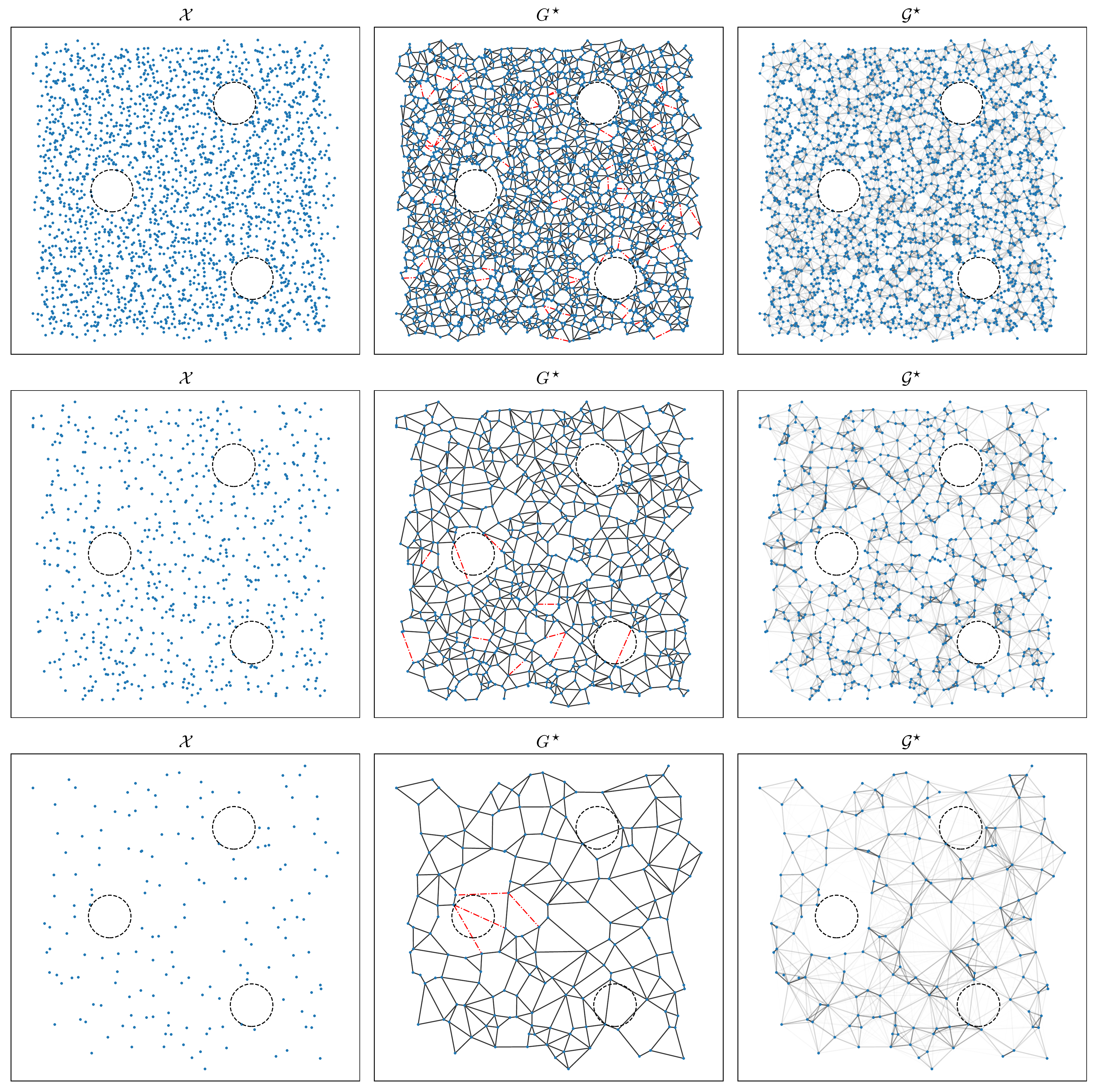}
    \caption{Sampled Swiss cheese results (cf. Figure~\ref{fig:swiss-cheese-pts}). The original sampled points (true holes outlined) are shown, together with the converged graphs. In the sparse case (bottom), sampling is close to locally uniform so not all holes are correctly inferred. As sampling gets denser (top two rows), no holes are violated. 
    }
  \label{fig:swiss-cheese}
\end{figure}

In closing this section, we return to one of our introductory examples and show, in Figure~\ref{fig:swiss-cheese}, the resulting graphs for the {\it sampling Swiss cheese} patterns (from Figure~\ref{fig:swiss-cheese-pts}). 
When sampling is too sparse (bottom), there is only so much that can be inferred, and not all holes are free of edges after convergence. As sampling gets denser, however, the algorithm correctly identifies that edges across holes should be pruned  (middle). When it is very dense (top), even the initial Gabriel graph is able to correctly infer the true holes.


\subsection{Comparison with other kernel methods}

\begin{figure}[h!]
  \centering
   \includegraphics[width=1\textwidth]{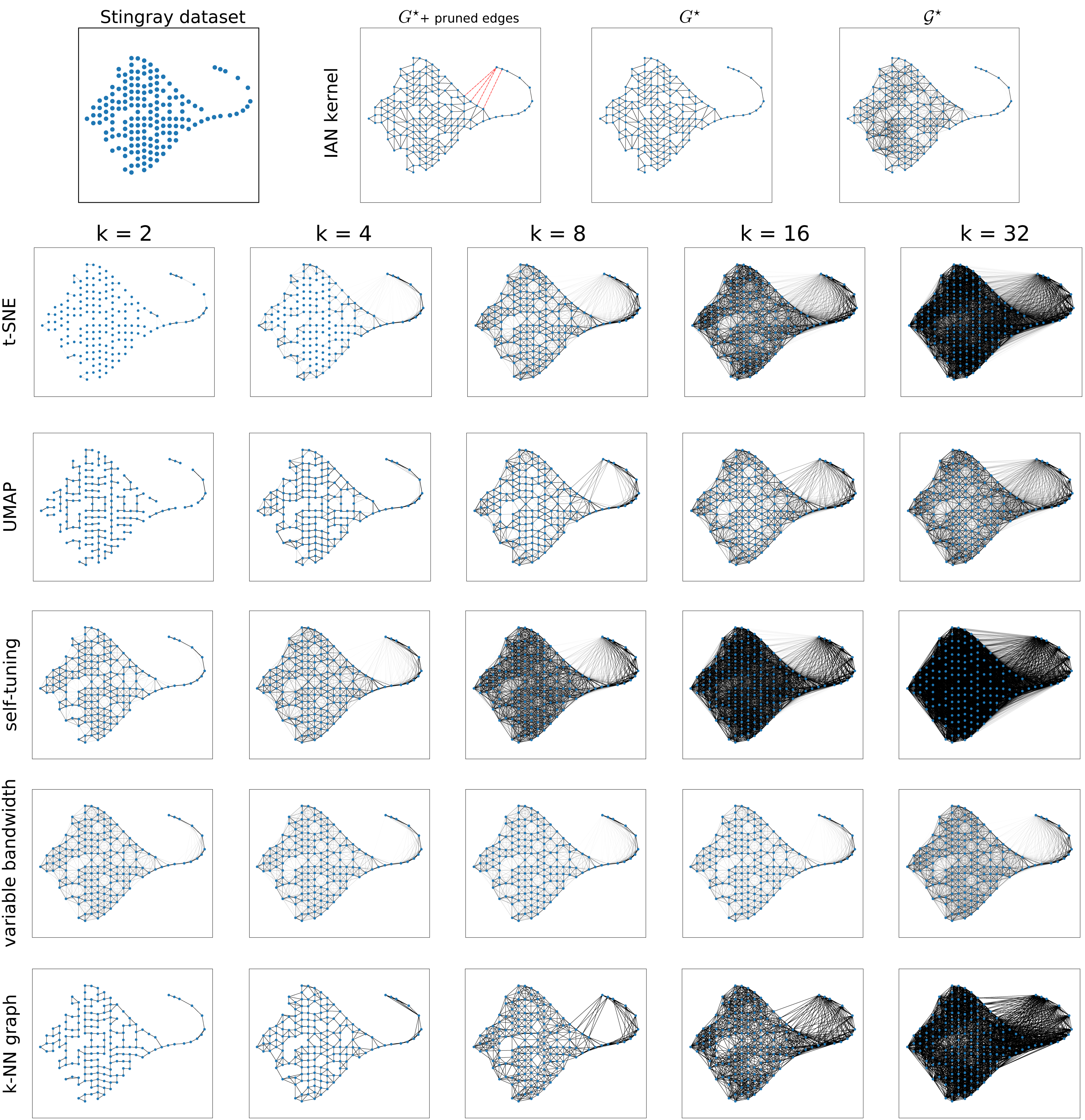}
  \caption{\emph{Top:} The stingray dataset and the converged graphs, $G^{\star}$ and $\mathcal{G}^{\star}$; pruned edges are shown in red. \emph{Bottom:} Other algorithms produce qualitatively different graphs depending on the neighborhood size parameter, $k$. All graphs shown are weighted (using a continuous kernel) except for the $k$-nearest neighbors graph (bottom row). Edge weights are visualized as the intensity of the line segments (each $w_{ij}$ is divided by the kernel value when $r_{ij}$ equals the scale, for a fair comparison across algorithms).
  }
  \label{fig:graphs-stingray}
\end{figure}

We now compare the data graphs obtained using our iterated adaptive neighborhoods (IAN) with those from other popular manifold learning methods. In Figure~\ref{fig:graphs-stingray}, a synthetic ``stingray'' dataset exhibits a transition of apparent dimension from 2 (body) to 1 (tail), a variation of the scenario explored in Figure~\ref{fig:manifold-probs}. Points were uniformly sampled, with 20\% deleted at random. 

Our converged, unweighted graph, $G^{\star}$ (top row in Figure~\ref{fig:graphs-stingray}), can be compared with the traditional $k$-nearest neighbors graph (bottom row), used in a variety of methods, including Isomap \citep{isomap}. In the latter, when $k$ = 2, the tail exhibits perfect connectivity, but the body is too sparse. If $k$ = 4, the body is more properly connected but the tail becomes overly connected, and ``folding'', or ``short-circuits'', start to appear. Finally, for $k \geq 8$, the connectivity is inappropriate as the tip of the tail connects directly to the body. In contrast, $G^{\star}$ manages to retain a minimally-connected tail while covering the body almost everywhere, creating appropriate edges across many of the sampling gaps (compare with the holes that remain in the $k$-NN graph with $k$ = 4, some of which are present even when $k$ = 8).

Our weighted graph, $\mathcal{G}^{\star}$, can be compared against methods that use a Gaussian-like kernel, and where each point has an individual scale. Some of these methods were described in section~\ref{section:kernels}: t-SNE \citep{tsne}, UMAP \citep{umap}, self-tuning \citep{zelnikmanor}, and variable bandwidth \citep{BGH1, BGH2}; their resulting connectivity can be visualized in Figure~\ref{fig:graphs-stingray}, where edges have intensity proportional to their weight.

\begin{figure}[h!]
  \centering
  \includegraphics[width=1\textwidth]{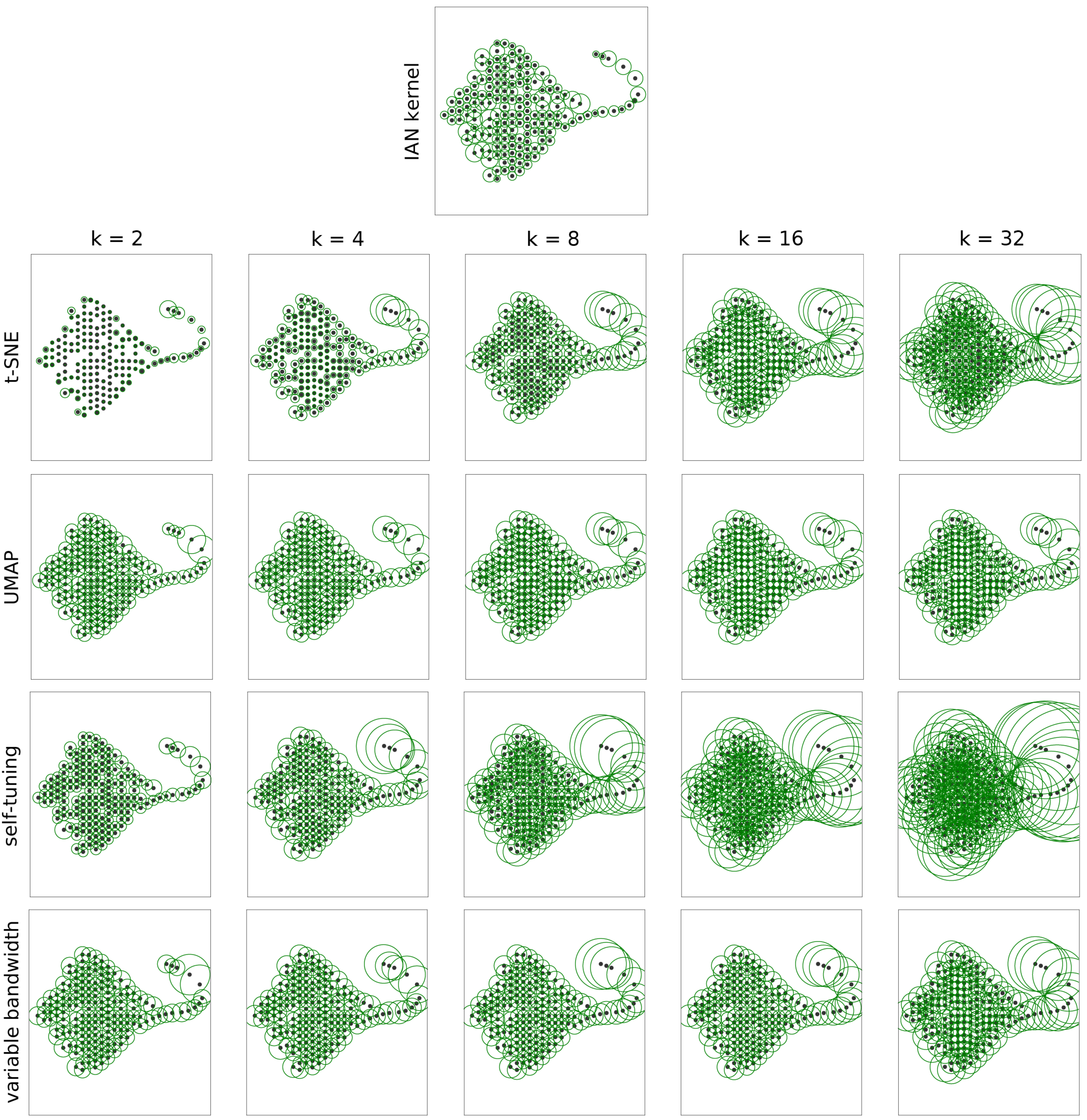}
 \caption{Individual scales obtained using our algorithm (top) compared to other methods (bottom table), as represented by their level sets for a (single-scale) kernel value of 0.75. 
  }
  \label{fig:scales-stingray}
\end{figure}

In Figure~\ref{fig:scales-stingray}, we visualize the individual scales resultant from each of these methods. Each $\sigma_i$ is represented, around each point $i$, as the level set corresponding to a (single-scale) kernel value of 0.75.
At the top, we see that the scales found by our kernel seem to nicely conform to the space between each point and its neighbors. Especially illuminating is what happens along the tail, where scales either ``expand'' or ``shrink'' so as to minimally cover the spaces between neighboring points; this illustrates what our scale optimization achieves. Among the other methods, with few exceptions, the scales seem to cover either too much (collapsing the tail on itself) or too little (leaving holes in the body). 

The weighted graphs in Figure~\ref{fig:graphs-stingray} reveal the result of the interaction between these individual scales (namely, the edge weights). Our $\mathcal{G}^{\star}$ (top right) manages to cover almost the entire body with edges, while keeping the tail minimally connected---in fact, resembling the unweighted version in $G^{\star}$, and therefore respecting the original curvature and reach. Other methods, in contrast, have a hard time achieving both things with a global value for $k$.
In t-SNE, the scales over the body are much small when $k$ $\leq$ 4, so its weighted graph looks too sparse; for $k$ $\geq$ 8, the scales over the tail become too large, and therefore strong edges appear, connecting it to the body.
In UMAP, the scales do not grow as much with increasing $k$, but at $k$ = 4 the body in the weighted graph is still too sparse, while for $k$ $\geq$ 8 the tail is strongly connected to the body.
With the self-tuning, scales seem to grow faster with $k$, while with variable bandwidth this growth is somewhat counteracted by the action of their global scale, $\epsilon$ (equation~\ref{eq:bgh}). In fact, the graph that most resembles our own $\mathcal{G}^{\star}$ is the one using the variable bandwidth kernel with $k$ = 2, the main difference being that the big sampling gap near the tip of the body is poorly connected, while in our case it is slightly overly connected (due to connections in $G^{\star}$ crossing that gap).

\section{Applications}\label{section:applications}

We now provide examples of application of our kernel to three different manifold learning tasks: dimensionality reduction, by means of a non-linear embedding algorithm; geodesic estimation, which typically finds application in computational geometry, vision, and graphics; and local intrinsic dimensionality estimation. 

\subsection{Low-dimensional embeddings}\label{section:low-dim-embeddings}

Dimensionality reduction is now ubiquitous in visualization of high-dimensional data. Several methods exist \citep{saul2006spectral,lee2007nonlinear, goldberg2009procrustes,vdmaaten-review}, and most of the non-linear methods are manifold-based \citep{LLE,isomap,roweis2001global,sne,eigenmaps,donoho2003hessian,zhang2004principal,lafon1,weinberger2006unsupervised,tsne,largevis,umap,moon2019phate}. Given a collection of points in high-dimensional space sampled from a low-dimensional manifold $\mathcal{M}$, the goal is to find a good parametrization for the data in terms of intrinsic coordinates over $\mathcal{M}$, which in turn can be used to produce a low-dimensional embedding.

In surveying the literature, it is common to find a heuristic, or a range of values, suggested for choosing the neighborhood size (see section~\ref{section:kernels}), but rarely do we see examples of the sensitivity of the results to that choice. In this section, we ran a few of the most popular methods using a wide range of values for the kernel scale parameter, $k$, and compared their results to those using our own kernel.

\begin{figure}[h!]
  \centering
  \includegraphics[width=1\textwidth]{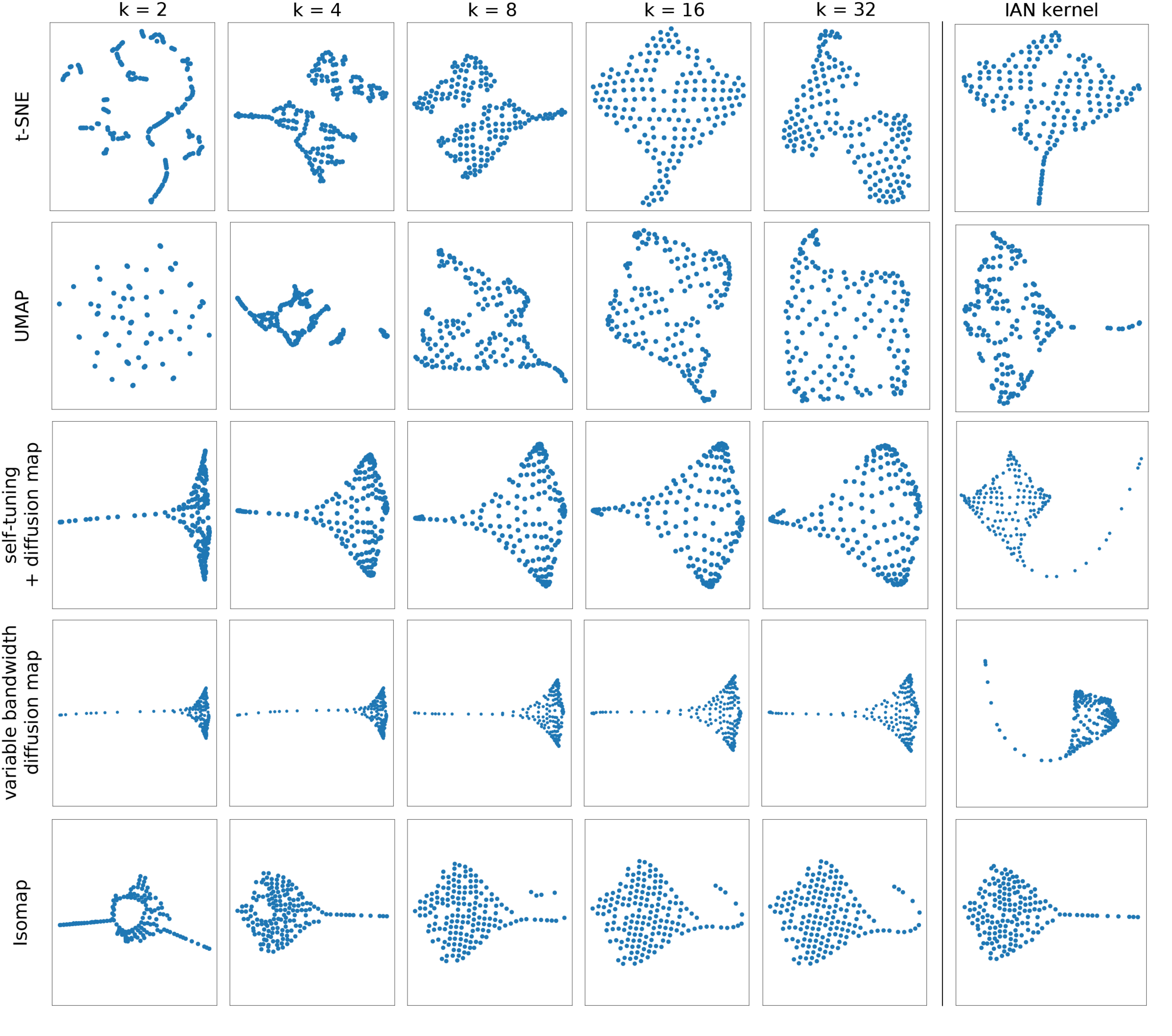}
  \caption{
  Running different embedding algorithms on the ``stingray'' dataset (see Figure~\ref{fig:graphs-stingray}).  Different choices of the neighborhood size, $k$, may produce qualitatively different results, depending on the algorithm. Running those same algorithms using the IAN kernel (right) typically gives a reasonable result. Refer to main text for details. 
  }
  \label{fig:embeds-stingray}
\end{figure}

We have limited our comparison to some of the embedding methods that use a neighborhood kernel and for which pairwise information is sufficient as input (i.e., do not require positional information): diffusion maps \citep{coifman2005diffmaps,lafon1}, Isomap \citep{isomap}, t-SNE \citep{tsne}, and UMAP \citep{umap}. 
As shown in Figures~\ref{fig:embeds-stingray}--\ref{fig:embeds-bentplane}, results can vary qualitatively depending on the choice of $k$. Five values were tested for each dataset, spanning a wide range of scales and different geometries. Next, we summarize each of these methods and their results.

\subsubsection{Diffusion maps + self-tuning kernel}

Diffusion maps are based on the spectral properties of the random walk matrix (normalized graph Laplacian) over the weighted data graph; integration over all paths in the graph makes diffusion distances, in principle, more robust to ``short-circuiting'' than graph geodesics. For better comparison with IAN, instead of the standard single-scale Gaussian kernel we use the self-tuning approach of \citet{zelnikmanor} from equation~\ref{eq:multiscale}. Our kernel was applied to diffusion maps by directly using $\mathcal{G}^{\star}$ as similarity matrix (weighted adjacency matrix). We use the diffusion map parameters $\alpha$ = 1 and $t$ = 1 \citep[cf.][]{lafon1}.

With the stingray dataset (Figure~\ref{fig:embeds-stingray}), we see that the fully-extended tail at $k$ = 2 becomes progressively more folded and compressed as $k$ increases. The body appears contracted at $k$ = 2, but expands with larger $k$. Using our own $\mathcal{G}^{\star}$, although we obtain excellent embeddings of both body and tail (right-most column), they are represented by separate sets of coordinates (two for the body, and a third for the tail), which happens due to the change in dimensionality.

Applying self-tuning to the spiral dataset (Figure~\ref{fig:embeds-spiral}), only $k$ = 2 and $k$ = 4 were able to prevent folding. The bent plane (Figure~\ref{fig:embeds-bentplane}) was more tolerant, with good results for all $k$ except 64, for which the plane remained folded. When using IAN, a good parametrization was obtained for both datasets.

\subsubsection{Variable bandwidth diffusion embedding}

We also tested a variant of diffusion maps using the variable bandwidth kernel of \citet{BGH1}, in which a distinct type of multiscale kernel is proposed, along with a specific normalization of the weighted graph Laplacian.
Because it computes an additional global scale, $\epsilon$, based on the individual scales, in order to apply our algorithm to this method we replaced the density estimates, $q_{\epsilon}$ (equation~\ref{eq:bgh}), with the inverse of our optimal scales. We used $\alpha$ = 0 and $\beta$ = -1/2, as recommended in \cite{BGH2}; eigenvectors were scaled by the square-root of the inverse of their respective eigenvalues \citep{saerens2004principal,noe2016commute}, following the implementation in \cite{bgh-github}. 

This method produced good embeddings for the stingray, especially for $k$ = 8 (Figure~\ref{fig:embeds-stingray}). For the spiral (Figure~\ref{fig:embeds-spiral}), using $k$ $\leq$ 8 caused some points to drift apart, and although it returned basically the original curve when $k$ = 16 or 32, a spectral algorithm such as this is expected to ``unroll'' the spiral, finding a good (1-D) parametrization of it. 
The same happened with the bent plane (Figure~\ref{fig:embeds-bentplane}), which could not be embedded into 2 coordinates for any choice of $k$. Using our scales, however, the algorithm managed to find appropriate parametrizations for all three datasets. 

\subsubsection{Isomap}

Isomap applies classical multidimensional scaling (MDS) to geodesic distances computed as shortest paths over a $k$-nearest neighbors graph (equation~\ref{eq:knn-graph}). Because the graph is unweighted, this method is particularly sensitive to the choice of $k$. Our kernel was applied to Isomap by directly replacing the $k$-NN graph with $G^{\star}$.

With the stingray (Figure~\ref{fig:embeds-stingray}), Isomap produced a good embedding with $k$ = 4. The result with $k$ = 2 was completely wrong (an additional tail appears), and with $k$ = 8 the tip of the tail was disconnected. With $k$ = 16 and $k$ = 32, it essentially returned the original data, without any dimensionality reduction. Our $G^{\star}$ improved on the result of $k$ = 4 by making the points in the body more uniformly spread. 

The spiral (Figure~\ref{fig:embeds-spiral}) was properly embedded (1-dimensional) only when $k \leq 4$. With the bent plane (Figure~\ref{fig:embeds-bentplane}), good results were obtained for $k$ between 4 and 16, but $k$ = 2 produced 1-dimensional curves, and $k$ = 64 did not completely unfold it. Our $G^{\star}$ produced the correct mapping in either case.

\subsubsection{t-SNE and UMAP}

t-SNE and UMAP are related methods that have gained popularity in recent years \citep{tsne-vs-umap}. Both compute similarities between data points using individual scales based on $\log_2 k$ (section~\ref{section:kernels}), and adopt a secondary kernel for computing similarities between embedded points: t-SNE uses a Student t-distribution (Cauchy kernel), while UMAP uses an non-normalized variant requiring a hyperparameter, $\texttt{min\_dist}$. In t-SNE, embedding coordinates are initialized at random, while UMAP adopts the strategy of refining an initial spectral embedding. 
Both then optimize their embeddings by running gradient descent on an information-theoretic cost function between similarities in input space \textit{vs.} embedded space: t-SNE minimizes the KL-divergence; UMAP uses a variant of cross-entropy. 

Alternative initializations are typically used with t-SNE (e.g., PCA) to improve results \citep{kobak2019art,linderman2019fast,kobak2021initialization}; in our experiments, for better comparison with UMAP, we used a spectral embedding initialization computed from its own symmetrized similarity matrix (equation~\ref{eq:tsne-sym}).
The IAN kernel was applied to t-SNE by replacing the individual scales (equation~\ref{eq:tsne-kernel}) with those in $\bm{\sigma}^{\star}$; with UMAP, because a different kernel function is used, we directly replaced the weighted graph (with adjacencies given by $U_{ij}$ in equation~\ref{eq:umap-sym}) with $\mathcal{G}^{\star}$.

\begin{sloppypar}
We executed t-SNE assigning the various $k$ values to the perplexity parameter, leaving the remaining parameters to their defaults in the scikit-learn implementation \citep{scikit-learn}. We used the Barnes-Hut method \citep{van2014accelerating} for the cylinder dataset; and the ``exact'' method for all others.
In UMAP, the $\texttt{n\_neighbors}$ parameter was set to $k$, with remaining parameters using default values (in particular, $\texttt{min\_dist}$ = 0.1). Because of the stochastic nature of both algorithms (even when using a fixed initialization), different runs will produce slightly different results. Therefore, in order to avoid ``cherry-picking'', both algorithms were executed a single time, using the same random seed.
\end{sloppypar}

Results for the stingray (Figure~\ref{fig:embeds-stingray}) were quite analogous between the two algorithms: both produced artificial clustering for $k$ $\leq$ 8, while for $k$ $\geq$ 16 the tail began to fuse with the body. The gaps in sampling within the body were accentuated by both algorithms, even at $k$ = 32, where we see a big hole in the UMAP embedding; in t-SNE, it almost breaks into two pieces (despite the large neighborhood size). This example is illustrative of how much an embedding algorithm based on attractive \textit{vs.} repulsive forces can end up exaggerating nonuniform sampling.

\begin{figure}[h!]
  \centering
  \includegraphics[width=1\textwidth]{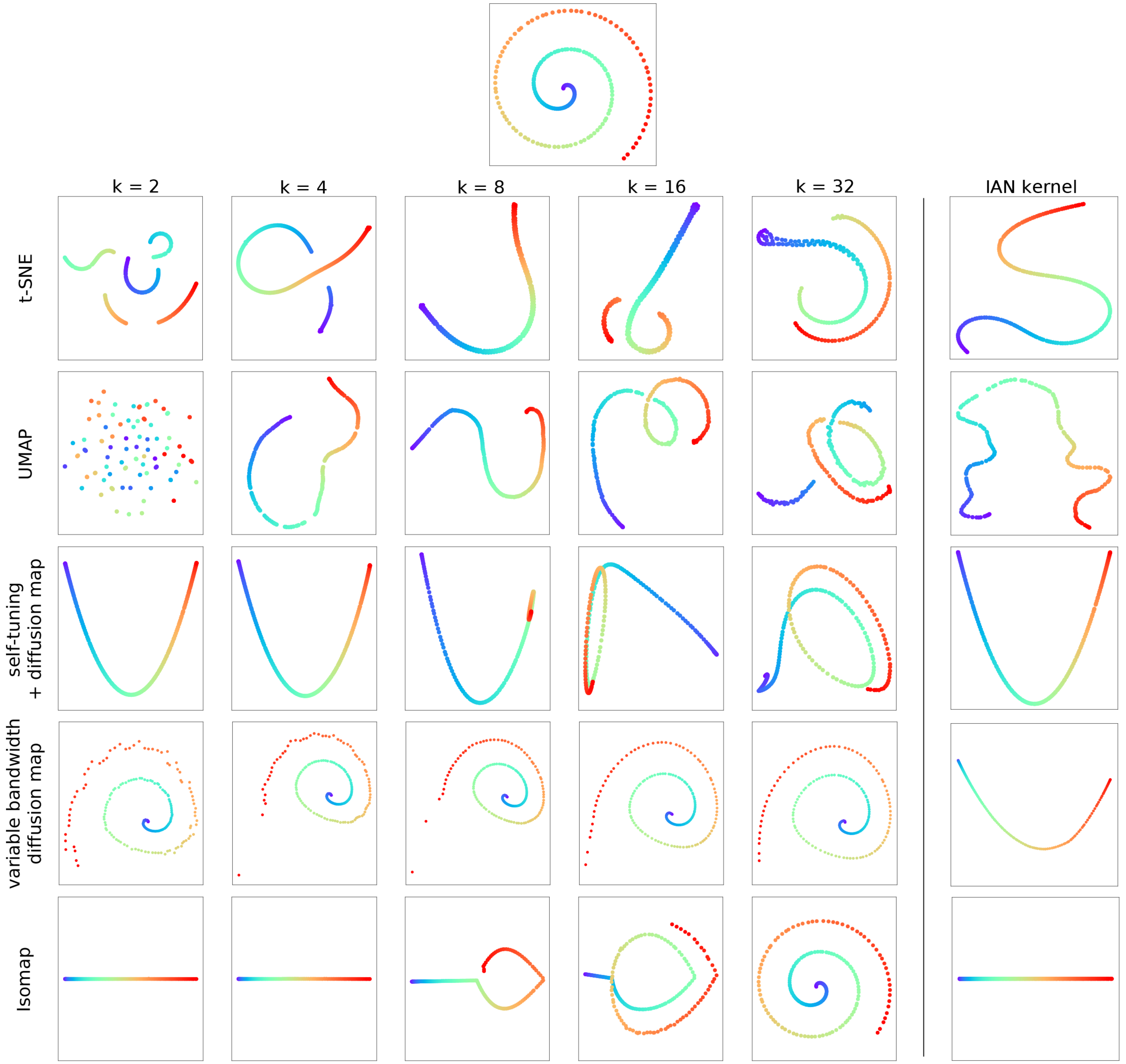}
  \caption{
  Running different embedding algorithms on the spiral dataset (top), in which points are sampled from a unit-speed parametrized Archimedean spiral. Different choices of the neighborhood size, $k$, may produce qualitatively different results, depending on the algorithm. Running those same algorithms using the IAN kernel (right) typically gives a reasonable result. Refer to main text for details. 
  }
  \label{fig:embeds-spiral}
\end{figure}

The spiral (Figure~\ref{fig:embeds-spiral}) was disconnected by t-SNE for all values of $k$ except 8. UMAP produced reasonable results for $k$ between 4 and 8; however, for $k$ = 2 a multitude of clusters was obtained, and when $k$ $\geq$ 16 the curve twisted over itself. Using our kernel (right column) produced a connected, non-self-intersecting curve. Neither algorithm was capable of returning a good arc-length parametrization of the spiral, however.

\begin{figure}[h!]
  \centering
  \includegraphics[width=1\textwidth]{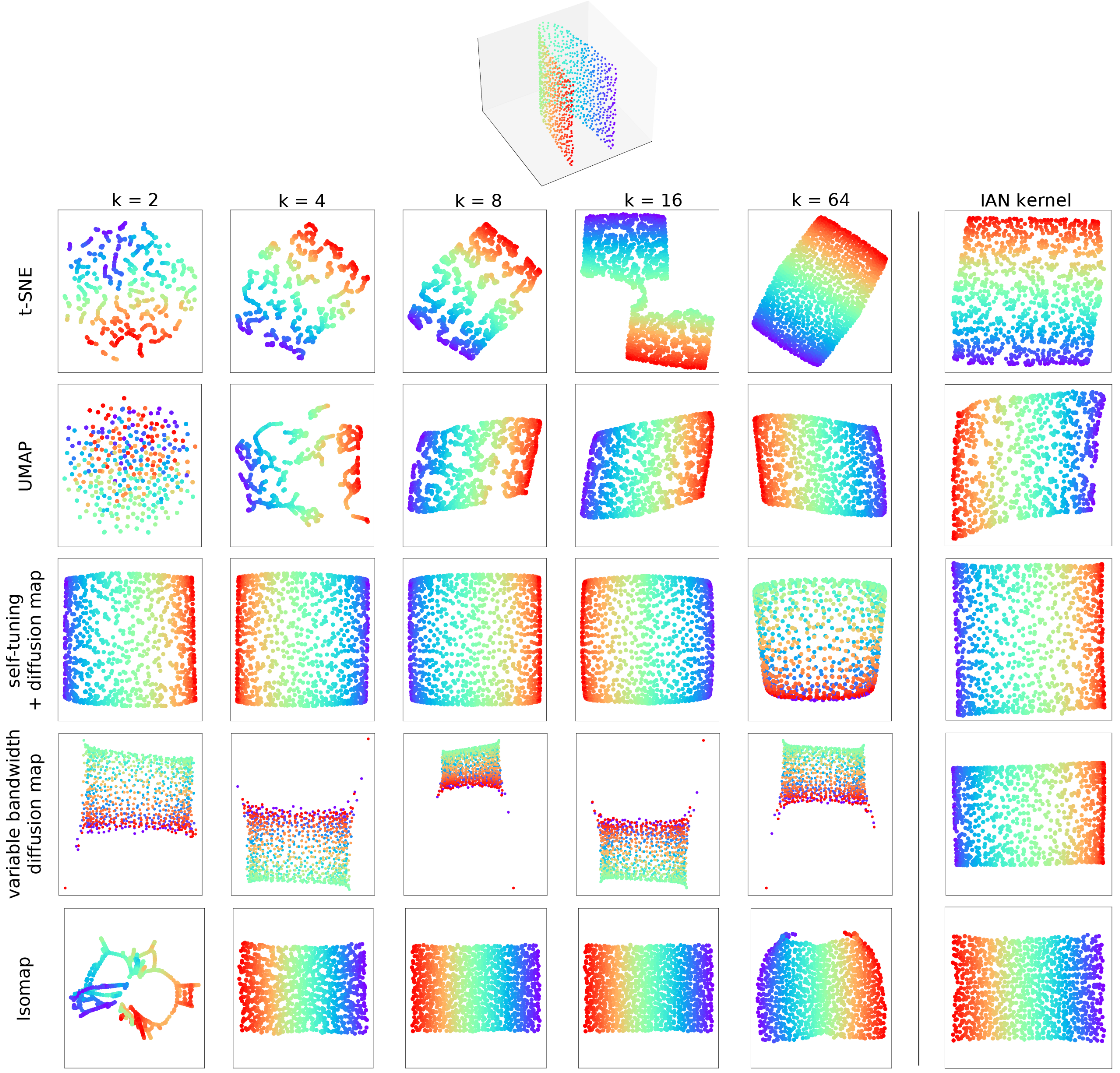}
  \caption{
  Running different embedding algorithms on the bent plane dataset (top), generated by extending a unit-speed parametrized catenary curve into two dimensions. Different choices of the neighborhood size, $k$, may produce qualitatively different results, depending on the algorithm. Running those same algorithms using the IAN kernel (right) typically gives a reasonable result. Refer to main text for details. 
  }
  \label{fig:embeds-bentplane}
\end{figure}

With the bent plane (Figure~\ref{fig:embeds-bentplane}), although both algorithms succeeded in unfolding it, t-SNE was only able to produce a fully two-dimensional plane (with no gaps) when setting $k$ = 32 (not shown) or 64, while UMAP required $k$ $\geq$ 16. Both gave reasonable results using our kernel.

\subsubsection{A higher dimensional example}

Because all of the examples above have $d \leq 2$, we also tested our kernel when applied to a higher dimensional manifold, namely a 5-dimensional cylinder ($\mathbb{R}^1 \times S^4$) with radius 1 and length 3, sampled uniformly at random ($N$ = 8403, ambient space $\mathbb{R}^6$). On the other hand, here we used a pure, connected manifold with no bottlenecks and low curvature in order to simplify interpretation.

\begin{figure}[h!]
  \centering
  \includegraphics[width=1\textwidth]{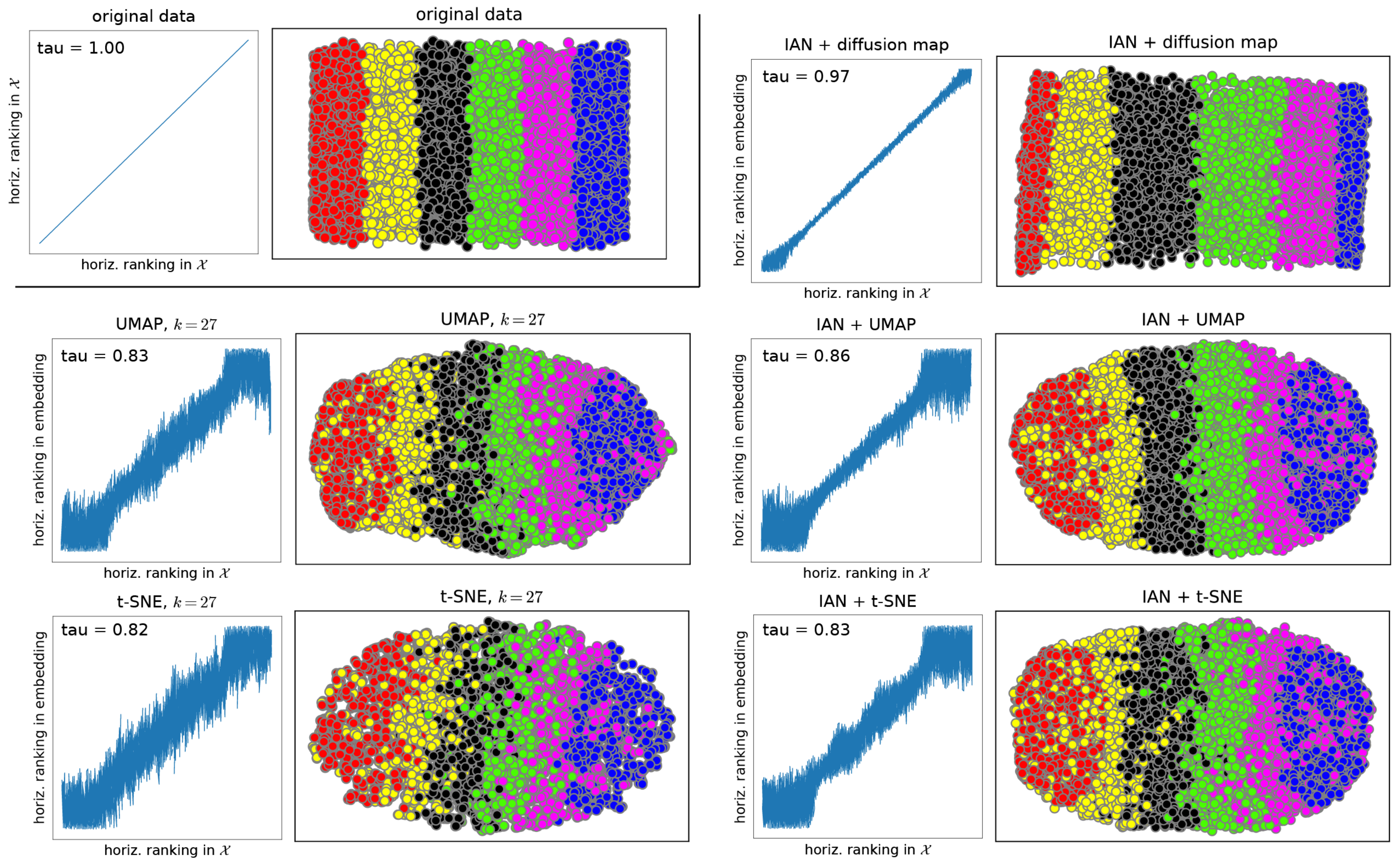}
  \caption{
  Performance of different embedding algorithms on a 5-dimensional cylinder ($\mathbb{R}^1 \times S^4$) sampled uniformly at random ($N$ = 8403, ambient space $\mathbb{R}^6$). Top left: original data, $\mathcal{X}$, projected onto first 2 coordinates (points colored according to their position along the cylinder's long axis). Other plots show embeddings using different kernels and/or algorithms. The resulting degree of mixing of the original color labels indicates the quality of the embedding. A quantitative assessment (plots to the left of each embedding) was computed as the rank correlation coefficient, tau (see main text), between the ranking (positional order) of each point along the horizontal axis in the original {\it vs.} embedded spaces (a value closer to 1 indicates fewer exchanges in the original order). Use of the IAN kernel produced similar or better results with both t-SNE and UMAP ($k$ = 27 was set based on the mean degree of $G^{\star}$, compatible with $d$ = 5). Diffusion maps resulted in very little mixing except near the boundaries.
  }
  \label{fig:embeds-cylinder}
\end{figure}

Figure~\ref{fig:embeds-cylinder} shows two-dimensional embeddings obtained by applying our kernel to different embedding algorithms. Although all correctly produced an oblong, various degrees of mixing of the original color labels were observed, which can be used to qualitatively indicate the quality of the embedding schemes. A quantitative assessment was computed as the rank correlation coefficient, or Kendall's tau \citep{kendall1948rank, knight1966computer} between the ranking (positional order) of each point along the main axis in the original {\it vs.} embedded spaces.

Both t-SNE and UMAP produced similar or better results when using the IAN kernel (we set $k$ = 27 based on the mean degree found in $G^{\star}$, compatible with $d$ = 5; results were robust to this particular choice). Despite their current popularity \citep[e.g.,][]{wattenberg2016use,arora-tsne,chan2018t,dimitriadis2018t,tsne-vs-umap,kobak2019art,fujiwara2021fast,kobak2021initialization,wang2021understanding}, produced considerably jittered outputs, however, implying that the original neighborhoods were not preserved. This appears to be caused by an attempt to reproduce the spherical shape of the cylinder's base along the main axis, so different ``slices'' ended up projected on top of one another. However, UMAP produced jittered results even when set to return 6 components (as in the original space) instead of 2.

Diffusion maps using IAN resulted in little mixing except near the boundaries, so neighborhoods were better preserved. Running it with either self-tuning or variable-bandwidth kernels using $k$ = 27 gave comparable results; Isomap also produced excellent results, with tau = 0.98 (not shown).

\FloatBarrier


\subsection{Geodesic computation}

Using the unweighted graph, $G^{\star}$, one may immediately compute graph geodesics (shortest paths using distances in ambient space as edge lengths) to estimate the true geodesics over $\mathcal{M}$ \citep{isomap}. The latter are likely to be underestimated by the former when sampling is sparse \citep{bernstein2000graph}, even when the graph connectivity is correct, e.g., due to curvature (cf. section~\ref{section:max-curvature}). It seems a good idea, then, to incorporate the continuous kernel values present in its weighted counterpart, $\mathcal{G}^{\star}$, as a means to possibly improve geodesic estimation.

We propose to use the heat method for geodesic computation of \citet{crane-geoheat}. It consists in solving the Poisson equation to find a function, $\phi$, whose gradient follows a unit vector field, $\bm{X}$, pointing along geodesics; $\bm{X}$ can be obtained by normalizing the temperature gradient, $\nabla \bm{u}$, due to a diffusion process in which heat, $\bm{u}$, is allowed to diffuse for a short time. Although this method is tailored to applications where positional information and dimensionality are known (in particular, surfaces in $\mathbb{R}^3$), here we apply it to $\mathcal{G}^{\star}$, since discrete versions of the operators used (Laplacian, gradient, and divergence) can be readily defined on a weighted graph \citep[see][]{desquesnes2013eikonal}.

\begin{figure}[h!]
  \centering
    \includegraphics[width=1\textwidth]{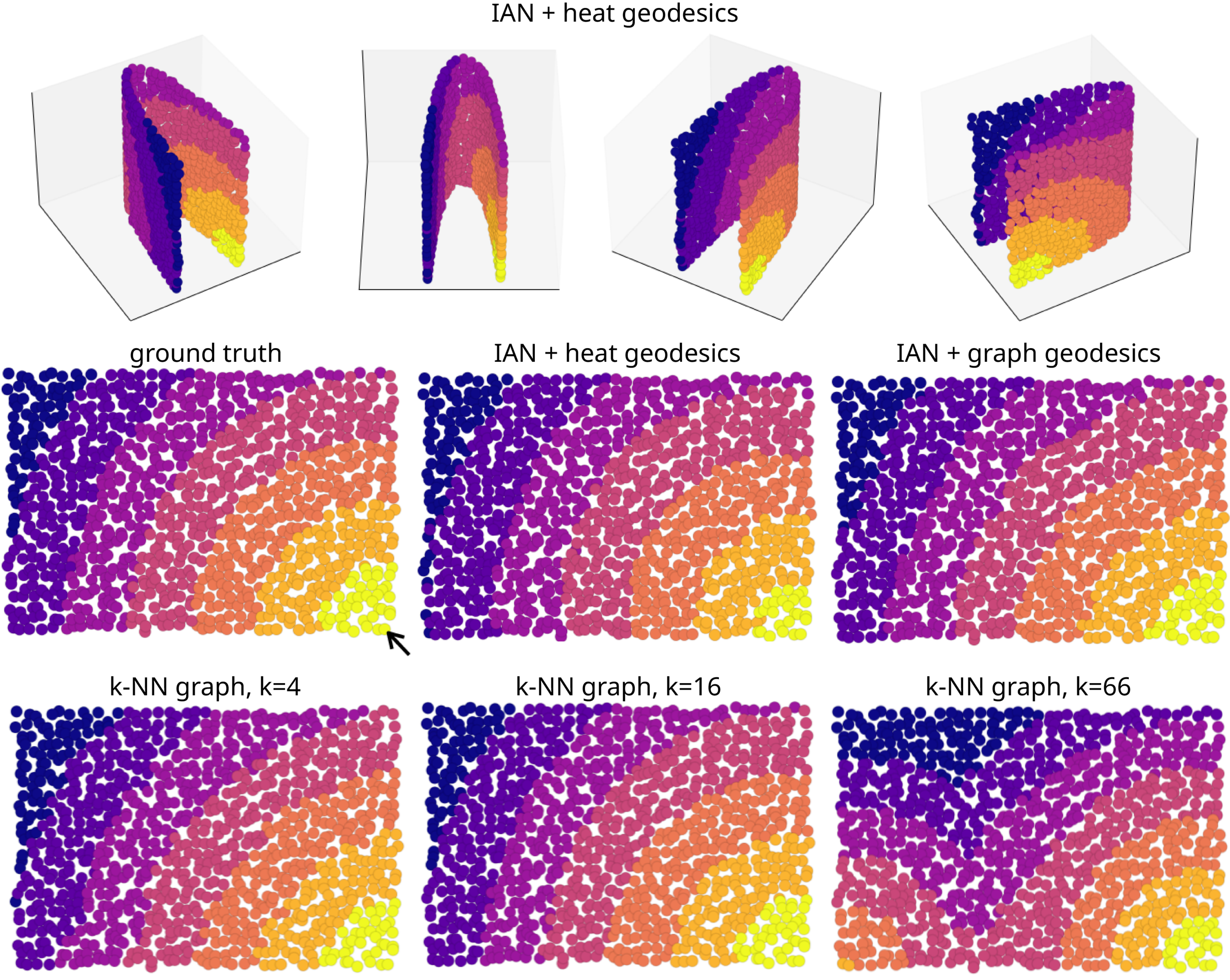}
  \caption{
  Geodesic estimation for the bent plane from Figure~\ref{fig:embeds-bentplane}; yellow points are closer to the source (marked with an arrow in the ground truth plot).
  \emph{Top:} different views of the data in 3-D, with points colored according to the heat geodesics computed from $\mathcal{G}^{\star}$. 
  \emph{Middle:} Geodesics displayed on an unbent version of the dataset: heat geodesics approximate well the true geodesics over $\mathcal{M}$, and graph geodesics computed from $G^{\star}$ follow closely.
  \emph{Bottom:} graph geodesics computed from $k$-NN graphs using different choices of $k$; choosing $k$ = 16 gives near-perfect results, but $k$ = 4 shows distortions, and $k$ = 66 misses completely.
  }
  \label{fig:geo-bentplane}
\end{figure}

\begin{figure}[h!]
\centering
  \includegraphics[width=1\textwidth]{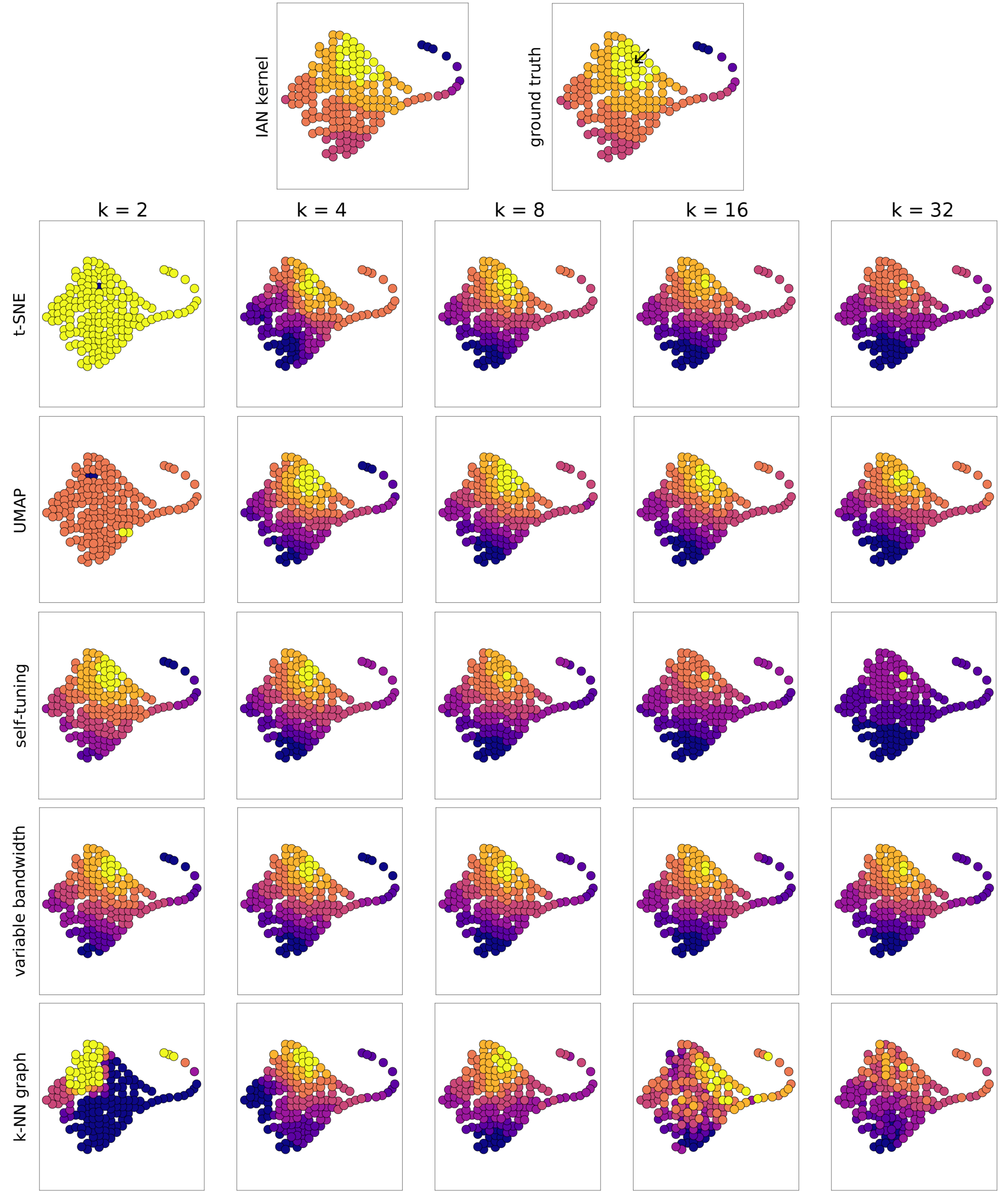}
\caption{Geodesics estimated using the heat method applied to $\mathcal{G}^{\star}$ are close to the ground truth (top). Other kernels yield suboptimal results for most choices of $k$ (bottom); in particular, notice how the tip of the tail is usually inferred to be closer than it should (due to its being directly connected to the body in the underlying graph, cf. Figure~\ref{fig:graphs-stingray}). Yellow points are closer to the source (marked with an arrow in the ground truth plot).
}
\label{fig:geo-stingray}
\end{figure}

Despite using pairwise information only, our method produces reasonable estimates, as shown in Figures~\ref{fig:geo-bentplane} and \ref{fig:geo-stingray}. To understand why, notice that IAN indirectly solves for a weighted graph for which a random walk starting at node $i$ has a higher probability of reaching a node in its discrete neighborhood, $\mathcal{N}(i)$, than any other non-neighboring node. Given that random walks are closely related to diffusion over a graph, one should expect $\mathcal{G}^{\star}$ to be able to provide reasonable information about how a diffusion process propagates over $\mathcal{M}$. In other words, the Laplacian obtained from $\mathcal{G}^{\star}$ should be a good approximation of a continuous operator over $\mathcal{M}$---this is empirically confirmed by our results.

In Figure~\ref{fig:geo-bentplane}, heat geodesics computed from $\mathcal{G}^{\star}$ for the bent plane dataset approximate well the true geodesics over $\mathcal{M}$, and graph geodesics obtained from $G^{\star}$ follow closely. Comparison with those from a naive $k$-NN graph illustrates that the choice of $k$ is critical (compare with the bottom row of Figure~\ref{fig:embeds-bentplane}).
In Figure~\ref{fig:geo-stingray}, we compare the results using weighted graphs from various kernels on the stingray dataset; interestingly, heat geodesics computed from $\mathcal{G}^{\star}$ hold reasonably well even when facing a continuous change in dimensionality. (The diffusion time parameter used by the heat method was optimized for each dataset.)

\FloatBarrier


\subsection{Local dimensionality estimation}\label{section:dim-estimation}

Intrinsic dimensionality (ID) estimation is tightly associated with dimensionality reduction tasks, especially in manifold learning, where knowledge of $d$ can help, among others, to determine the appropriate number of embedding dimensions. Informally, ID may be seen as the minimum number of parameters required to accurately describe the data. In the context of manifold learning, it is typically equivalent to the topological dimension of $\mathcal{M}$ (e.g., a general space curve has dimensionality 1 since it requires a single parameter, arc length).

There are many different ways to estimate it \citep{dimreview, camastra2016intrinsic}; global approaches are typically divided into two. The first group is based on some variant of PCA  \citep[e.g.,][]{fukunaga, little2017multiscale}, and use the number of significant eigenvalues to infer dimensionality; these may be applied globally or by combining local estimates. The second group of methods, termed geometric (or fractal, when a non-integer ID is computed), exploit the geometric relationships in the data, such as neighboring distances. Some are based on estimating packing numbers \citep{kegl2002packing} or on distances to nearest-neighbors \citep{trunk1976statistical, pettis1979intrinsic,verveer1995evaluation,costa2005estimating,facco2017estimating,block2021intrinsic}.

Among the most popular are the \emph{correlation dimension} methods \citep{camastra2002estimating,correlationdim, heindimensionality}, a variant of which has been specifically applied in the context of determining an appropriate kernel width for manifold learning \citep[see][]{laplaciantomography,BGH1,theiss}. The dimension is computed as the slope of a log-log plot of the number of neighboring points {\it vs.} neighborhood radius (see section~\ref{section:kernels}). A recent variation is \cite{kleindessner2015dim}; others cover the difficult case of high ID \citep{camastra2002estimating,rozza2012novel}.

In our scenario, since we do not assume a pure manifold (section~\ref{section:subtleties}), we focus on local (i.e., pointwise) ID estimation approaches, namely those in which dimension is estimated within a neighborhood around each data point \citep[e.g.,][]{farahmand2007manifold, he2014intrinsic}. This notion can be formalized as the \emph{local Hausdorff dimension} \citep{young1982dimension,camastra2016intrinsic}, and a global estimate is typically found by averaging over local values.

A popular approach is the maximum likelihood estimator (MLE) of \citet{levina2004maximum}, which computes local dimension based on $k$-nearest neighbors: 
\begin{equation}\label{eq:MLE}
    \hat{m}_k(\bm{x}_i) = \left ( \frac{1}{k-1}\sum_{j=1}^k \log \frac{T_k(\bm{x}_i)}{T_j(\bm{x}_i)} \right )
\end{equation}
where $T_j(\bm{x}_i)$ denotes the distance between $\bm{x}_i$ and its $j^{\mathrm{th}}$ nearest neighbor. We shall use this method in our experiments, in which we compute a final $m_k(\bm{x}_i)$ by averaging $\hat{m}_k(\bm{x}_i)$ over $i$'s neighbors in order to reduce the variance of the local estimates (in the original, this is done over all data points).

Notice that our kernel can be readily used with this method by simply replacing the $k$-NN graph with $G^{\star}$, therefore summing over nodes in the neighborhood $\mathcal{N}(i)$ instead of over the $k$ nearest. Additionally, we propose a correlation dimension-based method that allows for local estimates. We describe it next, then compare its results with those from the MLE method.

\subsubsection{Algorithm: Neighborhood Correlation Dimension}

Our proposed method is adapted from the approach from \citet{heindimensionality} \citep[also used in][]{laplaciantomography,theiss,BGH1}, where an estimate of correlation dimension is obtained using a general kernel. It consists in computing a curve, $Z(\sigma)$, over all pairwise kernel values (e.g., a Gaussian) at different values of the scale parameter $\sigma$:
\begin{equation}\label{eq:avgsumZ}
    Z = \sum_{i=1}^N \sum_{j=1}^N\exp{\frac{-\left\| \bm{x}_i - \bm{x}_j \right\|^2}{2\sigma^2}}.
\end{equation}
As in \cite{laplaciantomography} (and analogous to equations~\ref{eq:meanvaluegauss1}--\ref{eq:meanvaluegauss}), by assuming that for small values of $\sigma$ the manifold $\mathcal{M}$ looks locally like its tangent space, $\mathbb{R}^d$, we have
\begin{equation}
    Z \approx \frac{N^2(\sqrt{2\pi}\sigma)^d}{\mathrm{vol}^2(\mathcal{M})},
\end{equation}
which, after taking the logarithm, yields
\begin{equation}\label{eq:singer-log}
    \log Z \approx d \log \sigma + \log \frac{N^2(2\pi)^{d/2}}{\mathrm{vol}(\mathcal{M})},
\end{equation}
so the slope of $\log{Z} \times \log{\sigma}$ can be used to estimate the global dimensionality of the manifold, $d$. To do so, one typically looks for a region where this slope is most stable, i.e., the curve is approximately linear. Automated ways of finding the slope of such a region are: by linear regression of the middle portion of the curve \citep{heindimensionality} or by taking a point of maximum of $Z'(\sigma)$ \citep{BGH1,theiss}.

However, because we assume that intrinsic dimension may vary over $\mathcal{M}$, global averages cannot work in general. Moreover, nonuniform density, curvature, or multiple connected components may all create multiple peaks for $Z'(\sigma)$, so inspection of the log-log plot cannot be automated. 

Therefore, we modify this approach to use individual $Z_i(\sigma)$ curves for each data point $\bm{x}_i$. To keep the summation local, points are restricted to those in the neighborhood of $i$ in $G^{\star}$. Here, it is advantageous to work with an extended neighborhood (e.g., by also including neighbors-of-neighbors) due to the theoretical limit to the value of the dimension $d$ that can be accurately estimated given a set of $N$ points \citep{eckmann1992fundamental}, namely $d < 2\log_{10} N$. In fact, if $N$ is large compared to $d$, even additional hops away from $i$ may be considered. Because such extension is done by following edges in $G^{\star}$ (as opposed to naively expanding a ball in $\mathbb{R}^n$), we may thus obtain a larger (approximately tubular) neighborhood around $\bm{x}_i$ without ever leaving the manifold. We denote such a neighborhood $\mathcal{N}'(i)$, as opposed to the immediate neighborhood $\mathcal{N}(i)$; throughout this section, both will include the node $i$ itself.

Our algorithm involves the following steps:
\begin{enumerate}
\item For each data point $\bm{x}_i$ and its extended neighborhood, $\mathcal{N}'(i)$, define $Z_i$ as
\begin{equation}\label{eq:localZi}
        Z_i(\sigma) = \sum_{j \in|\mathcal{N}'(i)|}{\exp{\frac{-\left\| \bm{x}_i - \bm{x}_j \right\|^2}{2\sigma^2}}}.
\end{equation}
\item Analogous to equation~\ref{eq:singer-log}, by taking the logarithm we have that the slope of the $\log{Z_i} \times \log{\sigma}$ curve, i.e.,
\begin{equation}
Z'_i(\sigma) \stackrel{\text{def}}{=} \frac{\mathrm{d} \log{Z_i} }{\mathrm{d} \log{\sigma}},
\end{equation}
is an estimate of $d_i$, the dimension around $\bm{x}_i$, as a function of $\sigma$. Computationally, it is desirable to use the closed-form expression, for accuracy:
\begin{equation}\label{eq:slopei-closed}
Z'_i(\sigma) = 
\frac{\sum_{j=1}^{|\mathcal{N}'(i)|}{\left\| \bm{x}_i - \bm{x}_j \right\|^2\exp{\frac{-\left\| \bm{x}_i - \bm{x}_j \right\|^2}{2\sigma^2}}}}
{\sigma^2\sum_{j=1}^{|\mathcal{N}'(i)|}{\exp{\frac{-\left\| \bm{x}_i - \bm{x}_j \right\|^2}{2\sigma^2}}}}.
\end{equation}
\item A region of stability of $Z'_i$, i.e., a local maximum, is then an estimate of the dimension around $\bm{x}_i$.
\setcounter{dim-method}{\value{enumi}}
\end{enumerate}

A local maximum (``peak'') in $Z'_i(\sigma)$ can be interpreted as follows: as a ball around $\bm{x}_i$ is expanded, the rate at which neighbors are seen has stopped increasing and must decrease with larger $\sigma$, since no additional neighbors can be found after the ball encompasses all points in $\mathcal{N}'(i)$. Underlying is the assumption that $\mathcal{N}'(i)$ is sufficiently representative of the manifold around $\bm{x}_i$. I.e., if neighbors are approximately uniformly distributed and dimensionality is constant within it, then $Z'_i$ should remain constant over some appreciable range of $\sigma$, whence the notion of ``stability''.

Even though we work with a subset of $\mathcal{X}$, there may still be multiple maxima in $Z'_i$, e.g., when the neighbors of $\bm{x}_i$ are far from uniformly distributed around it. So, operationally, we use the global maximum of $Z'_i$, as this takes into account the information given by the majority of neighboring points. Now, because ${Z_i}\rightarrow 1$ as $\sigma \rightarrow 0$, and ${Z_i}\rightarrow N$ as $\sigma \rightarrow \infty$, the slope of $\log{Z_i}$ must approach 0 at both extremes, thus the global maximum of $Z'_i$ must also be a relative one (a ``peak'').

We now proceed to avoid boundary effects by \emph{re-centering neighborhoods}. The boundary, $\partial \mathcal{M}$, of a $d$-dimensional manifold (when present) has dimensionality $d-1$ \citep{intromanifolds}. The correlation integral approach often fails for these---it typically returns $d/2$ for points in $\partial \mathcal{M}$, since they have roughly half the number of neighbors compared to interior points. For the same reason, it tends to also underestimate $d$ for points near the boundary. Since we work locally over a graph, we can regularize the computation by moving the focus to a more central, nearby point (thus regularizing over sampling artifacts as well):
\begin{enumerate}
    \setcounter{enumi}{\value{dim-method}}
    \item Letting $\mathcal{N}(i)$ be the set of adjacent nodes to $i$ in $G^{\star}$ and including $i$ itself, define $\bar{\iota}$ as the node $j \in \mathcal{N}(i)$ with smallest median squared distance to all points in the extended neighborhood $\mathcal{N}'(i)$:
 \begin{equation}\label{eq:recentering}
     \bar{\iota} = \mathrm{argmin}_{j\in\mathcal{N}(i)} \mathrm{median}\left \{\|\bm{x}_j - \bm{x}_l\|^2), \forall l \in \mathcal{N'}(i) \right \}.
 \end{equation}
Thus $\bar{\iota}$ is, in effect, the most central node in $i$'s immediate neighborhood\footnote{Since we know $G^{\star}$, graph-theoretical quantities such as shortest-path betweenness centrality \citep{betweenness1,betweenness2} may also be used here.}.
 \item Use $\bar{\iota}$ as the point from which kernel values are computed for $Z_i(\sigma)$ by replacing $\bm{x}_i$ with $\bm{x}_{\bar{\iota}}$ in equation~\ref{eq:localZi}, thereby shifting the center of estimation of $d_i$. This assumes that the dimension does not change abruptly across neighboring points. Denote the resulting estimate by $\hat{d}_i$.
    \item As with the MLE method (section~\ref{section:dim-estimation}), we may obtain a smoother estimate, $\hat{d}'_i$, by averaging over immediate neighbors in $\mathcal{N}(i)$:
    \begin{equation}
        \hat{d}'_i = \frac{1}{|\mathcal{N}(i)|} \sum_{j \in \mathcal{N}(i)} \hat{d}_j.
    \end{equation}
 \setcounter{dim-method}{\value{enumi}}
\end{enumerate}

Finally, recall from section~\ref{section:stat-pruning} that we also obtain a degree-based estimate, $\tilde{d}_i$, when computing volume ratios (equation~\ref{eq:estimate-dim}); we can use this information to further improve our results. A final estimate, $d_i^{\star}$, is then obtained as follows:

\begin{enumerate}
    \setcounter{enumi}{\value{dim-method}}
    \item To avoid overestimating the true dimension, compute an average $\tilde{d}'_i$ over $\mathcal{N}(i)$ as
    \begin{equation}
        \tilde{d}'_i = \frac{1}{|\mathcal{N}(i)|} \sum_{j \in \mathcal{N}(i)} \left \lfloor \tilde{d}_j \right \rfloor = \frac{1}{|\mathcal{N}(i)|} \sum_{j \in \mathcal{N}(i)} \left \lfloor \log_2 \mathrm{deg}(j) \right \rfloor.
    \end{equation}
    \item Compute the optimal estimate, $d_i^{\star}$, as
    \begin{equation}\label{di-star}
        d_i^{\star} = \max \left \{ \hat{d}'_i, \tilde{d}'_i\right \}.
    \end{equation}
\end{enumerate}
Application of this technique and comparison with other methods are given next.

\subsubsection{Experimental results}

\begin{figure}[h!]
\centering
  \includegraphics[width=1\textwidth]{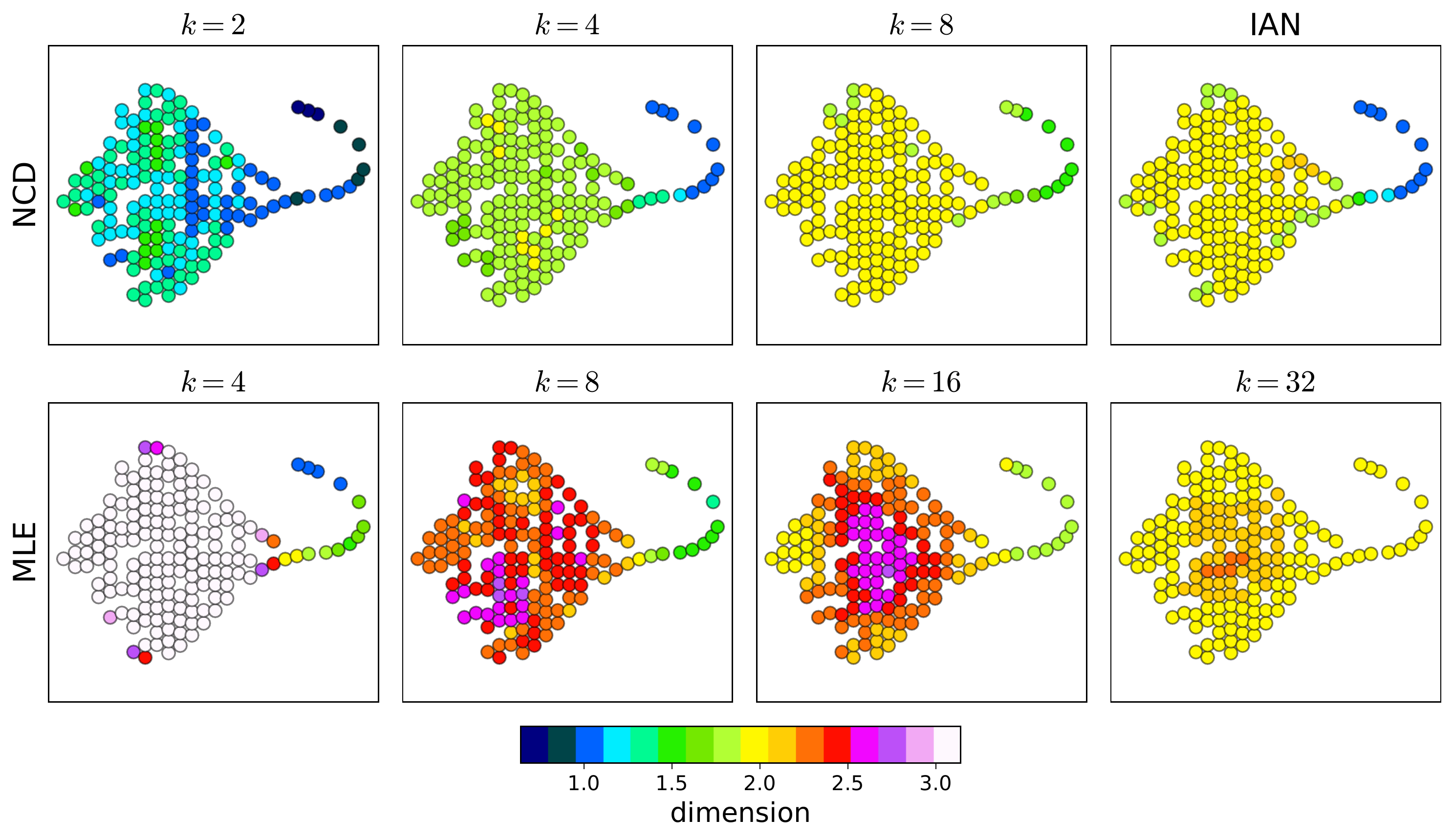}
\caption{Estimation of local intrinsic dimension on the stingray dataset. Top row shows results for our neighborhood correlation dimension (NCD) algorithm using $k$-NN graphs with various $k$ and using adaptive neighborhoods from $G^{\star}$ (IAN).
Bottom row shows results using Levina \& Bickel's MLE estimator, which was sensitive to the choice of $k$: using a small value grossly overestimated the dimension over the body, and a large $k$ ignored the geometry of the tail. NCD using IAN gave the best results, estimating dimension 2 for the body and 1 for the tail, with intermediate values for the transition tail-body and the boundary.
}
\label{fig:dims-stingray}
\end{figure}

\begin{figure}[h!]
\centering
  \includegraphics[width=1\textwidth]{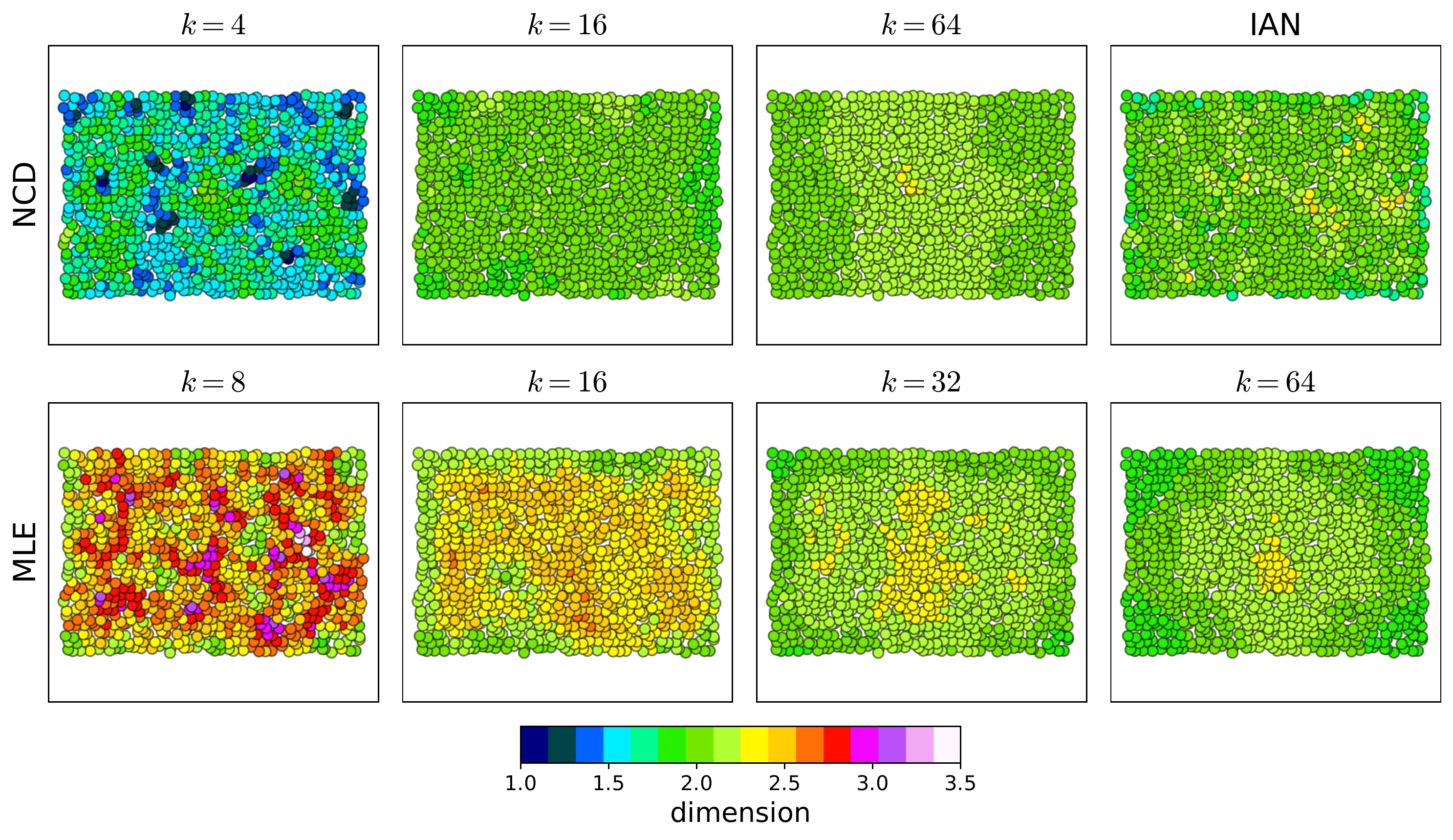}
\caption{Estimation of local intrinsic dimension on the bent plane dataset. As with the stingray (Figure~\ref{fig:dims-stingray}), results are sensitive to the choice of $k$, but here a wider range of values work due to the constant dimension. For NCD, results with IAN are comparable to those using the best $k$-NN graph ($k$ = 16). With MLE, larger $k$ improved results (comparable to those using NCD).
}
\label{fig:dims-bentplane}
\end{figure}

\begin{figure}[h!]
\centering
  \includegraphics[width=1\textwidth]{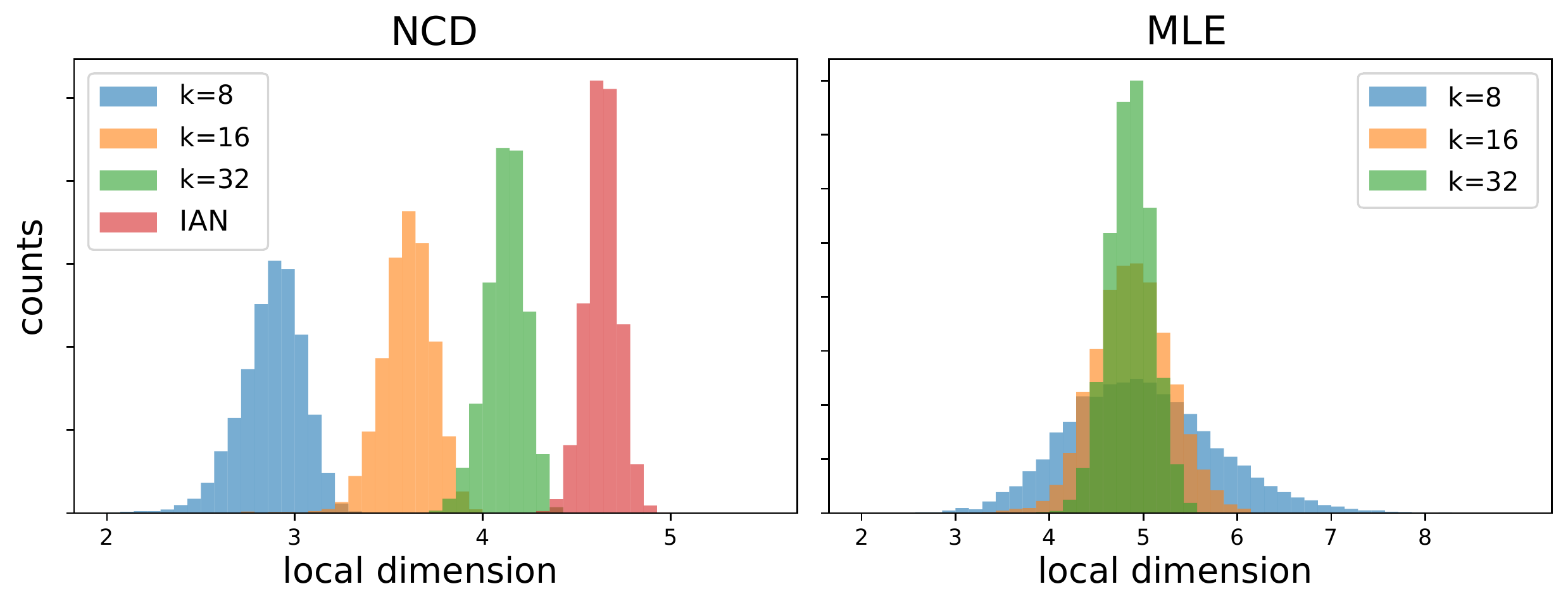}
\caption{Estimation of local intrinsic dimension for the 5-D cylinder dataset of Figure~\ref{fig:embeds-cylinder}. With NCD, results using IAN underestimated the true dimensionality (mean 4.63), but are still better than using a $k$-NN graph with arbitrary $k$. With MLE, larger values of $k$ gave tighter distributions centered near the correct value (mean 4.86 for $k$ = 32).
}
\label{fig:dims-cylinder}
\end{figure}

Results of applying our neighborhood correlation dimension (NCD) algorithm compared to Levina \& Bickel's MLE estimator (equation~\ref{eq:MLE}) are shown in Figures~\ref{fig:dims-stingray}--\ref{fig:dims-cylinder}. For NCD, we compared results using IAN against those from $k$-NN graphs using various values of $k$ (a range was chosen that included the best results for each algorithm). The IAN kernel was applied by using the discrete neighborhoods of $G^{\star}$, re-centered using neighbors-of-neighbors at most 3 hops away from $i$ (equation~\ref{eq:recentering}).

Using IAN, we obtained near-optimal results for the stingray and the bent plane. For the 5-dimensional cylinder, the dimension was underestimated (mean 4.6). Methods based on correlation dimension are known to underestimate the true $d$ when the sample size is not sufficiently large \citep{camastra2016intrinsic}. In these cases, the method of \citep{camastra2002estimating} can be applied \emph{a posteriori} to improve results.

For the MLE method, using large values of $k$ tended to improve results, but only when dimension was constant (as in the bent plane and cylinder datasets). For the stingray, however, no value of $k$ gave correct results: small values of $k$ increased the dimension estimates due to a bias, and large values tended to produce a uniform value throughout (thus giving better estimates only when $d$ is constant). We found that computing the neighborhood averages using the correction of \citet{mackay2005comments}, i.e., averaging the inverse of the estimators to reduce bias when $k$ is small, gave slightly better results. (We did not use the final smoothing procedure which involves choosing two additional neighborhood size parameters, $k_1$ and $k_2$.)

\begin{figure}[!ht]
\centering
  \includegraphics[width=1\textwidth]{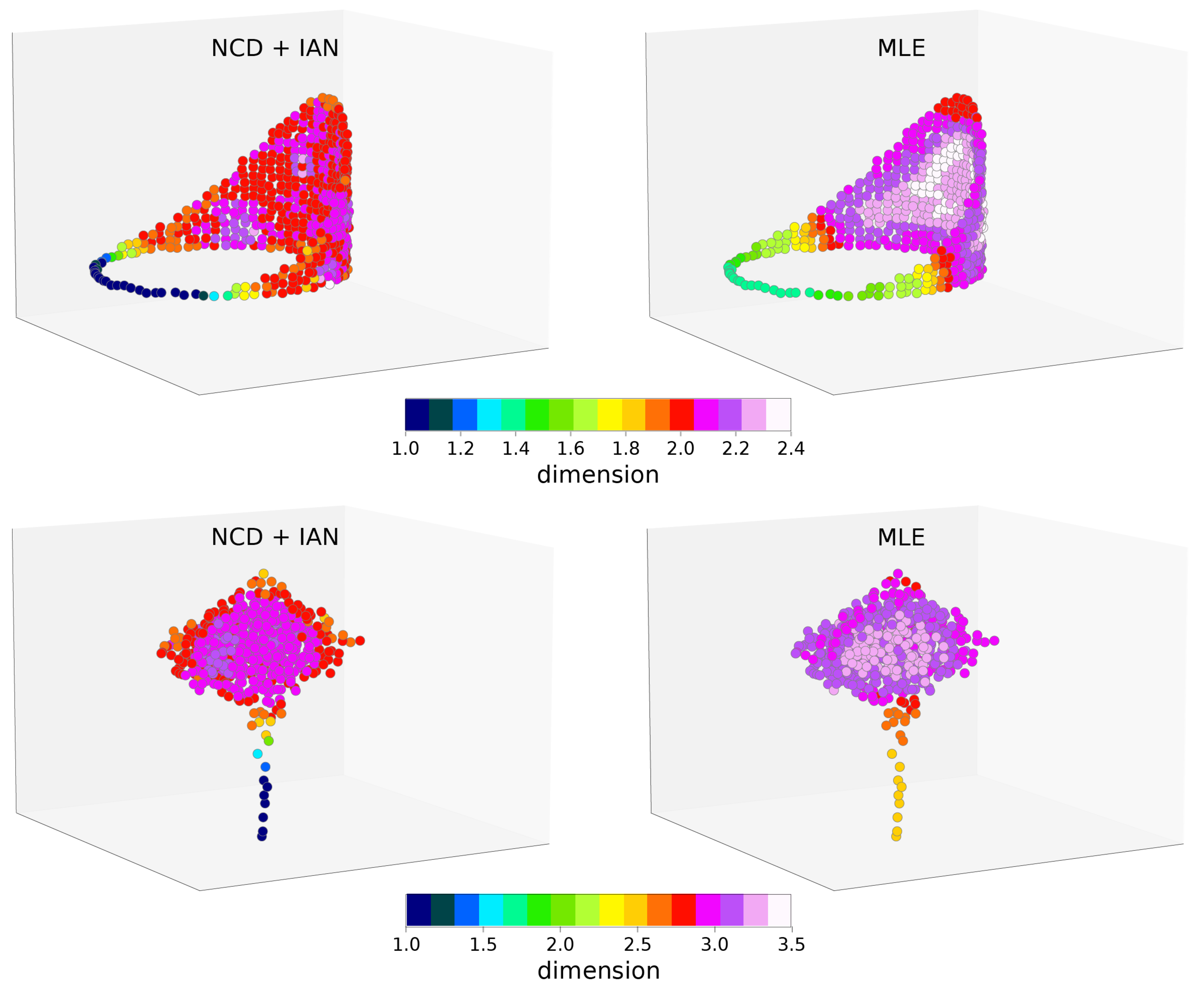}
\caption{Dimensionality estimation for two datasets with non-uniform local $d_i$: a ``tiara'' (top row), where dimension varies smoothly from 2 to 1, and a ``spinning top'' (bottom row, middle cross-section shown), where dimension reduces from 3 to 1 as one moves from the bulky part toward the tip. Using the optimal $k$ for MLE could not produce good results for the entire dataset (here, $k$ = 32 for both datasets). The NCD method, in contrast, was able to correctly adapt to the local geometry by taking advantage of the data graph produced by the IAN kernel.}
\label{fig:dims-crown-top}
\end{figure}

Finally, we confirmed these observations by testing two additional datasets with non-uniform dimensionality (Figure~\ref{fig:dims-crown-top}). Again, while our algorithm achieved good results locally, there was no single value of $k$ that allowed MLE to find appropriate local estimates everywhere.

\section{Summary and Conclusions}

In theory, applying the manifold assumption requires prior knowledge about the manifold: its geometry, topology, as well as how it was sampled. In practice, however, these manifold properties are rarely known.
Instead, one typically imposes an assumption about the manifold's dimension, $d$, which in turn suggests that $k = 2^d$ nearest neighbors should suffice. This is how many---most!---of the data graphs underlying manifold inference and non-linear dimensionality reduction are built. Since it is difficult to know whether this  assumption about dimension is accurate, it is common practice to test a few values of $k$ and choose among the results. 

Apart from the subjective nature of this choice, there are more general problems. Manifolds may not have a fixed dimension, they may be curved or with boundary, and sampling may vary. The intrinsic dimension may vary across the data, and so should the number of neighbors. In such cases, finding a compromise $k$ may be far from ideal. We suggest a different approach: that one should build the nearest-neighbor graph, and hence the graph-Laplacian approximation, in as data-driven a manner as possible, while being aware of the manifold properties.

Our algorithm of iterated adaptive neighborhoods (IAN) starts with a conservative assumption: that nearest neighbors should have no ``nearer'' neighbors between them. We then alternate between a discrete and a continuous view of neighborhood graphs, and use a volumetric statistic to check for outliers. A linear program keeps the scales minimal while providing a global cover. This optimization is convex, so results are deterministic; other approaches, such as t-SNE and UMAP, are stochastic, so depend critically on the initialization.

Our kernel has been applied successfully to a variety of datasets, and compared against some of the most popular algorithms available. In all cases our performance dominates. Furthermore, IAN can be incorporated directly into many embedding algorithms, including diffusion maps, Isomap, UMAP, and t-SNE, improving their results. Most of these algorithms involve several free parameters; we have none other than the robust requirement for an outlier.

Other popular embedding algorithms, e.g., LLE \citep{LLE}, approximate the tangent space over a local neighborhood around each point. Although not explored here, using $G^{\star}$ to automatically provide such neighborhoods is straightforward (analogous to what was done in section~\ref{section:dim-estimation} to estimate the local dimensionality). Applications to clustering need to be explored.

Our weighted graph has also been applied to geodesic estimation, achieving comparable results to those obtained from graph geodesics. In contrast, the graphs obtained from other similarity kernels produced less than optimal results.

Our unweighted graph has found application in local dimensionality estimation. Our proposed algorithm, neighborhood correlation dimension (NCD), takes advantage of the adaptive connectivity of our graph to improve results based on correlation dimension, namely by restricting the correlation integral to an approximately tubular neighborhood around $\bm{x}_i$ in $\mathcal{M}$. As a result, we obtained accurate estimates of the local dimension in datasets where it is not uniform.

Several theoretical bounds are implied throughout this paper; these need to be proved. Multiscale kernels, such as those from equations~\ref{eq:multiscale} and \ref{eq:bgh}, are known to approximate Laplacian operators asymptotically \citep{mjordan, BGH2}. Using our application examples as evidence, we conjecture that our version also results in good approximations.

In conclusion, understanding the interplay between manifold geometry, topology, and sampling lies at the heart of many data science applications. We have taken a first step to illustrate how discrete relates to continuous, how local estimates relate to global ones, and how uncertainties in data gathering relate to both. Applying data science in a way that leads to rigorous, scientifically-appropriate conclusions must take all of these into account.

\appendix
\section{Greedy splitting}\label{section:greedy}

As an alternative to the optimization from section~\ref{section:optimization} (which can be expensive when the number of edges in $G$ is very large, mainly due to large dimensionality), we have developed a greedy approach in which   scales that ``$C$-cover'' each edge $e_{ij}$ are assigned in decreasing order of length, $r_{ij}$ (the Euclidean distance between $\bm{x}_i$ and $\bm{x}_j$ in $\mathbb{R}^n$). We call this algorithm \emph{greedy splitting}.

Starting with the edge $e_{ij}$ with largest $r_{ij}$, set $\sigma_i = \sigma_j = Cr_{ij}$, with $C \leq 1$, thereby satisfying $\sigma_i\sigma_j \geq (Cr_{ij})^2$ with equality---we say $Cr_{ij}$ is evenly ``split'' between $\sigma_i$ and $\sigma_j$. Moreover, since $r_{ij} = r_i^{\mathrm{FN}} = r_j^{\mathrm{FN}}$, we know the constraints $\sigma_i \leq r_i^{\mathrm{FN}}$ and $\sigma_j \leq r_j^{\mathrm{FN}}$ are also satisfied.

Continue with the edge $e_{ij}$ that has the next largest length, $r_{ij}$. Here we are met with three possible cases in which a (re)assignment of scales is needed:
\begin{enumerate}
    \item If neither of the nodes have been assigned a scale yet, evenly split the scaled distance between $\sigma_i$ and $\sigma_j$, as above.
    \item If one of the nodes does not have a scale yet (without loss of generality, let that node be $j$), set $\sigma'_{j}$ to the minimum scale that ensures $\sigma_{i}\sigma'_{j} \geq (Cr_{ij})^2$, i.e., $\sigma'_{j} = (Cr_{ij})^2/\sigma_{i}$;
    \item If both nodes have previously been assigned a scale but $e_{ij}$ is not $C$-covered by the current values of $\sigma_{i}$ and $\sigma_{j}$, then set the quotient $a = \frac{Cr_{ij}}{\sqrt{\sigma_{i}\sigma_{j}}}$ and update both scales: $\sigma'_{i} = a\sigma_{i}$ and $\sigma'_{j} = a\sigma_{j}$, thereby evenly splitting the quotient between the two nodes.
\end{enumerate}
After cases (2) and (3), the updated scales might need to be ``rebalanced'' in order to meet the constraints $\sigma'_i \leq r_i^{\mathrm{FN}}$ and $\sigma'_j \leq r_j^{\mathrm{FN}}$. Without loss of generality, let $\sigma'_i > r_i^{\mathrm{FN}}$. Then, we set $\sigma''_i = r_i^{\mathrm{FN}}$ and $\sigma''_j = \sigma'_j\frac{\sigma'_i}{\sigma''_i}$. Only one of the two scales may exceed its upper bound: in (2), this is trivially true since only the newly-assigned scale may be greater than $Cr_{ij}$; in (3), since both $\sigma_i$ and $\sigma_j$ have been previously assigned, we have $\sigma_i \leq r_i^{\mathrm{FN}}$ and $\sigma_j \leq r_j^{\mathrm{FN}}$, as well as $r_{ij} \leq r_i^{\mathrm{FN}}$ and $r_{ij} \leq r_j^{\mathrm{FN}}$, so therefore it must be the case that $r_i^{\mathrm{FN}}r_j^{\mathrm{FN}} \geq r_{ij}^2 = \sigma'_i \sigma'_j$. Note that, as a corollary, both scales must meet their respective constraints after being re-balanced as above.

The above is repeated until all edges have been visited. By covering the largest edges first, we assign the largest, most constrained scales first, allowing for the later, less constrained scales, to be as small as possible. Because in most cases this tends to evenly split the scaled edge lengths $Cr_{ij}$ between $\sigma_i$ and  $\sigma_j$, the algorithm produces reasonable (but usually sub-optimal) results when compared to the linear program of section~\ref{section:constraints}. 

\bibliography{bibfile}

\end{document}